\documentclass[10pt,journal,compsoc]{IEEEtran}
\usepackage{blindtext}
\usepackage{times}
\usepackage{epsfig}
\usepackage{amssymb}
\usepackage{amsthm}
\usepackage{graphicx}
\usepackage{amsfonts}
\usepackage{amsmath}
\usepackage{array}
\usepackage{mdwmath}
\usepackage{mdwtab}
\usepackage{eqparbox}
\usepackage{fixltx2e}
\usepackage{stfloats}
\usepackage{url}
\usepackage{diagbox}
\usepackage{verbatim}
\usepackage{booktabs}
\usepackage{multirow}
\usepackage{bm}
\usepackage{mathtools}
\usepackage{breqn}
\usepackage{float}
\usepackage{tcolorbox}
\usepackage{todonotes}
\usepackage{pgfmath}
\usepackage{siunitx}
\usepackage{adjustbox}
\usepackage{threeparttable}
\usepackage{hyperref}

\usepackage[font=footnotesize,labelfont=bf]{caption}
\usepackage{subcaption}

\usepackage[mode=buildnew]{standalone} % for directly including standalone diagrams directly. requires -shell-escape

\def\etal{{\it et al. }}
\theoremstyle{definition}

\usepackage{threeparttable}
\usepackage{flushend}

\ifCLASSOPTIONcompsoc
  \usepackage[nocompress]{cite}
\else
  \usepackage{cite}
\fi

\DeclareMathOperator*{\argmin}{arg\,min}

\def\Minimize{\mathop{\rm minimize}\limits}
\def\st{\mathop{\rm subject\ to}}

\def\ourmethod{NAT}
\def\ourmodel{NATNet}

\usepackage{xcolor}

\definecolor{color_blue}{RGB}{167, 222, 250}
\definecolor{color_yellow}{RGB}{237, 211, 83}
\definecolor{color_green}{RGB}{217,255,219}
\definecolor{color_red}{RGB}{237, 150, 83}

\definecolor{gautamblue}{RGB}{215, 238, 247}
\definecolor{gautamred}{RGB}{255, 186, 184}
\definecolor{gautamgreen}{RGB}{217,255,219}
\definecolor{gautamorange}{RGB}{255, 219, 153}
\definecolor{gautamviolet}{RGB}{224, 220, 244}

\def\HiLiBlue{\leavevmode\rlap{\hbox to \hsize{\color{gautamblue}\leaders\hrule height .8\baselineskip depth .5ex\hfill}}}
\def\HiLiYellow{\leavevmode\rlap{\hbox to \hsize{\color{gautamorange}\leaders\hrule height .8\baselineskip depth .5ex\hfill}}}
\def\HiLiGreen{\leavevmode\rlap{\hbox to \hsize{\color{gautamgreen}\leaders\hrule height .8\baselineskip depth .5ex\hfill}}}
\def\HiLiRed{\leavevmode\rlap{\hbox to \hsize{\color{gautamred}\leaders\hrule height .8\baselineskip depth .5ex\hfill}}}

\usepackage{algorithmic}
\usepackage[ruled, vlined, linesnumbered]{algorithm2e}

\begin{document}

\title{Neural Architecture Transfer}
\author{Zhichao Lu,~\IEEEmembership{Student Member,~IEEE,} Gautam Sreekumar, Erik Goodman, Wolfgang Banzhaf, \\Kalyanmoy Deb,~\IEEEmembership{Fellow, IEEE} and Vishnu~Naresh~Boddeti,~\IEEEmembership{Member, IEEE}% <-this % stops a space
\thanks{Z. Lu is with Southern University of Science and Technology, Shenzhen, China. The majority of this work was done when Z. Lu was with Michigan State University. E-mail: luzc@sustech.edu.cn,}
\thanks{G. Sreekumar, E. Goodman, W. Banzhaf, K. Deb, and V. N. Boddeti are with Michigan State University, East Lansing, MI, 48824 USA. E-mail: \{sreekum1,goodman,banzhafw,kdeb,vishnu\}@msu.edu.}
}

\IEEEtitleabstractindextext{%
\begin{abstract}
Neural architecture search (NAS) has emerged as a promising avenue for automatically designing task-specific neural networks. Existing NAS approaches require one complete search for each deployment specification of hardware or objective. This is a computationally impractical endeavor given the potentially large number of application scenarios. In this paper, we propose \emph{Neural Architecture Transfer} (\ourmethod{}) to overcome this limitation. \ourmethod{} is designed to efficiently generate task-specific custom models that are competitive under multiple conflicting objectives. To realize this goal we learn task-specific supernets from which specialized subnets can be sampled without any additional training. The key to our approach is an integrated online transfer learning and many-objective evolutionary search procedure. A pre-trained supernet is iteratively adapted while simultaneously searching for task-specific subnets. We demonstrate the efficacy of \ourmethod{} on 11 benchmark image classification tasks ranging from large-scale multi-class to small-scale fine-grained datasets. In all cases, including ImageNet, \ourmodel{}s improve upon the state-of-the-art under mobile settings ($\leq$ 600M Multiply-Adds). Surprisingly, small-scale fine-grained datasets benefit the most from \ourmethod{}. At the same time, the architecture search and transfer is orders of magnitude more efficient than existing NAS methods. Overall, experimental evaluation indicates that, across diverse image classification tasks and computational objectives, \ourmethod{} is an appreciably more effective alternative to \textcolor{black}{conventional transfer learning of fine-tuning weights of an existing network architecture learned on standard datasets}. Code is available at \url{https://github.com/human-analysis/neural-architecture-transfer}.
\end{abstract}

}

\maketitle
\IEEEdisplaynontitleabstractindextext
\IEEEpeerreviewmaketitle

\section{Introduction}

\IEEEPARstart{I}{mage} classification is a fundamental task in computer vision, where given a dataset and, possibly, multiple objectives to optimize, one seeks to learn a model to classify images. Solutions to address this problem fall into two categories: (a) Sufficient Data: A custom convolutional neural network architecture is designed and its parameters are trained from scratch using variants of stochastic gradient descent, and (b) Insufficient Data: An existing architecture designed on a large scale dataset, such as ImageNet~\cite{imagenet}, along with its pre-trained weights (e.g., VGG\cite{vgg}, ResNet\cite{resnet}), is fine-tuned for the task at hand. These two approaches have emerged as the mainstays of present day computer vision.

Success of the aforementioned approaches is primarily attributed to architectural advances in convolutional neural networks. Initial efforts at designing neural architectures relied on human ingenuity. Steady advances by skilled practitioners has resulted in designs, such as AlexNet\cite{alexnet}, VGG\cite{vgg}, GoogLeNet\cite{googlenet}, ResNet\cite{resnet}, DenseNet\cite{densenet} and many more, which have led to performance gains on the ImageNet Large Scale Visual Recognition Challenge \cite{imagenet}. In most other cases, a recent large scale study \cite{transfer_learning} has shown that, across many tasks, transfer learning by fine-tuning ImageNet pre-trained networks outperforms networks that are trained from scratch on the same data.

Moving beyond manually designed network architectures, Neural Architecture Search (NAS) \cite{nasnet} seeks to automate this process and find not only good architectures, but also their associated weights for a given image classification task. This goal has led to notable improvements in convolutional neural network architectures on standard image classification benchmarks, such as ImageNet, CIFAR-10~\cite{cifar}, CIFAR-100~\cite{cifar} etc., in terms of predictive performance, computational complexity and model size. However, apart from transfer learning by fine-tuning the \emph{weights}, current NAS approaches have failed to deliver new models for both \emph{weights} and \emph{topology} on custom non-standard datasets. The key barrier to realizing the full potential of NAS is the large data and computational requirements for employing existing NAS algorithms on new tasks.

In this paper, we introduce \emph{Neural Architecture Transfer} (\ourmethod{}) to breach this barrier. Given an image classification task, \ourmethod{} obtains custom neural networks (both \emph{topology} and \emph{weights}), optimized for possibly many conflicting objectives, and does so without the steep computational burden of running NAS for each new task from scratch. A single run of \ourmethod{} efficiently obtains multiple custom neural networks spanning the entire trade-off front of objectives.

Our solution builds upon the concept of a supernet \cite{one-shot} which comprises of many subnets. All subnets are trained simultaneously through weight sharing, and can be sampled very efficiently. This procedure decouples the network training and the search phases of NAS. A many-objective\footnote{Problems having more than three objectives are called many-objective problems \cite{nsga3}.} search can then be employed on top of the supernet to find all network architectures that provide the best trade-off among the objectives. However, training such supernets for each task from scratch is very computationally and data intensive. The key idea of \ourmethod{} is to leverage an existing supernet and efficiently transfer it into a task-specific supernet, whilst simultaneously searching for architectures that offer the best trade-off between the objectives of interest. Therefore, unlike standard supernet-based NAS, we combine supernet transfer learning with the search process. At the conclusion of this process, \ourmethod{} returns (i) subnets that span the entire objective trade-off front, and (ii) a task-specific supernet. The latter can now be utilized for all future deployment-specific NAS, i.e., new and different hardware or objectives, without any additional training.

\begin{figure*}[t]
    \centering
    \begin{subfigure}{0.95\textwidth}
    \centering
    \includegraphics[width=\textwidth]{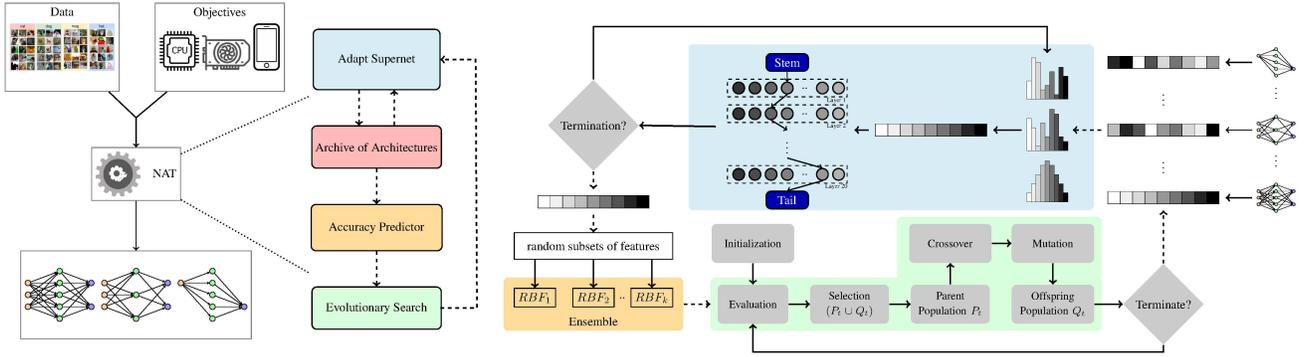}
    \end{subfigure}
\caption{\textbf{Overview:} Given a dataset and objectives to optimize, \ourmethod{} designs custom architectures spanning the objective trade-off front. \ourmethod{} comprises of two main components, supernet adaptation and evolutionary search, that are iteratively executed. \ourmethod{} also uses an online accuracy predictor model to improve its computational efficiency.
\label{fig:overview}}
\vspace{-0.3cm}
\end{figure*}

The core of \ourmethod{}'s efficiency lies in only adapting the subnets of the supernet that will lie on the efficient trade-off front of the new dataset, instead of all possible subnets. But, the structure of the corresponding subnets is unknown before adaptation. We resolve this ``chicken-and-egg problem" by adopting an online procedure that alternates between the two primary stages of NAT: (a) \emph{supernet adaptation} of subnets that are at the current trade-off front, and (b) \emph{evolutionary search} for subnets that span the many-objective trade-off front. A pictorial overview of the entire \ourmethod{} method is shown in Fig.\ref{fig:overview}. 

In the \emph{adaptation} stage, we first construct a layer-wise empirical distribution from the promising subnets returned by evolutionary search. Then, subnets sampled from this distribution are fine-tuned. In the \emph{search} stage, to improve the efficiency of the search, we adopt a surrogate model to quickly predict the objectives of any sampled subnet without a full-blown and costly evaluation. Furthermore, the predictor model itself is also learned online from previously evaluated subnets. We alternate between these two stages until our computational budget\footnote{We manually set the computational budget to a maximum of 1 day on a 8-GPU (NVIDIA 2080Ti) server. This is equivalent to the computational resources available to a small lab.} is exhausted. The key contributions of this paper are:

\vspace{2pt}
\noindent\textbf{--} We introduce \emph{Neural Architecture Transfer} as a NAS-powered alternative to fine-tuning based transfer learning. \ourmethod{} is powered by a simple, yet highly effective online supernet fine-tuning and online accuracy predicting surrogate model.

\vspace{2pt}
\noindent\textbf{--} We demonstrate the scalability and practicality of \ourmethod{} on multiple datasets corresponding to different scenarios; large-scale multi-class (ImageNet~\cite{imagenet}, CINIC-10\cite{cinic10}), medium-scale multi-class (CIFAR-10, CIFAR-100~\cite{cifar}), small-scale multi-class (STL-10\cite{stl-10}), large-scale fine-grained (Food-101\cite{food-101}), medium-scale fine-grained (Stanford Cars\cite{stanford_cars}, FGVC Aircraft\cite{aircraft}) and small-scale fine-grained (DTD\cite{dtd},  Oxford-IIIT Pets\cite{pets}, Oxford Flowers102~\cite{flowers102}) datasets.

\vspace{2pt}
\noindent\textbf{--} Under mobile settings ($\leq$ 600M MAdds), \ourmodel{}s lead to state-of-the-art performance across all these tasks. For instance, on ImageNet, \ourmodel{} achieves a Top-1 accuracy of 80.5\% at 600M MAdds. 

\vspace{2pt}
\noindent\textbf{--} We also demonstrate the utility of \ourmethod{} in searching for a backbone for semantic segmentation, a dense prediction task. On Cityscapes \cite{cityscapes}, \ourmethod{} matches the mIoU performance of Auto-DeepLab \cite{autodeeplab} while using 4$\times$ fewer MAdds.

\vspace{2pt}
\noindent\textbf{--} Finally we demonstrate the scalability and utility of \ourmethod{} across many objectives and on dense image prediction. Optimizing accuracy, model size and one of MAdds, CPU or GPU latency, \ourmodel{}s dominate MobileNetV3~\cite{mobilenetv3} across all objectives. We also consider a 12 objective problem of finding a common architecture across eleven datasets while minimizing MAdds. The best trade-off \ourmodel{} dominates all models across these datasets under mobile settings.

\section{Related Work}
\begin{table*}[!t]
\caption{Comparison of NAT and existing NAS methods. \textcolor{black}{$^\dagger$ indicates methods that scalarize multiple objectives into one composite objective or as an additional constraint, see text for details.}}
\label{tab:related-work}
\centering
\resizebox{0.8\textwidth}{!}{%
\begin{tabular}{@{\hspace{2mm}}l|c|ccc|c@{\hspace{2mm}}}
\toprule
Methods \hspace{1mm} & \hspace{1mm} \begin{tabular}[c]{@{}c@{}}Search Method \\ \end{tabular} \hspace{1mm} & \hspace{1mm} \begin{tabular}[c]{@{}c@{}} Performance \\ Prediction \end{tabular} & \hspace{1mm} \begin{tabular}[c]{@{}c@{}} Weight \\ Sharing \end{tabular} & \hspace{1mm} \begin{tabular}[c]{@{}c@{}} Multiple\\Objective \end{tabular} \hspace{1mm} & \begin{tabular}[c]{@{}c@{}}Dataset Searched \\ \end{tabular} \\ \midrule
NASNet \cite{nasnet} & RL &  &  &  & C10 \\
PNAS \cite{PNAS} & SBMO & \checkmark &  &  & C10 \\
DARTS \cite{darts} & gradient &  & \checkmark &  & C10 \\
LEMONADE \cite{LEMONADE} & EA &  & \checkmark & \checkmark & C10, C100, ImageNet64 \\
ProxylessNAS \cite{proxylessnas} & RL / \textcolor{black}{gradient} &  & \checkmark & \textcolor{black}{\checkmark$^{\dagger}$} & \textcolor{black}{C10}, ImageNet \\
MnasNet \cite{mnasnet} & RL &  &  & \textcolor{black}{\checkmark$^{\dagger}$} & ImageNet \\
EfficientNet \cite{efficientnet} & RL+scaling &  &  &  & ImageNet \\
ChamNet \cite{chamnet} & EA & \checkmark &  & \textcolor{black}{\checkmark$^{\dagger}$} & ImageNet \\ 
MobileNetV3 \cite{mobilenetv3} & RL+expert &  &  & \textcolor{black}{\checkmark$^{\dagger}$} & ImageNet \\
\textcolor{black}{SPOS NAS} \cite{guo2019single} & \textcolor{black}{EA} &  & \textcolor{black}{\checkmark} & \textcolor{black}{\checkmark$^{\dagger}$} & \textcolor{black}{ImageNet} \\ 
OnceForAll \cite{onceforall} & EA & \checkmark & \checkmark & \textcolor{black}{\checkmark$^{\dagger}$} & ImageNet \\
FBNetV2 \cite{fbnetv2} & gradient &  & \textcolor{black}{\checkmark} & \textcolor{black}{\checkmark$^{\dagger}$} & ImageNet \\
\midrule
\textbf{\ourmethod{} (this paper)} & EA+transfer & \checkmark & \checkmark & \checkmark & \begin{tabular}[c]{@{}c@{}}ImageNet, C10, C100,\\CINIC-10, STL-10, Flowers102,\\Pets, DTD, Cars, Aircraft, Food-101\end{tabular} \\ \bottomrule
\end{tabular}%
}
\vspace{-0.2cm}
\end{table*}
Recent years have witnessed growing interests in neural architecture search. The promise of being able to automatically search for task-dependent network architectures is particularly appealing as deep neural networks are widely deployed in diverse applications and computational environments. Early methods \cite{yao1999evolving,stanley2002evolving} made efforts to simultaneously evolve the topology of neural networks along with weights and hyperparameters. These methods perform competitively with hand-crafted networks on simple control tasks with shallow fully connected networks. Recent efforts \cite{nas} primarily focus on designing deep convolutional neural network architectures. 

The development of NAS largely happened in two phases. Starting from NASNet \cite{nasnet}, the focus of the first wave of methods was primarily on improving the predictive accuracy of CNNs including Block-QNN \cite{block-qnn}, Hierarchical NAS \cite{liu2018hierarchical}, and AmoebaNet \cite{amoebanet}, etc. These methods relied on Reinforcement Learning (RL) or Evolutionary Algorithm (EA) to search for an optimal modular structure that is repeatedly stacked together to form a network architecture. The search was typically carried out on relatively small-scale datasets (e.g. CIFAR-10/100 \cite{cifar}), following which the best architectures were transferred to ImageNet for validation. A steady stream of improvements over state-of-the-art on numerous datasets were reported. The focus of the second wave of NAS methods was on improving the search efficiency.

A few methods have also been proposed to adapt NAS to other scenarios. These include meta-learning based approaches \cite{lian2020towards,elsken2020meta} with application to few-shot learning tasks. XferNAS~\cite{wistuba2019xfernas} and EAT-NAS~\cite{fang2019eat} illustrate how architectures can be transferred between similar datasets or from smaller to larger datasets. Some approaches \cite{wong2018transfer,kokiopoulou2019fast} proposed RL-based NAS methods that search on multiple tasks during training and transfer the learned search strategy, as opposed to searched networks, to new tasks at inference. Next, we provide short overviews on methods that are closely related to the technical approach in this paper. Table~\ref{tab:related-work} provides a comparative overview of \ourmethod{} to existing NAS approaches.

\vspace{3pt}
\noindent\textbf{Performance Prediction:} Evaluating the performance of an architecture requires a computationally intensive process of iteratively optimizing model weights. To alleviate this computational burden, regression models have been learned to predict an architecture's performance without actually training it. Baker \etal \cite{baker2017accelerating} use a radial basis function to estimate the final accuracy of architectures from its accuracy in the first 25\% of training iterations. PNAS \cite{PNAS} uses a multilayer perceptron (MLP) and a recurrent neural network to estimate the expected improvement in accuracy if the current modular structure (which is later stacked together to form a network) is expanded with a new branch. Conceptually, both of these methods seek to learn a prediction model that extrapolate (rather than interpolate), resulting in poor correlation in prediction. OnceForAll \cite{onceforall} also uses a MLP to predict accuracy from architecture encoding. However, the model is trained offline for the entire search space, thereby requiring a large number of samples for learning (16K samples -$>$ 2 GPU-days for just constructing the surrogate model). Instead of using uniformly sampled architectures to train the prediction model to approximate the entire landscape, ChamNet \cite{chamnet} trains many architectures through full SGD and selects only 300 samples of high accuracy with diverse efficiency (Multiply-adds, Latency, Energy) to train a prediction model offline. In contrast, \ourmethod{} learns a prediction model in an online fashion only on the samples at the current trade-off front as we explore the search space. Such an approach only needs to interpolate over a much smaller space of architectures constituting the current trade-off front. Consequently, this procedure significantly improves both the accuracy and the sample complexity of constructing the prediction model.

\vspace{3pt}
\noindent\textbf{Weight Sharing:} Approaches in this category involve training a \emph{supernet} that contains all searchable architectures as its subnets. They can be broadly classified into two categories depending on whether the supernet training is coupled with architecture search or decoupled into a two-stage process. Approaches of the former kind \cite{enas,darts,proxylessnas} are computationally efficient but return sub-optimal models. Numerous studies \cite{li2019random,xie2019exploring,Yu2020Evaluating} allude to weak correlation between performance at the search and final evaluation stages. Methods of the latter kind \cite{smash,one-shot,onceforall} use performance of subnets (obtained by sampling the trained supernet) as a metric to select architectures during search. However, training a supernet beforehand for each new task is computationally prohibitive. In this work, we take an integrated approach where we train a supernet on large-scale datasets (e.g. ImageNet) once and couple it with our architecture search to quickly adapt it to a new task. An elaborated discussion connecting our method to existing approaches is provided in Section~\ref{sec:one-shot}.

\vspace{3pt}
\noindent\textbf{Multi-Objective NAS:} Methods that consider multiple objectives for designing hardware specific models have also been developed. The objectives are optimized either through (i) scalarization, or (ii) Pareto-based solutions. The former include, ProxylessNAS \cite{proxylessnas}, MnasNet \cite{mnasnet}, ChamNet \cite{chamnet}, MobileNetV3 \cite{mobilenetv3}, and FBNetV2 \cite{fbnetv2} which use a scalarized objective or an additional constraint to encourage high accuracy and penalize compute inefficiency at the same time, e.g., maximize $Acc * (Latency / Target)^{-0.07}$. Conceptually, the search of architectures is still guided by a single objective and only one architecture is obtained per search. Empirically, multiple runs with different weighting of the objectives are needed to find an architecture with the desired trade-off, or multiple architectures with different complexities. Methods in the latter category include \cite{NSGANet,LEMONADE,dppnet,chu2019fairnas,muxconv} and aim to approximate the entire Pareto-efficient frontier simultaneously---i.e. multiple architectures with different complexities are obtained in a single run. These approaches rely on heuristics (e.g., EA) to efficiently navigate the search space allowing practitioners to visualize the trade-off between the objectives and to choose a suitable network \emph{a posteriori} to the search. \ourmethod{} falls into the latter category and uses an accuracy prediction model and weight sharing for efficient architecture transfer to new tasks.

\section{Proposed Approach}

\emph{Neural Architecture Transfer} consists of three main components: an accuracy predictor, an evolutionary search routine, and a supernet. \ourmethod{} starts with an archive $\mathcal{A}$ of architectures (subnets) created by uniform sampling from our search space. We evaluate the performance $f_i$ of each subnet ($\bm{a}_i$) using weights inherited from the supernet. The accuracy predictor is then constructed from $(\bm{a}_i, f_i)$ pairs which (jointly with any additional objectives provided by the user) drives the subsequent many-objective evolutionary search towards  optimal architectures.  Promising architectures at the conclusion of the evolutionary process are added to the archive $\mathcal{A}$. The (partial) weights of the supernet corresponding to the top-ranked subnets in the archive are fine-tuned. \ourmethod{} repeats this process for a pre-specified number of iterations. At the conclusion, we output both the archive and the task-specific supernet. Networks that offer the best trade-off among the objectives can be post-selected from the archive. Detailed descriptions of each component of \ourmethod{} are provided in the following subsections. Figure~\ref{fig:overview} and Algorithm~\ref{algo:framework} provide an overview of our entire approach.

\begin{algorithm}[tbh]
\SetAlgoLined
\SetKwInOut{Input}{Input}
\SetKwInOut{Output}{Output}
\SetKwFor{For}{for}{do}{end for}
\footnotesize
\Input{Training data $\mathcal{D}_{trn}$, validation data $\mathcal{D}_{vld}$,\\additional objectives $\tilde{f}$, supernet $\mathcal{S}_w$, archive size $N$, \\ \# of iterations $T$, \# of epochs $E$, \# of generations $G$.}
    $t$ $\leftarrow$ 0 \textcolor{gray}{// initialize an iteration counter}.\\
    $\mathcal{A}$ $\leftarrow$ randomly initialize an archive of archs with a size of $N$.\\
    \While{$t < T$}{
        \textcolor{gray}{// compute accuracy by inheriting weights and inference.}\\
        $f \leftarrow \mathcal{S}_w(\mathcal{A}, \mathcal{D}_{vld})$\\
        \textcolor{gray}{// construct the accuracy predictor.}\\
        \HiLiYellow $\mathcal{S}_f \leftarrow$ \emph{Accuracy Predictor}($\mathcal{A}, f$) \hspace{20mm} $\triangleleft$ Algo.~\ref{algo:acc_predictor}\\
        \textcolor{gray}{// find promising archs by evolutionary search.}\\
        \HiLiGreen $P_t \leftarrow$ \emph{Evolutionary Search}($\mathcal{S}_f$, $\tilde{f}$, $\mathcal{A}$, $G$) \hspace{10mm} $\triangleleft$ Algo.~\ref{algo:evolution}\\
        \textcolor{gray}{// keep the top-$N$ ranked archs in archive.}\\
        \HiLiRed $\mathcal{A} \leftarrow$ \emph{Selection}($\mathcal{A} \cup P_t, N$) \hspace{25mm} $\triangleleft$ Algo.~\ref{algo:nsga3}\\
        \textcolor{gray}{// fine tune supernet to promising archs.}\\
        \HiLiBlue $\mathcal{S}_w \leftarrow$ \emph{Adapt}($\mathcal{S}_w, \mathcal{A}, \mathcal{D}_{trn}, E$) \hspace{21mm} $\triangleleft$ Algo.~\ref{algo:supernet}\\
        $t$ $\leftarrow$ $t + 1$
    }
    \textcolor{gray}{// optional in case of no preferences from users.}\\
    $\mathcal{A}^*$ $\leftarrow$ choose a subset of archs from $\mathcal{A}$ based on trade-offs by method presented in Section~\ref{sec:decision}.\\
\textbf{Return} $\mathcal{S}_w, \mathcal{A}, \mathcal{A}^*$.
\caption{Neural Architecture Transfer\label{algo:framework}}
\end{algorithm}

\subsection{Problem Formulation}

The problem of neural architecture search for a target dataset $\mathcal{D} = \{\mathcal{D}_{trn}, \mathcal{D}_{vld}, \mathcal{D}_{tst}\}$ with many objectives can be formulated as the following bilevel optimization problem \cite{bilevel},
\begin{equation}
\begin{aligned}
\Minimize & \hspace{3mm} \bm{F}(\bm{a}) = \big(f_1(\bm{a}; \bm{w}^*(\bm{a})), \ldots, f_m(\bm{a}; \bm{w}^*(\bm{a}))\big)^T, \\
\st  & \hspace{3mm} \bm{w}^*(\bm{a}) \in \argmin~\mathcal{L}(\bm{w};\bm{a}), \\
     & \hspace{3mm} \bm{a} \in \mathbf{\Omega}_{a}, \hspace{3mm} \bm{w} \in \mathbf{\Omega}_{w},
\end{aligned}
\label{def:nas}
\end{equation}
where the upper-level variable $\bm{a}$ defines a candidate architecture, and the lower-level variable $\bm{w}(\bm{a})$ denotes its associated weights. $\mathcal{L}(\bm{w};\bm{a})$ is the cross-entropy loss on the training data $\mathcal{D}_{trn}$ for an architecture $\bm{a}$. $\bm{F}: \mathbf{\Omega} \rightarrow \mathbb{R}^m$ constitutes $m$ (user-) desired, possibly competing, objectives---e.g., predictive performance on validation data $\mathcal{D}_{vld}$, number of parameters (\#Params), multiply-adds (\#MAdds), latency / power consumption / memory footprint on specific hardware etc.

The bi-level optimization is typically solved in an iterative fashion, with an inner optimization loop over the weights of the network for a given architecture, and an outer optimization loop over the network architectures themselves. The computational challenge of solving this problem stems from both the upper and lower level optimization. Learning optimal weights of a network in the lower level necessitates costly iterations of stochastic gradient descent over multiple epochs. Similarly, searching the optimal architecture on the upper level is prohibitive due to the discrete nature of the architecture description, size of search space and our desire to optimize many, possibly conflicting, objectives.

\subsection{Search Space and Encoding}
\begin{figure}[t]
    \centering
    \begin{subfigure}{0.48\textwidth}
    \centering
    \includegraphics[width=\textwidth]{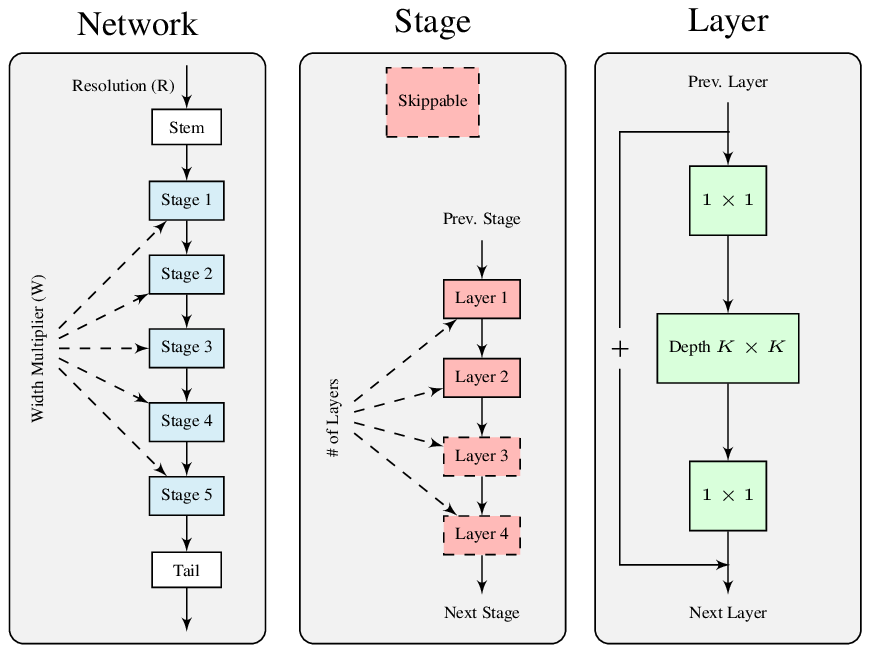}
    \caption{Search Space\label{fig:search_space}}
    \end{subfigure}
    \begin{subfigure}{0.48\textwidth}
    \centering
    \vspace{1em}
    \includegraphics[width=0.8\textwidth]{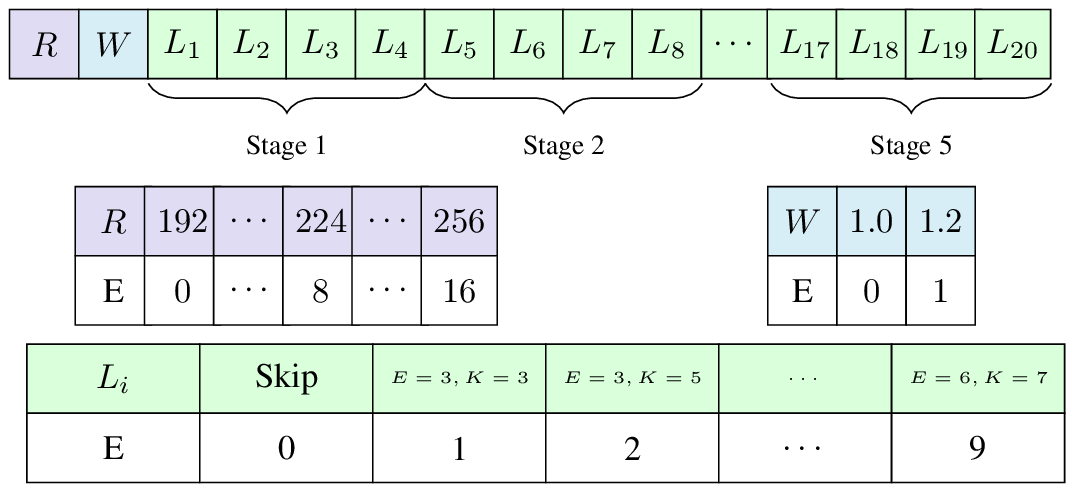}
    \caption{Encoding\label{fig:encoding}}
    \end{subfigure}
\caption{The architectures in our search space are variants of MobileNetV2 family of models \cite{mobilenetv2,mnasnet,efficientnet,mobilenetv3}. (a) Each networks consists of five stages. Each stage has two to four layers. Each layer is an inverted residual bottleneck block. The search space includes, input image resolution (R), width multiplier (W), the number of layers in each stage, the \# of output channels (expansion ratio E) of the first $1\times1$ convolution and the kernel size (K) of the depth-wise separable convolution in each layer. (b) Networks are represented as 22-integer strings, where the first two correspond to resolution and width multiplier, and the rest correspond to the layers. Each value indicates a choice, e.g. the third integer ($L_1$) takes a value of ``1'' corresponds to using expansion ratio of 3 and kernel size of 3 in layer 1 of stage 1. \label{fig:search_encode}}
\vspace{-0.3cm}
\end{figure}

The search for optimal network architectures can be performed over many different search spaces. The generality of the chosen search space has a major influence on the quality of results that are feasible. We adopt a modular design for overall structure of the \emph{network}, consisting of a stem, multiple stages and a tail (see Fig.~\ref{fig:search_space}). The \emph{stem} and \emph{tail} are common to all networks and not searched. Each \emph{stage} in turn comprises of multiple layers, and each \emph{layer} itself is an inverted residual bottleneck structure \cite{mobilenetv2}.

\noindent\textbf{-Network:} We search for the input image resolution and the width multiplier (a factor that scales the \# of output channels of each layer uniformly \cite{mobilenet}). Following previous work \cite{mnasnet,efficientnet,onceforall}, we segment the CNN architecture into five sequentially connected stages. The stages gradually reduce the feature map size and increase the number of channels (Fig.~\ref{fig:search_space} \emph{Left}).

\noindent\textbf{-Stage:} We search over the number of layers, where only the first layer uses stride 2 if the feature map size decreases, and we allow each block to have minimum of two and maximum of four layers (Fig.~\ref{fig:search_space} \emph{Middle}).

\noindent\textbf{-Layer:} We search over the expansion ratio (between the \# of output and input channels) of the first $1\times1$ convolution and the kernel size of the depth-wise separable convolution (Fig.~\ref{fig:search_space} \emph{Right}).

Overall, we search over four primary hyperparameters of CNNs i.e., the depth (\# of layers), the width (\# of channels), the kernel size, and the input resolution. The resulting volume of our search space is approximately \num{3.5E+19} for each combination of image resolution and width multiplier.

To encode these architectural choices, we use an integer string of length 22, as shown in Fig.~\ref{fig:encoding}. The first two values represent the input image resolution and width multiplier, respectively. The remaining 20 values denote the expansion ratio and kernel size settings for each of the 20 layers. The available options for expansion ratio and kernel size are [3, 4, 6] and [3, 5, 7], respectively. It is worth noting that we sort the layer settings in ascending \#MAdds order, which is beneficial to the mutation operator used in our evolutionary search algorithm.

\subsection{Accuracy Predictor}
The main computational bottleneck of NAS arises from the nested nature of the bi-level optimization problem. The inner optimization requires the weights of the subnets to be thoroughly learned prior to evaluating its performance. Methods like weight-sharing \cite{enas,smash,onceforall} allow sampled subnets to inherit weights among themselves or from a supernet, avoiding the time-consuming  process (typically requiring hours) of learning weights through SGD. However, standalone weight-sharing still requires inference on validation data (typically requiring minutes) to assess performance. Therefore, simply having to evaluate the subnets 
can still render the overall process computationally prohibitive for methods \cite{nasnet,amoebanet,mnasnet} that sample thousands of architectures during search.
\begin{figure}[t]
    \centering
    \begin{subfigure}{0.45\textwidth}
    \centering
    \includegraphics[width=\textwidth]{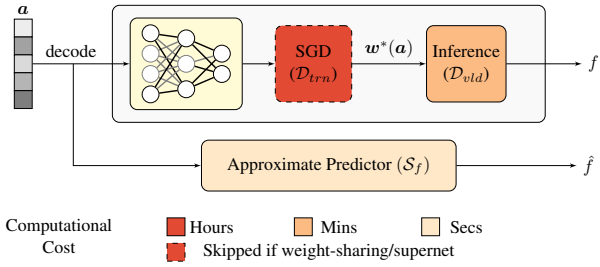}
    \end{subfigure}
\caption{\textbf{Top Path:} A typical process of evaluating an architecture in NAS algorithms. \textbf{Bottom Path:} Accuracy predictor aims to bypass the time-consuming components for evaluating a network's performance by directly regressing its accuracy $f$ from $\bm{a}$ (architecture in the encoded space).\label{fig:acc_pred}}
\vspace{-0.3cm}
\end{figure}

To mitigate the computational burden of fully evaluating the subnets, we adopt a surrogate accuracy predictor that regresses the performance of a sampled subnet without performing training or inference. By learning a functional relation between the integer-strings (subnets in the encoded space) and the corresponding performance, this approach decouples the evaluation of an architecture from data-processing (including both SGD and inference). Consequently, the evaluation time reduces from hours/minutes to seconds. We illustrate this concept in Fig.~\ref{fig:acc_pred}. The effectiveness of this idea, however, is critically dependent on the quality of the surrogate model. Below we identify three desired properties of such a model:

\begin{enumerate}
    \item Reliable prediction: high rank-order correlation\footnote{Low mean square error is also desirable, but not necessary since the selection of architectures in the subsequent evolutionary search compares relative performance between architectures.} between predicted and true performance.
    \item Consistent prediction: the quality of the prediction should be consistent across different datasets.
    \item Sample efficiency: minimizing the number of training examples necessary to construct an accurate predictor model, since each training sample requires costly training and evaluation of a subnet.
\end{enumerate}

Current approaches~\cite{PNAS,chamnet,onceforall} that use surrogate based accuracy predictors, however, do not satisfy property (1) and (3) simultaneously. For instance, PNAS~\cite{PNAS} uses 1,160 subnets to build the surrogate but only achieves a rank-order correlation of 0.476. Similarly, OnceForAll~\cite{onceforall} uses 16,000 subnets to build the surrogate. The poor sample complexity and rank-order correlation of these approaches, is due to the offline learning of the surrogate model. Instead of focusing on models that are at the trade-off front of the objectives, these surrogate models are built for the entire search space. Consequently, these methods require a significantly larger and more complex surrogate model.
\begin{algorithm}[t]
\SetAlgoLined
\SetKwInOut{Input}{Input}
\SetKwInOut{Output}{Output}
\SetKwFor{For}{for}{do}{end for}
\footnotesize
\Input{Training data $X$, training targets $Y$, ensemble size $K$}
    $k$ $\leftarrow$ 0 \textcolor{gray}{// initialize an counter.}\\
    \emph{pool} $\leftarrow \emptyset$ \textcolor{gray}{// initialize a pool to store all models.}\\
    \While{$k < K$}{
        ($\tilde{X}, \tilde{Y}) \leftarrow$ randomly create a subset of the training data. \\
        \emph{idx} $\leftarrow$ randomly pick a subset of the features in training data. \\
        \emph{rbf} $\leftarrow$ fit a RBF model from $\tilde{X}[:, idx]$ and $\tilde{Y}$.\\
        \emph{pool} $\leftarrow$ \emph{pool} $\cup$ (\emph{rbf}, \emph{idx}) \textcolor{gray}{// append the fitted model to the pool.}\\
        $k$ $\leftarrow$ $k + 1$
    }
\textbf{Return} a \emph{pool} of $K$ RBF models.
\caption{Accuracy Predictor (RBF Ensemble)\label{algo:acc_predictor}}
\end{algorithm}
\begin{figure}[t]
    \centering
    \begin{subfigure}{0.45\textwidth}
    \centering
    \includegraphics[width=0.9\textwidth]{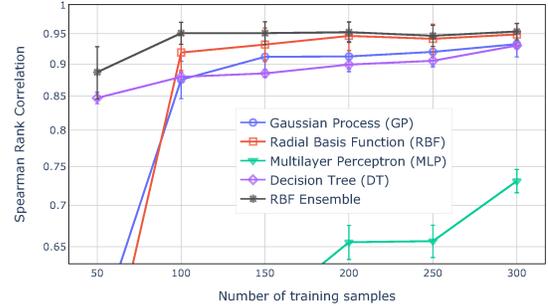}
    \end{subfigure}
\caption{Accuracy predictor performance as a function of training samples. For each model, we show the mean and standard deviation of the Spearman rank correlation on 11 datasets (Table~\ref{tab:dataset}). The size of RBF ensemble is 500.\label{fig:accuracy_predictor1}}
\end{figure}

We overcome the aforementioned limitation by restricting the surrogate model to the search space that constitutes the current objective trade-off. Such a solution significantly reduces the sample complexity of the surrogate and increases the reliability of its predictions. We adopt four low-complexity predictors, namely, Gaussian Process (GP) \cite{chamnet}, Radial Basis Function (RBF) \cite{baker2017accelerating}, Multilayer Perceptron (MLP) \cite{PNAS}, and Decision Tree (DT) \cite{ae-cnn+e2epp}. Empirically, we observe that RBFs are consistently better than the other three models if the \# of training samples is more than 100. To further improve RBF's performance, especially under a high sample efficiency regime, we construct an ensemble of RBF models. As outlined in Algorithm~\ref{algo:acc_predictor}, each RBF model is constructed with a subset of samples and features randomly selected from the training instances. The correlation between predicted accuracy and true accuracy from an ensemble of 500 RBF models outperforms all other models across all regimes. Fig.~\ref{fig:accuracy_predictor1} compares the performance of the different surrogate models we considered. Practically, we observed that the RBF ensemble can be learned under a minute.

\subsection{Many-Objective Evolutionary Search\label{sec:many-obj}}
Given the accuracy predictor, we employ a customized evolutionary algorithm (EA) to search for optimal architectures that offer the best trade-off between many objectives. The EA is an iterative process in which initial architectures, selected from the archive of previously explored architectures, are gradually improved as a group, referred to as a \emph{population}. In every generation (iteration), a group of \emph{offspring} (i.e., new architectures) are created by applying variations through crossover and mutation (described below) operations on the most promising architectures, also known as \emph{parents}, found so far in the population. Every member of the population, i.e., both parents and offspring, competes for survival and reproduction (becoming a parent) in each generation. See Fig.~\ref{fig:overview} (bottom right shaded in green) for a pictorial overview, and Algorithm~\ref{algo:evolution} for the pseudocode. 

\begin{algorithm}[t]
\SetAlgoLined
\SetKwInOut{Input}{Input}
\SetKwInOut{Output}{Output}
\SetKwFor{For}{for}{do}{end for}
\footnotesize
\Input{Accuracy predictor $\mathcal{S}_f$, additional objectives $\tilde{f}$, archive of archs $\mathcal{A}$, max. \# of generations $G$, population size $K$, crossover probability $p_{c}$, mutation probability $p_{m}$.}
    $g$ $\leftarrow$ 0 \textcolor{gray}{// initialize an generation counter.}\\
    $f \leftarrow \mathcal{S}_f(\mathcal{A})$ \textcolor{gray}{// compute accuracy of all archs in archive.}\\
    $P \leftarrow$ \emph{Selection}($\mathcal{A}, f, \tilde{f}(\mathcal{A}), K$) \textcolor{gray}{// initialize the parent population with top-$K$ ranked archs from $\mathcal{A}$.}\\
    \While{$g < G$}{
        \textcolor{gray}{// choose parents through tournament selection for mating.}\\
        $P \leftarrow$ \emph{Binary Tournament Selection}($P$) \\
        \textcolor{gray}{// create offspring population by crossover between parents.}\\
        $Q \leftarrow$ \emph{Crossover}($P, p_{c}$) \\
        \textcolor{gray}{// induce randomness to offspring population through mutation.}\\
        $Q \leftarrow$ \emph{Mutation}($Q, p_{m}$) \\
        $R \leftarrow P \cup Q$ \textcolor{gray}{// merge parent and offspring population.}\\
        \textcolor{gray}{// survive the top-$K$ archs to next generation.}\\
        $P \leftarrow$ \emph{Selection}($R, \mathcal{S}_f(R), \tilde{f}(R), K$) \\
        $g$ $\leftarrow$ $g + 1$
    }
\textbf{Return} parent population $P$.
\caption{Evolutionary Search\label{algo:evolution}}
\end{algorithm}

\begin{figure}[t]
    \centering
    \begin{subfigure}{0.4\textwidth}
    \centering
    \includegraphics[width=\textwidth]{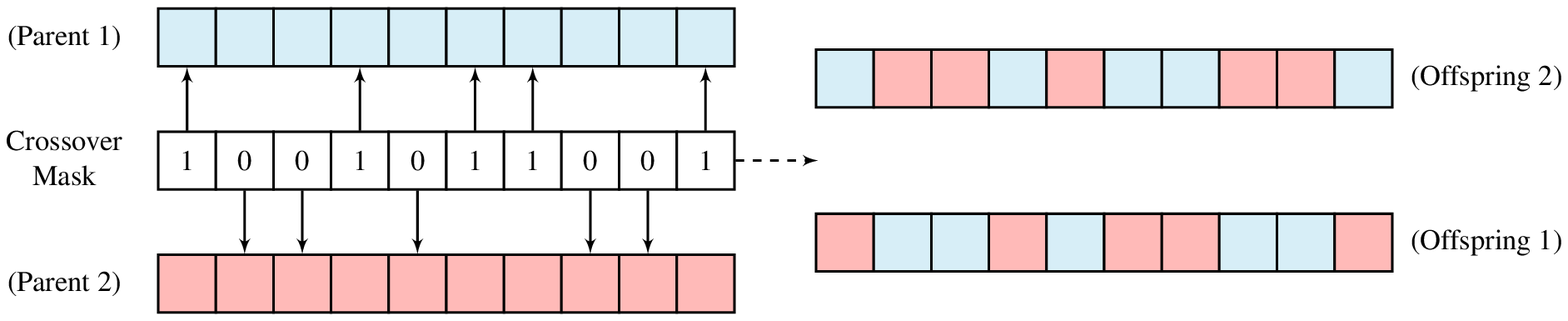}
    \caption{\label{fig:crossover}}
    \end{subfigure}
    \centering
    \begin{subfigure}{0.48\textwidth}
    \centering
    \includegraphics[width=\textwidth]{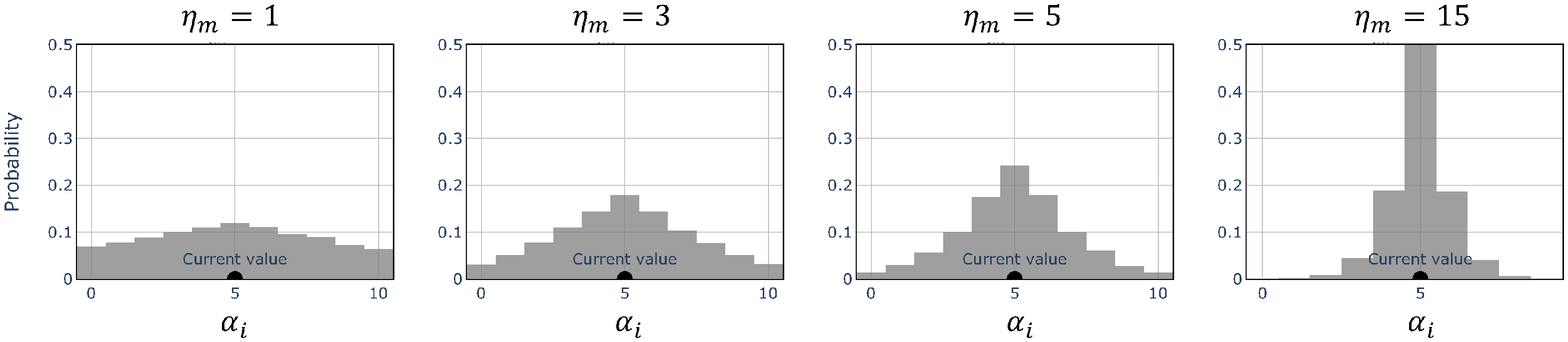}
    \caption{\label{fig:mutation}}
    \end{subfigure}
    \caption{(a) \textbf{Crossover Operator}: new offspring architectures are created by recombining integers from two parent architectures. The probability of choosing from either one of the parents is equal. (b) \textbf{Mutation Operator}: histograms showing the probabilities of mutated values with current value at 5 under different hyperparameter $\eta_m$ settings.}
    \vspace{-0.3cm}
\end{figure}

\textbf{Crossover} exchanges information between two or more population members to create two or more new members. Designing an effective crossover between non-standard solution representations can be difficult and has been largely ignored by existing EA-based NAS algorithms \cite{real2017large,liu2018hierarchical,amoebanet}. Here we adopt a customized, homogeneous crossover that uniformly picks integers from parent architectures to create offspring architectures. This crossover operator offers two properties: (1) it preserves common integers shared between parents; and (2) it is free of additional hyperparameters. Fig.~\ref{fig:crossover} visualizes our implementation of the crossover operation. We generate two offspring architectures with each crossover, and an offspring population of the same size as the parent population is generated in each generation.

\textbf{Mutation} is a \emph{local} operator that perturbs a solution to produce a new solution in its vicinity. In this work, we use a discretized version of the polynomial mutation (PM) operator \cite{deb1995simulated} and apply it to every solution created by the crossover operator. For a given  architecture $\bm{a}$, PM is carried out integer-wise with probability $p_m$, and the mutated $i^{th}$ integer, $a_i$, of the mutated offspring is:
\begin{equation}
  \small
  a_i^{\prime} =
    \begin{cases}
      a_i + ((2u)^{1/(1 + \eta_m)} - 1)(a_i - a_i^{(L)}), & \text{for } u \leq 0.5,\\
      a_i + (1 - \big(2(1 - u)\big)^{1/(1 + \eta_m)})(a_i^{(U)} - a_i), & \text{for } u > 0.5\\
    \end{cases}
  \normalsize
\end{equation}
where $u$ is a uniform random number in the interval $[0, 1]$. $a_i^{(L)}$ and $a_i^{(U)}$ are the lower and upper bounds of $a_i$, respectively. Each mutated value in an offspring is rounded to the nearest integer. The PM operator inherits the \emph{parent-centric} convention, in which the offspring are intentionally created around the parents. The centricity is controlled via an index hyperparameter $\eta_m$. In particular, high-values of $\eta_m$ tend to create mutated offspring around the parent, and low-values encourage mutated offspring to be further away from the parent architecture. See Fig.~\ref{fig:mutation} for a visualization of the effect of $\eta_m$. It is the worth noting that the PM operator was originally proposed for continuous optimization where distances between variable values are naturally defined. In contrast, in context of our encoding, our variables are categorical in nature, indicating a particular layer hyperparameter. So we sort the searched subnets in ascending order of \#MAdds, such that $\eta_m$ now controls the difference in \#MAdds between the parent and the mutated offspring.

We apply PM to every member in the offspring population (created from crossover). We then merge the mutated offspring population with the parent population and select the top half using many-objective selection operator described in Algorithm~\ref{algo:nsga3}. This procedure creates the parent population for the next generation. We repeat this overall process for a pre-specified number of generations and output the parent population at the conclusion of the evolution.

\subsection{Many-Objective Selection\label{sec:selection}}
In addition to high predictive accuracy, real-world applications demand NAS algorithms to simultaneously balance a few other conflicting objectives that are specific to the deployment scenarios. For instance, mobile or embedded devices often have restrictions in terms of model size, multiply-adds, latency, power consumption, and memory footprint. With no prior assumption on the correlation among these objectives, a scalable (to the number of objectives) selection is required to drive the search towards the high dimensional Pareto front. In this work, we adopt the reference point guided selection originally proposed in NSGA-III \cite{nsga3}, which has been shown to be effective in handling problems with as many as 15 objectives. In the remainder of this section, we provide an overview of NSGA-III procedure and refer readers to the original publication for more details.

\begin{algorithm}[t]
\SetAlgoLined
\SetKwInOut{Input}{Input}
\SetKwInOut{Output}{Output}
\SetKwFor{For}{for}{do}{end for}
\footnotesize
\Input{A set of archs $R$, their objectives $F$, number of archs to select $N$, reference directions $Z$.}
    \textcolor{gray}{// put archs into different fronts (rank levels) based on domination.} \\
    $(F_1, F_2, \ldots) \leftarrow$ \emph{non}\_\emph{dominated}\_\emph{sort}($F$) \\
    $S \leftarrow \emptyset$, $i \leftarrow 1$ \\
    \lWhile{$\vert S \vert + \vert F_i \vert < N$}{$S \leftarrow S \cup F_i$; $i \leftarrow i + 1$}
    $F_{L} \leftarrow F_i$  \textcolor{gray}{// next front is the split front where we cannot accommodate all archs associated with it.}\\
    \lIf{$\vert S \vert + \vert F_L \vert = N$}{$S \leftarrow S \cup F_L$}
    \Else{
        $(\tilde{S}, \tilde{F_L}) \leftarrow$ \emph{Normalize}($S, F_L$) \textcolor{gray}{// normalize the objectives based the ideal and nadir points derived from $R$.}\\
        $d \leftarrow$ compute orthogonal dist to $Z_i$ for each $i$\\
        $\rho \leftarrow$ count \#associated solutions for $Z_i$ based on $d$ for each $i$.\\
        \textcolor{gray}{// remaining archs from $F_L$ to fill up $S$.}\\
        $S \leftarrow S$ $\cup$ \emph{Niching}($\tilde{F_L}$, $N - \vert S \vert$, $\rho$, $d$)\\
    }
\textbf{Return} $S$.
\caption{Reference Point Based Selection\label{algo:nsga3}}
\end{algorithm}

\emph{Domination} is a widely-used partial ordering concept for comparing two objective vectors. For a generic many-objective optimization problem: $\min_{\bm{\bm{a}}} \left\{f_1(\bm{a}), \dots, f_m(\bm{a})\right\}$, where $f_i(\cdot)$ are the objectives (say, loss functions) to be optimized and $\bm{a}$ is the representation of a neural network architecture. For two given solutions $\bm{a}_1$ and $\bm{a}_2$, solution $\bm{a}_1$ is said to dominate $\bm{a}_2$ (i.e., $\bm{a}_1 \preceq \bm{a}_2$) if following conditions are satisfied:
\begin{enumerate}
    \item $\bm{a}_1$ is no worse than $\bm{a}_2$ for all objectives $(f_i(\bm{a}_1) \leq f_i(\bm{a}_2)$, $\forall i \in \{0, \dots, m\})$, and
    \item $\bm{a}_1$ is strictly better than $\bm{a}_2$ in at least one objective $\exists$ $i \in \{0, \dots, m\} \mid f_i(\bm{a}_1) < f_i(\bm{a}_2))$.
\end{enumerate}
\noindent A solution $\bm{a}_i$ is said to be non-dominated if these conditions hold against all the other solutions $\bm{a}_j$ (with $j\neq i$) in the entire search space of $\bm{a}$.

\begin{figure}[t]
    \centering
    \begin{subfigure}{0.24\textwidth}
    \centering
    \includegraphics[width=0.98\textwidth]{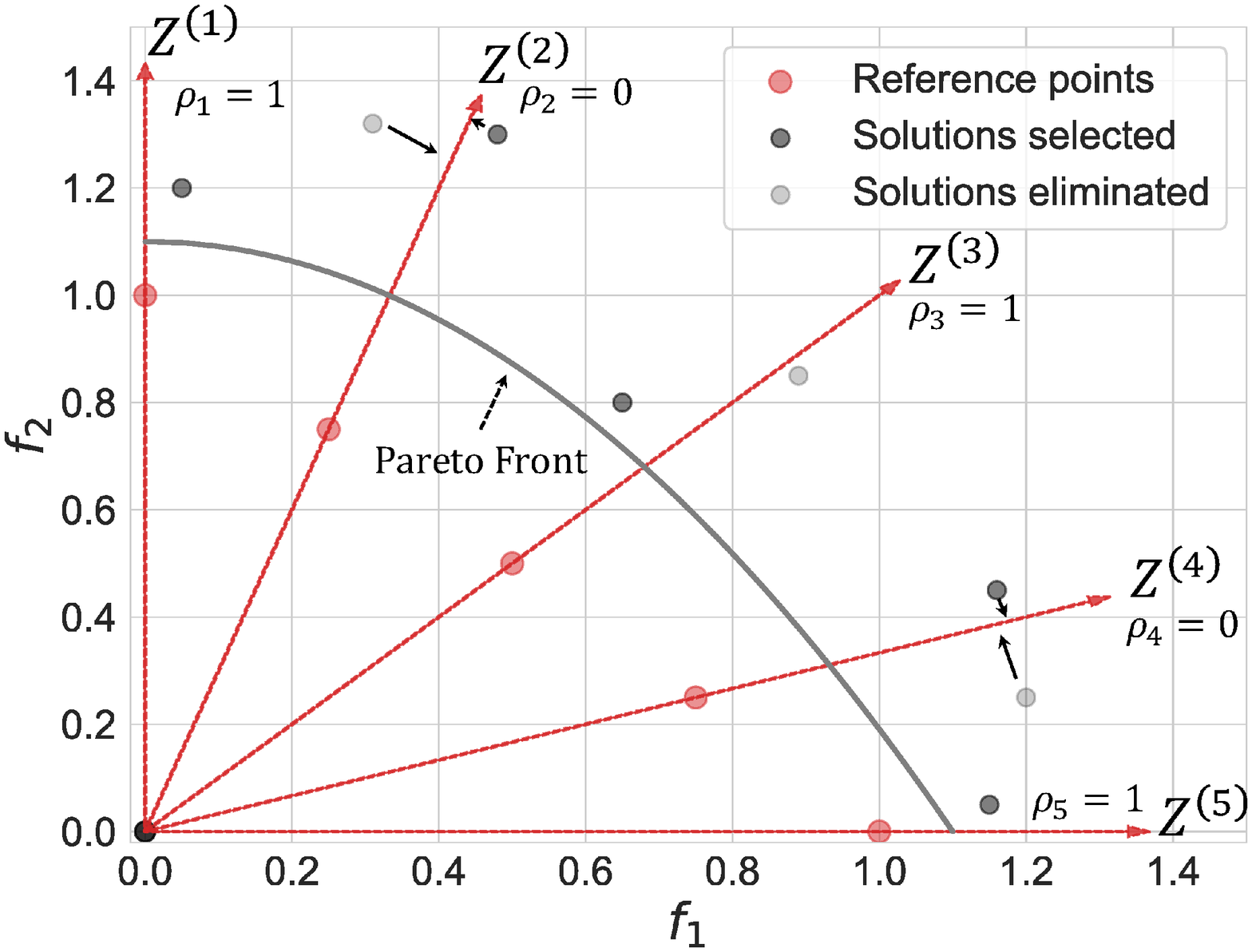}
    \caption{}
    \end{subfigure}
    \centering
    \begin{subfigure}{0.24\textwidth}
    \centering
    \includegraphics[width=0.98\textwidth, trim=1.5cm 0 0 1cm, clip=true]{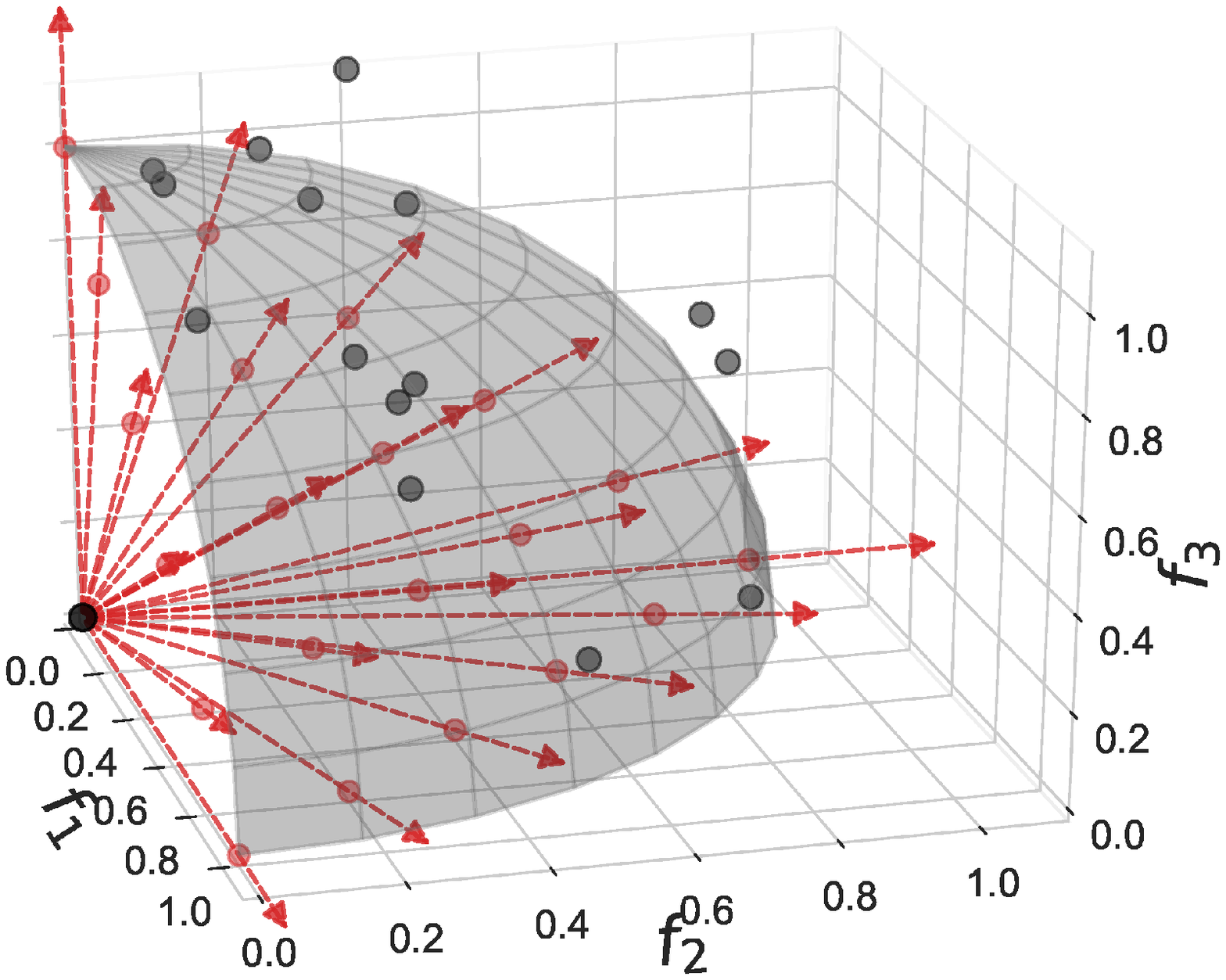}
    \caption{}
    \end{subfigure}
\caption{(a) An example (assuming minimization of all objectives) of the selection process in Algo~\ref{algo:nsga3}: We first create reference directions $Z$ by joining reference points with the ideal solution (origin). Then through \emph{non}\_\emph{dominated}\_\emph{sort}, three non-dominated solutions are identified, associated with reference directions $Z^{(1)}$, $Z^{(3)}$ and $Z^{(5)}$. We then select the remaining solutions by the orthogonal distances to the reference directions with no associated solutions---i.e. $Z^{(2)}$ and $Z^{(4)}$. This selection is scalable to larger \# of objectives. A tri-objective example is shown in (b). \label{fig:selection}}
\vspace{-0.3cm}
\end{figure}

With the above definition, we can sort solutions to different ranks of domination, where solutions in the same rank are non-dominated to each other, and there exists at least one solution in lower rank that dominates any solution in the higher rank. Thus, a lower non-dominated ranked set is lexicographically better than a higher ranked set. This process is referred as \emph{non}\_\emph{dominated}\_\emph{sort}, and it is the first step in the selection process. During the many-objective selection process,  the lower ranked sets are chosen one at a time until no more sets can be included to maintain the population size. The final accepted set may have to be \emph{split} to choose only a part. For this purpose, we choose the most diverse subset based on a diversity-maintaining mechanism. We first create a set of reference directions from a set of uniformly distributed (in ($m-1$)-dimensional space) reference points in the unit simplex by using Das-and-Dennis method \cite{das-n-dennis}. Then we associate each solution to a reference direction based on orthogonal distance of the solution from the direction. Then, for every reference direction, we choose the closest associated solution in a systematic manner by adaptively computing a niche count $\rho$ so that every reference direction gets an equal opportunity to choose a representative closest solution in the selected population. The domination and diversity-preserving procedures are easily scalable to any number of objectives and importantly are free from any user-defined hyperparameter. See Algorithm~\ref{algo:nsga3} for the pseudocode and Fig.~\ref{fig:selection} for a graphical illustration. A more elaborated discussion on the necessity of the reference point based selection is provided in Section~\ref{sec:many-continued}.

\subsection{Supernet Adaptation}
Instead of training every architectures sampled during search from scratch, NAS with weight sharing \cite{enas,darts} inherits weights from previously-trained networks or from a supernet. Directly inheriting the weights obviates the need to optimize the weights from scratch and speeds up the search from thousands of GPU days to only a few. In this work, we focus on the supernet approach \cite{one-shot,onceforall}. It involves first training a large network model (in which searchable architectures become subnets) prior to the search. Then the performance of the subnets, evaluated with the inherited weights, is used to guide the selection of architectures during search. The key to the success of this approach is that the performance of the subnets with the inherited weights be highly correlated with the performance of the same subnet when thoroughly trained from scratch. Satisfying this desideratum necessitates that the supernet weights be learned in such a way that \emph{all} subnets are optimized \emph{simultaneously}.

\begin{algorithm}[t]
\SetAlgoLined
\SetKwInOut{Input}{Input}
\SetKwInOut{Output}{Output}
\SetKwFor{For}{for}{do}{end for}
\footnotesize
\Input{Supernet $\mathcal{S}_w$, archive of archs $\mathcal{A}$, training data $\mathcal{D}_{trn}$, number of epochs $E$.}
    $e$ $\leftarrow$ 0 \textcolor{gray}{// initialize an epoch counter.}\\
    \emph{Distr} $\leftarrow$ construct the distribution from $\mathcal{A}$ following Eq.~(\ref{eq:dist}).\\
    \While{$e < E$}{
        \For{each batch in $\mathcal{D}_{trn}$}{
            \emph{subnet} $\leftarrow$ sample from \emph{Distr}. \\
            $w \leftarrow$ set forward path of $\mathcal{S}_w$ according to \emph{subnet}. \\
            $\mathcal{L} \leftarrow$ compute cross-entropy loss on data \emph{batch}.\\
            $\nabla w \leftarrow$ compute the gradient by $\partial \mathcal{L} / \partial w$ \\
            $\mathcal{S}_w \leftarrow$ one step of SGD. \\
        }
        $e$ $\leftarrow$ $e + 1$
    }
\textbf{Return} supernet $\mathcal{S}_w$.
\caption{Adapt Supernet\label{algo:supernet}}
\end{algorithm}

Existing methods \cite{guo2019single,chu2019fairnas} attempt to achieve the above goal by imposing \emph{fairness} in training the supernet, where the probabilities of training any particular subnet for each batch of data is uniform in expectation. However, we argue that simultaneously training all the subnets in the search space is practically not feasible and, more importantly, not necessary. Firstly, it is evident from existing NAS approaches \cite{proxylessnas,fbnet} that different objectives (\#Params, \#MAdds, latency on different hardware, etc.) require different architectures in order to be efficient. In other words, not all subnets are equally important for the task at hand. Secondly, only a tiny fraction\footnote{For example, AmoebaNet \cite{amoebanet} samples a large number of 27K architectures which is still only about \num{E-13}\% of its search space.} of the search space can practically be explored by a NAS algorithms.

Based on the aforementioned observations, we propose a simple yet effective supernet training routine that only focuses on training the subnets recommended by the evolutionary search algorithm in Section \ref{sec:selection}. Specifically, we seek to exploit the knowledge gained from the search process so far. Recall that our algorithm uses an archive to keep track of the promising architectures explored so far. For each value in our architecture encoding, we construct a categorical distribution from architectures in the archive, where the probability for $i^{th}$ integer taking on the $j$ value is computed as:
\begin{equation}
p(X_i=j) = \frac{\#\mbox{ of architectures with option $j$ at $i^{th}$ integer}}{\mbox{total \# of architectures in the archive}}
\label{eq:dist}
\end{equation}
In each training step (batch of data), we sample an integer-string from the above distribution\footnote{A visualization of such distributions is shown in \ref{fig:rosenbrock_ours}.}. We then activate the sub parts of the supernet corresponding to the architecture decoded from the integer-string. Only weights corresponding to the activated sub parts in the supernet will be updated in each step. See Algorithm~\ref{algo:supernet} for pseudocode. A more in-depth discussion connecting our proposed approach to the existing supernet-based NAS approaches is provided in Section~\ref{sec:one-shot}. 

\section{Experimental Evaluation\label{sec:experiments}}

In this section, we present experimental results to evaluate the efficacy of \emph{Neural Architecture Transfer} on multiple image classification tasks. In addition, we also investigate the scalability of our approach to more than two objectives. For all the experiments in this section, we use the same set of hyperparmaters (see Table \ref{tab:hyperparameters}) for the different components of \ourmethod{}. These choices were guided by the ablation studies described in Section~\ref{sec:ablation}.

\begin{table}[t]
\centering
\caption{Hyperparameter Settings\label{tab:hyperparameters}}
\resizebox{0.42\textwidth}{!}{%
\scriptsize{
\begin{tabular}{@{\hspace{2mm}}l|l|c@{\hspace{2mm}}}
\toprule
Category & Parameter & Setting \\ \midrule
\multirow{2}{*}{Global} & Archive size & 300 \\
 & Number of iterations & 30 \\ \midrule
\multirow{2}{*}{Accuracy predictor} & Train size & 100 \\
 & Ensemble size & 500 \\ \midrule
\multirow{5}{*}{Evolutionary search} & Population size & 100 \\
 & Number of generations per iteration & 100 \\
 & Crossover probability & 0.9 \\
 & Mutation probability & 0.1 \\
 & Mutation index $\eta_m$ & 1.0 \\ \midrule
Supernet & Number of epochs per iteration & 5 \\ \bottomrule
\end{tabular}%
}}
\vspace{-0.2cm}
\end{table}
\begin{table}[t]
\centering
\caption{Benchmark Datasets for Evaluation}
\label{tab:dataset}
\resizebox{0.45\textwidth}{!}{%
\scriptsize{
\begin{tabular}{lccccc}
\toprule
Dataset & Type && Train Size & Test Size & \#Classes \\ \midrule
ImageNet \cite{imagenet} & \multirow{5}{*}{multi-class} && 1,281,167 & 50,000 & 1,000 \\
CINIC-10 \cite{cinic10} &&& 180,000 & 9,000 & 10 \\
CIFAR-10 \cite{cifar} &&& 50,000 & 10,000 & 10 \\
CIFAR-100 \cite{cifar} &&& 50,000 & 10,000 & 10 \\
STL-10 \cite{stl-10} &&& 5,000 & 8,000 & 10 \\
\midrule
Food-101 \cite{food-101} & \multirow{6}{*}{fine-grained} && 75,750 & 25,250 & 101 \\
Stanford Cars \cite{stanford_cars} &&& 8,144 & 8,041 & 196 \\
FGVC Aircraft \cite{aircraft} &&& 6,667 & 3,333 & 100 \\
DTD \cite{dtd} &&& 3,760 & 1,880 & 47 \\
Oxford-IIIT Pets \cite{pets} &&& 3,680 & 3,369 & 37 \\
Oxford Flowers102 \cite{flowers102} &&& 2,040 & 6,149 & 102 \\ \bottomrule
\end{tabular}%
}}
\vspace{-0.2cm}
\end{table}

\subsection{Datasets}
We consider eleven image classification datasets for evaluation with sample size varying from 2,040 to 180,000 images (20 to 18,000 images per class; Table~\ref{tab:dataset}). These datasets span a wide variety of image classification tasks, including superordinate-level recognition (ImageNet~\cite{imagenet}, CIFAR-10 \cite{cifar}, CIFAR-100 \cite{cifar}, CINIC-10 \cite{cinic10}, STL-10 \cite{stl-10}); fine-grained recognition (Food-101 \cite{food-101}, Stanford Cars \cite{stanford_cars}, FGVC Aircraft \cite{aircraft}, Oxford-IIIT Pets \cite{pets}, Oxford Flowers102 \cite{flowers102}); and texture classification (DTD \cite{dtd}). We use the ImageNet dataset for training the supernet, and use the other ten datasets for architecture transfer.

\subsection{Supernet Preparation\label{sec:implementation}}
Our supernet is constructed by setting the architecture encoding at the maximum value, i.e. four layers in each stage and every layer uses expand ratio of six and kernel size of seven. Adapting subnets of a supernet with randomly initialized weights leads to training instability and large variance in its performance. Therefore, we warm-up the supernet weights on ImageNet following the \emph{progressive shrinking} algorithm \cite{onceforall}, where the supernet is first trained at full-scale, with subnets corresponding to different options (expand ratio, kernel size, \# of layers) being gradually activated during the training process. This procedure, which takes about 6 days on a server with eight V100 GPUs, is optimized with only the cross-entropy loss i.e., a single objective. We note that supernet preparation expense is a one-time cost that amortizes over any subsequent transfer to different datasets and objective combinations we show in the following subsections.

\subsection{ImageNet Classification\label{sec:imagenet}}
Before we evaluate our approach for architecture transfer to other datasets, we first validate its effectiveness on the ImageNet-1K dataset. This experiment evaluates the effectiveness of \ourmethod{} in adapting and searching for architectures that span trade-off between two objectives. For this experiment, we consider accuracy and \#MAdds as the two objective of interest. {\color{black} We randomly sample 50,000 images from the original ImageNet training set as the validation set to guide the architecture search.} We run \ourmethod{} for 30 iterations, and from the final archive of architectures, we select four models ranging from 200M MAdds to 600M MAdds (for high-end mobile devices). Following \cite{onceforall}, we fine-tune\footnote{Section~\ref{sec:abl_supernet} studies the impact of this fine-tuning step.} each model to further boost the performance. Our fine-tune training largely follows \cite{mnasnet}: RMSProp optimizer with decay 0.9 and momentum 0.9; batch normalization momentum 0.99; weight decay 1e-5. We use a batch size of 512 and an initial learning rate of 0.012 that gradually reduces to zero following the cosine annealing schedule. Our regularization settings are similar as in \cite{efficientnet}: we use augmentation policy \cite{cubuk2019randaugment}, drop connect ratio 0.2, and dropout ratio 0.2.

\begin{figure}[t]
    \centering
    \begin{subfigure}{0.48\textwidth}
    \centering
    \hspace*{1.5mm}
    \includestandalone[width=0.78\textwidth]{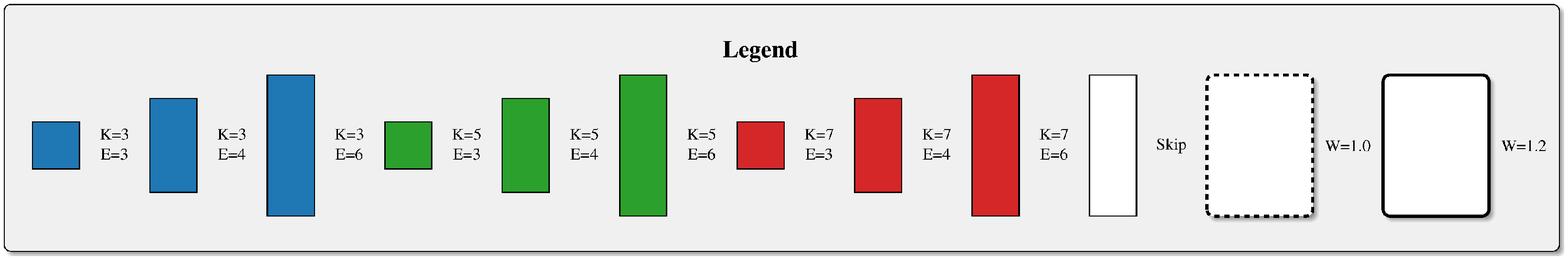} \\
    \includestandalone[width=0.95\textwidth]{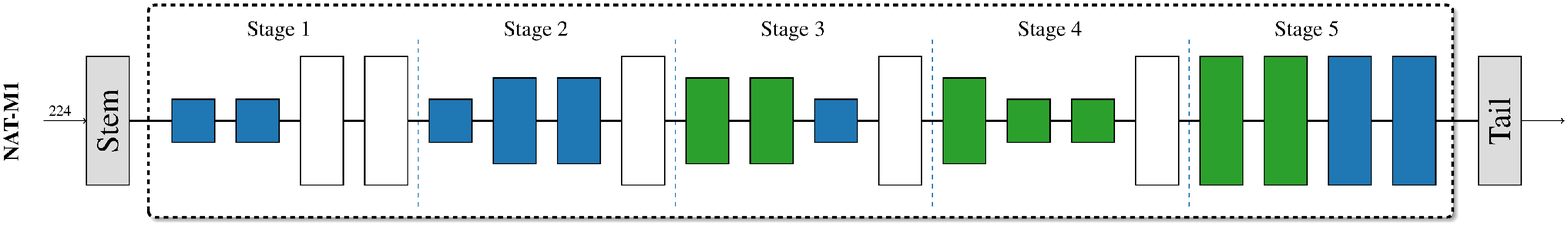} \\
    \includestandalone[width=0.95\textwidth]{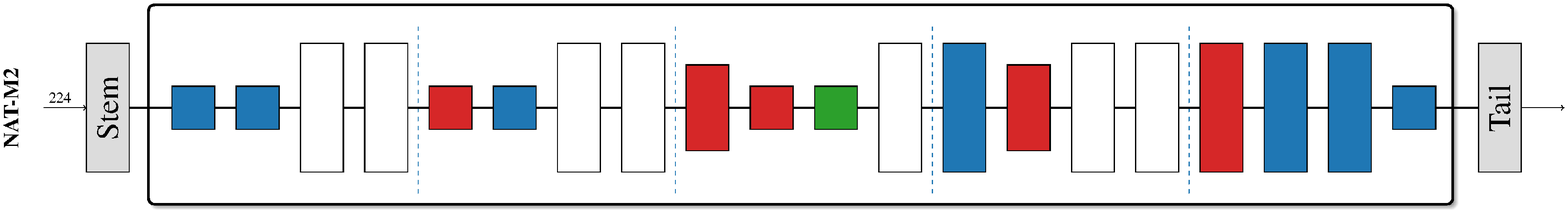} \\
    \includestandalone[width=0.95\textwidth]{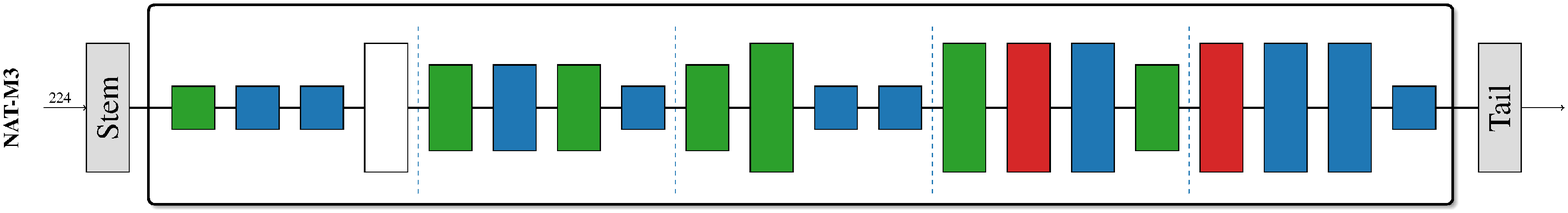} \\
    \includestandalone[width=0.95\textwidth]{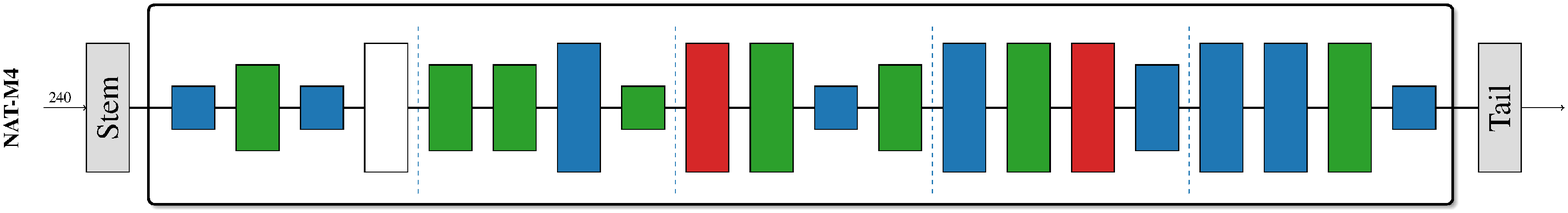} \\
    \end{subfigure}
\caption{ImageNet Architectures from Trade-Off Front.\label{fig:imagenet_archs}}
\vspace{-0.3cm}
\end{figure}

\begin{figure}[t]
    \centering
    \begin{subfigure}{0.45\textwidth}
    \centering
    \includegraphics[width=0.95\textwidth]{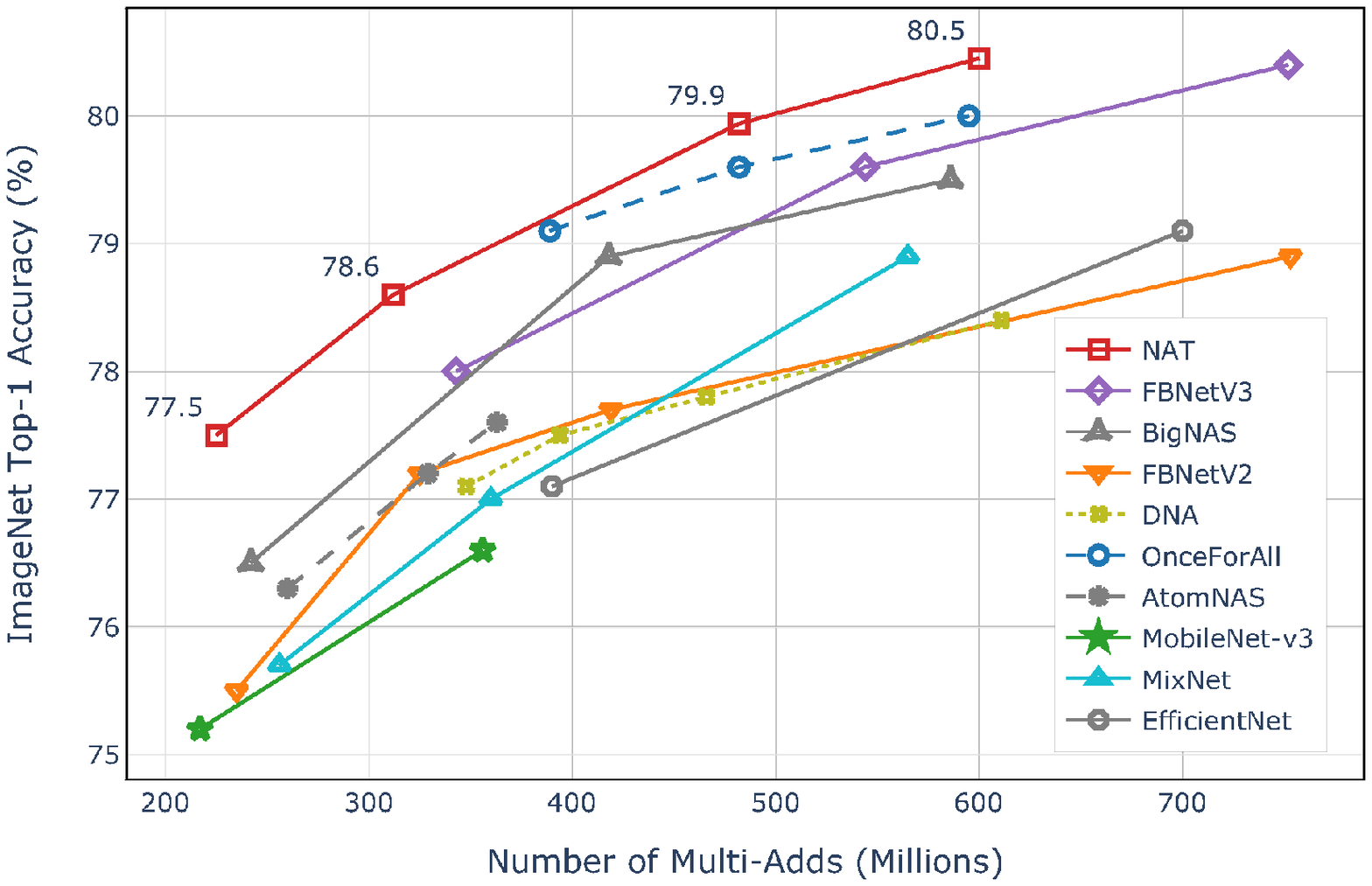}
    \end{subfigure}
\caption{\textbf{MAdds vs. ImageNet Accuracy}. \ourmodel{}s outperform other models in both objectives. In particular, \ourmethod{}-M4 achieves a new state-of-the-art top-1 accuracy of 80.5\% under mobile setting ($\leq$ 600M MAdds). \ourmethod{}-M1 improves MobileNetV3 top-1 accuracy by 2.3\% with similar \#MAdds. \label{fig:imagenet_mobilnet}}
\vspace{-0.3cm}
\end{figure}

Table~\ref{tab:imagenet} shows the performance of \ourmethod{} models obtained through bi-objective optimization of maximizing accuracy and minimizing \#MAdds. Our models, referred to as \ourmethod{}-M\{1,2,3,4\}, are in ascending order of \#MAdds (Fig.~\ref{fig:imagenet_archs}). Fig.~\ref{fig:imagenet_mobilnet} shows the full \#MAdds-accuracy trade-off curve comparison between \ourmethod{} and existing NAS methods. 

Results indicate that \ourmodel{}s completely dominate (i.e. better in both \#MAdds and accuracy) all existing designs, both manual and from other NAS algorithms, under mobile settings ($\leq$ 600M MAdds). Compared to manually. designed networks, \ourmethod{} is noticeably more efficient. \ourmethod{}-M1 is \textbf{2.3\%} and \textbf{1.5\% more accurate} than MobileNetV3 \cite{mobilenetv3} and FBNetV2-F4 \cite{fbnetv2} respectively, while being equivalent in efficiency (i.e. \#MAdds, CPU and GPU latency). Furthermore, \ourmodel{}s are consistently \textbf{6\% more accurate} than MobileNetV2 \cite{mobilenetv2} scaled by width multiplier from 200M to 600M \#MAdds. Our largest model, \ourmethod{}-M4, achieves a new state-of-the-art ImageNet top-1 accuracy of 80.5\% under mobile settings ($\leq$ 600M \#MAdds). Interestingly, even though this experiment did not explicitly optimize for CPU or GPU latency, \ourmodel{}s are faster than those (MobileNet-V3 \cite{mobilenetv3}, MNasNet \cite{mnasnet}) that explicitly do optimize for latency.

\begin{table*}[!t]
\caption{\textbf{ImageNet-1K Classification~\cite{imagenet}:} \ourmodel{}s comparison with manual and automated design of efficient convolutional neural networks. Models are grouped into sections for better visualization. Our results are underlined and the best result in each section is in bold. CPU latency (batchsize=1) is measured on Intel i7-8700K and GPU latency (batchsize=64) is measured on 1080Ti. ``WS'' stands for weight sharing. All methods are under single crop and single model condition, without any additional data. 
\label{tab:imagenet}}
\centering
\resizebox{0.9\textwidth}{!}{%
\scriptsize{
\begin{tabular}{@{\hspace{2mm}}l|l|cccc|cc@{\hspace{2mm}}}
\specialrule{1.5pt}{1pt}{1pt}
Model & Method & \#Params & \#Multi-Adds\hspace{-2mm} & CPU Lat (ms)\hspace{-2mm} & GPU Lat (ms) & Top-1 Acc (\%) & Top-5 Acc (\%) \\
\specialrule{1.5pt}{1pt}{1pt}
\textbf{\ourmethod{}-M1} & WS+EA & \underline{6.0M} & \underline{225M} & \underline{9.1} & \underline{30} & \textbf{\underline{77.5}} & \textbf{\underline{93.5}} \\
MobileNetV2~\cite{mobilenetv2} & manual & 3.5M & 300M & \textbf{8.3} & \textbf{23} & 72.0 & 91.0 \\
\color{black}SPOS NAS~\cite{guo2019single} & \color{black}WS+EA & \color{black}\textbf{3.4M} & \color{black}328M & - & - & \color{black}74.7 & \color{black}92.0 \\
ProxylessNAS~\cite{proxylessnas} & RL/gradient & 4.0M & 465M & 8.5 & 27 & 75.1 & 92.5 \\
MnasNet-A1~\cite{mnasnet} & RL & 3.9M & 312M & 9.3 & 31 & 75.2 & 92.5 \\
MobileNetV3~\cite{mobilenetv3} & RL/NetAdapt & 5.4M & \textbf{219M} & 10.6 & 33 & 75.2 & - \\
MUXNet-m~\cite{muxconv} & EA & \textbf{3.4M} & \textbf{218M} & 14.7 & 42 & 75.3 & 92.5 \\
FBNetV2-F4~\cite{fbnetv2} & gradient & 7.0M & 238M & 15.6 & 44 & 76.0 & - \\
\midrule
\textbf{\ourmethod{}-M2} & WS+EA & \underline{7.7M} & \textbf{\underline{312M}} & \textbf{\underline{11.4}} & \textbf{\underline{37}} & \textbf{\underline{78.6}} & \textbf{\underline{94.3}} \\
MUXNet-l~\cite{muxconv} & EA & \textbf{4.0M} & 318M & 19.2 & 74 & 76.6 & 93.2 \\
EfficientNet-B0 \cite{efficientnet} & RL/scaling & 5.3M & 390M & 14.4 & 46 & 77.1 & 93.2 \\
AtomNAS-C+ \cite{atomnas} & WS+shrinkage & 5.9M & 363M & - & - & 77.6 & 93.5 \\
AutoNL-L \cite{autonl} & gradient & 5.6M & 353M & - & - & 77.7 & 93.7 \\
DNA-c \cite{li2019blockwisely} & gradient & 5.3M & 466M & 14.5 & 67 & 77.8 & 93.7 \\
\midrule
\textbf{\ourmethod{}-M3} & WS+EA & \underline{9.1M} & \textbf{\underline{490M}} & \textbf{\underline{16.1}} & \textbf{\underline{62}} & \textbf{\underline{79.9}} & \textbf{\underline{94.9}} \\
ResNet-152~\cite{resnet} & manual & 60M & 11.3B & 66.7 & 176 & 77.8 & 93.8 \\
MixNet-L \cite{mixnet} & RL & \textbf{7.3M} & 565M & 29.4 & 105 & 78.9 & 94.2 \\
EfficientNet-B1~\cite{efficientnet} & RL/scaling & 7.8M & 700M & 19.5 & 67 & \color{black}79.1 & \color{black}94.4 \\
\midrule
\textbf{\ourmethod{}-M4} & WS+EA & \underline{9.1M} & \textbf{\underline{0.6B}} & \underline{17.3} & \underline{78} & \textbf{\underline{80.5}} & \textbf{\underline{95.2}} \\
\color{black}BigNASModel-L \cite{yu2020bignas} & \color{black}WS & \color{black}\textbf{6.4M} & \color{black}\textbf{0.6B} & \color{black}- & \color{black}- & \color{black}79.5 & \color{black}- \\
\color{black}OnceForAll \cite{onceforall} & \color{black}WS+EA & \color{black}9.1M & \color{black}\textbf{0.6B} & \textbf{\color{black}16.5} & \textbf{\color{black}72} & \color{black}80.0 & \color{black}94.9\\
Inception-v4~\cite{inceptionv4} & manual & 48M & 13B & 84.6 & 206 & 80.0 & 95.0 \\
Inception-ResNet-v2~\cite{inceptionv4} & manual & 56M & 13B & 99.1 & 289 & 80.1 & 95.1 \\
\specialrule{1.5pt}{1pt}{1pt}
\end{tabular}}
}
\vspace{-0.2cm}
\end{table*}

\subsection{\textcolor{black}{Scalability to Datasets}\label{sec:dataset}}
Existing NAS approaches are rarely applied to datasets beyond standard ones (i.e. CIFAR-10 \cite{cifar} and ImageNet \cite{imagenet}), where the classification task is at superordinate-level and the \# of training images are sufficiently large. Instead, they adopt a conventional transfer learning setup~\cite{transfer_learning}, in which the architectures found by searching on standard benchmark datasets are transferred as is, with weights fine-tuned to new datasets. We argue that such a process is conceptually contradictory to the goal of NAS. The architectures transferred from standard datasets are sub-optimal either with respect to accuracy, efficiency or both. On the other hand, by transferring both architecture and weights \ourmethod{} can indeed design bespoke models for each dataset.

We evaluated \ourmethod{} on ten image classification datasets (see Table~\ref{tab:dataset}) that present different challenges in terms of diversity in classes (superordinate vs. fine-grained) and size of training set (large vs small). For each dataset, we run \ourmethod{} with two objectives: maximize top-1 accuracy on validation data (20\% randomly separated from the training set) and minimize \#MAdds. We start from the supernet trained on ImageNet (which is created once before all experiments; see Section~\ref{sec:implementation}) and adapt it to the new dataset. During this procedure, the last linear layer is reset depending on the number of categories in the new dataset. \ourmethod{} is now applied for a total of 30 iterations. In each iteration the supernet is adapted for 5 epochs using SGD with a momentum of 0.9. The learning rate is initialized at 0.01 and annealed to zero in 150 epochs (30 iterations with five epochs in each). All hyperparameters are set at default values from Table~\ref{tab:hyperparameters}. For each dataset, the overall \ourmethod{} process takes slightly under a day on a server with eight 2080Ti GPUs.

\begin{figure*}[t]
    \centering
    \begin{subfigure}{0.95\textwidth}
    \centering
    \includegraphics[width=\textwidth]{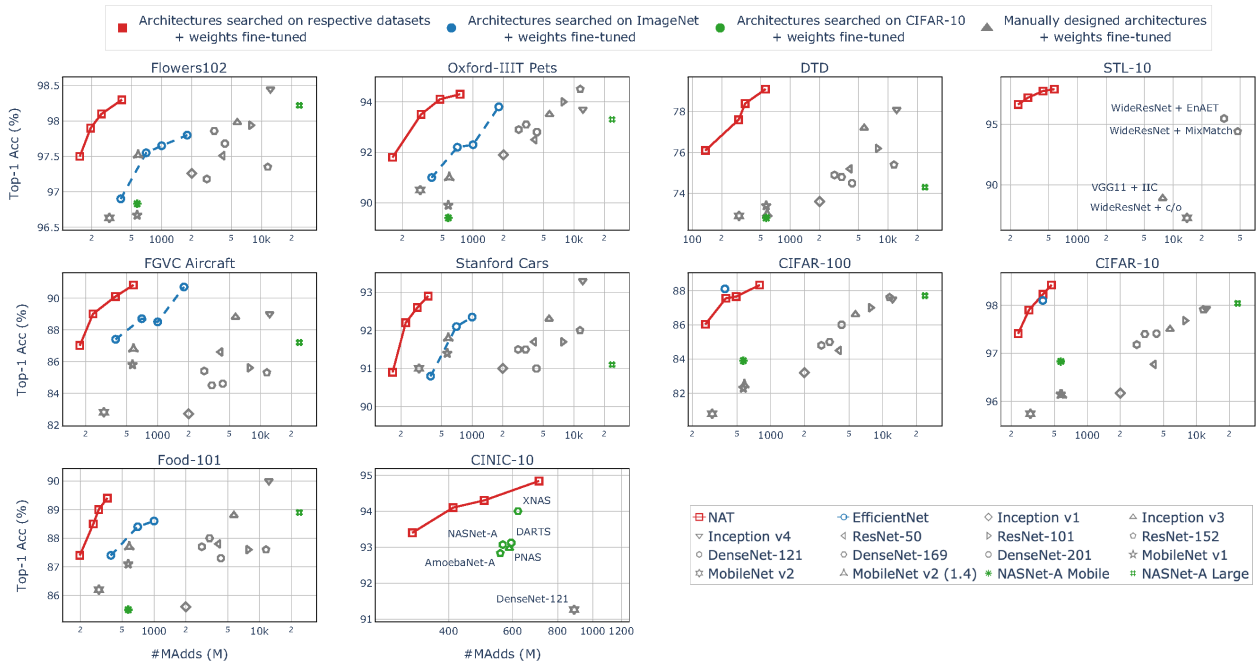}
    \end{subfigure}
\caption{\textbf{MAdds vs. Accuracy} trade-off curves comparing \ourmethod{} and existing architectures on a diverse set of datasets. The datasets are arranged in ascending order of training set size. Methods shown in the legend pre-train on ImageNet and fine-tune the weights on the target dataset. Methods with names annotated in sub-figures train from scratch or use external training data.\label{fig:dataset_anno}}
\vspace{-0.3cm}
\end{figure*}

\begin{figure}[!hbt]
    \centering
    \begin{subfigure}{0.45\textwidth}
    \centering
    \hspace*{1.5mm}
    \includestandalone[width=0.78\textwidth]{figs/legend}\\
    \includestandalone[width=0.95\textwidth]{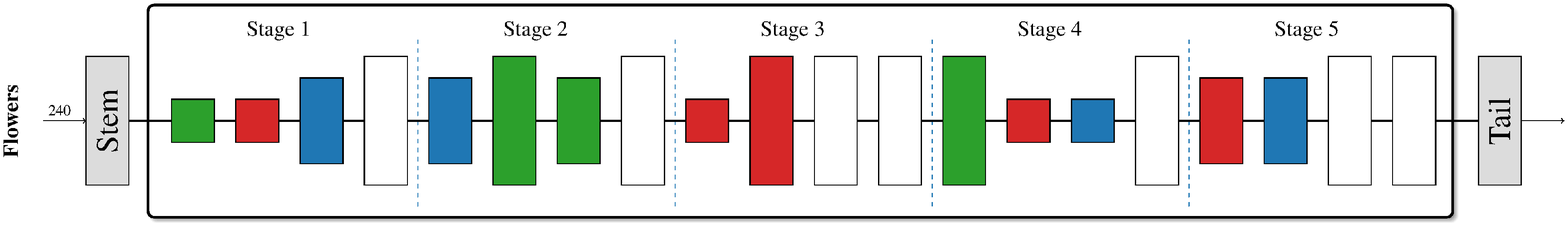}\\
    \includestandalone[width=0.95\textwidth]{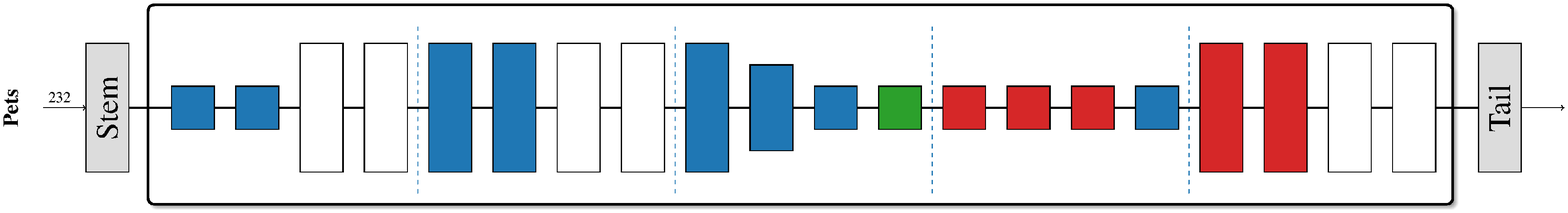}\\
    \includestandalone[width=0.95\textwidth]{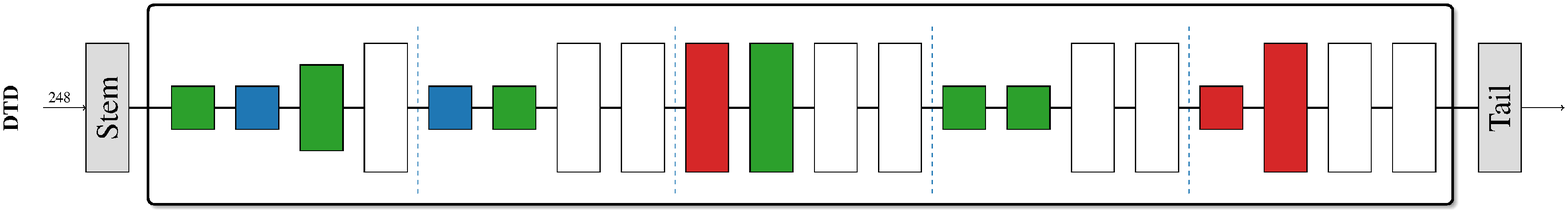}\\
    \includestandalone[width=0.95\textwidth]{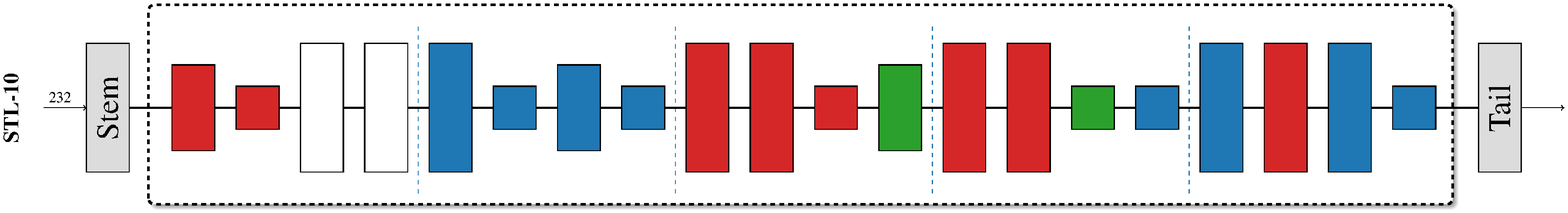}\\
    \includestandalone[width=0.95\textwidth]{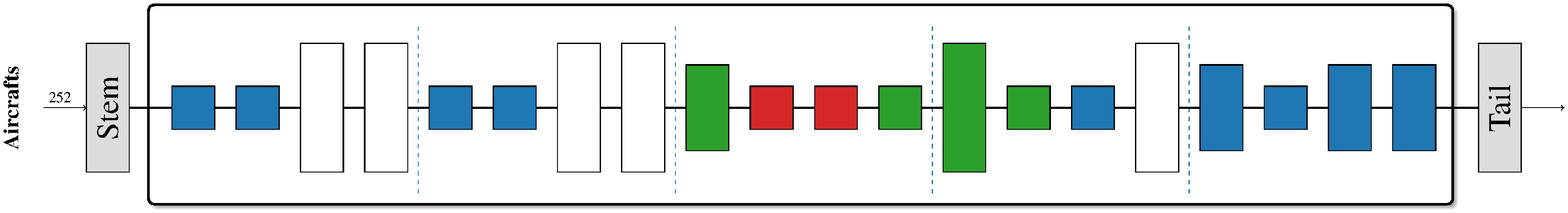}\\
    \includestandalone[width=0.95\textwidth]{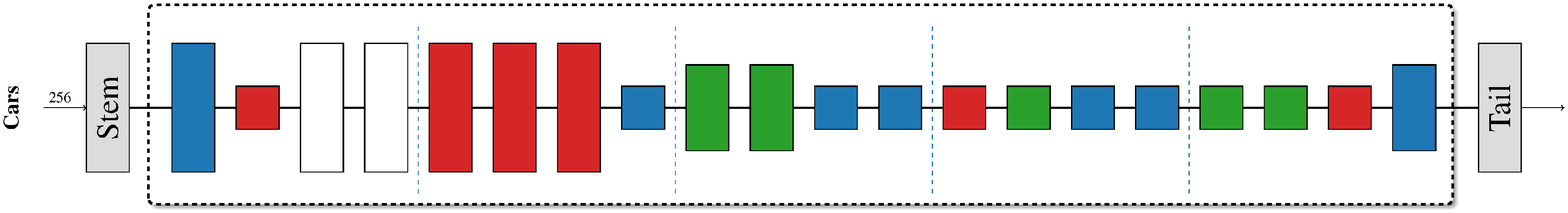}\\
    \includestandalone[width=0.95\textwidth]{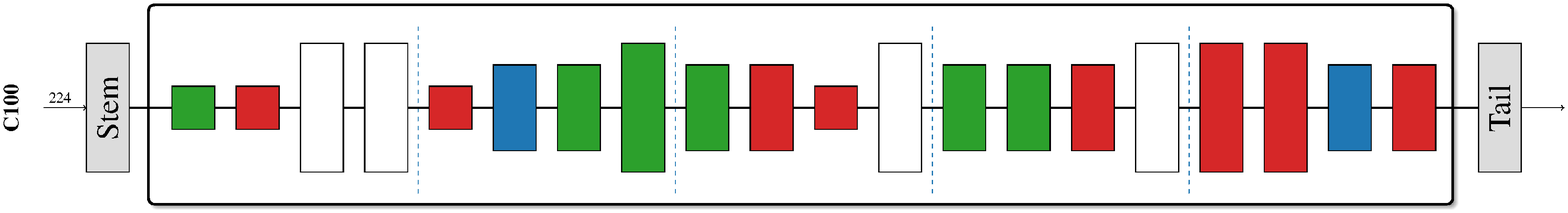}\\
    \includestandalone[width=0.95\textwidth]{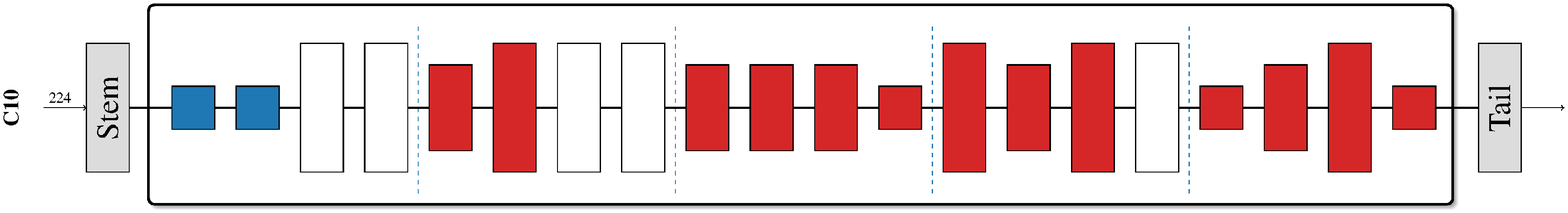}\\
    \includestandalone[width=0.95\textwidth]{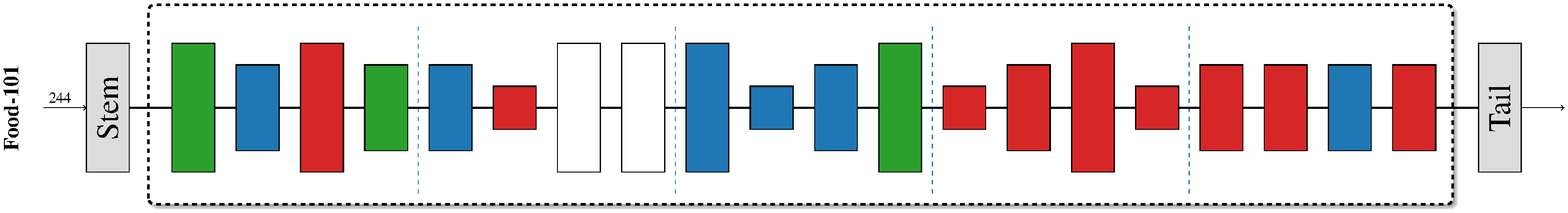}\\
    \includestandalone[width=0.95\textwidth]{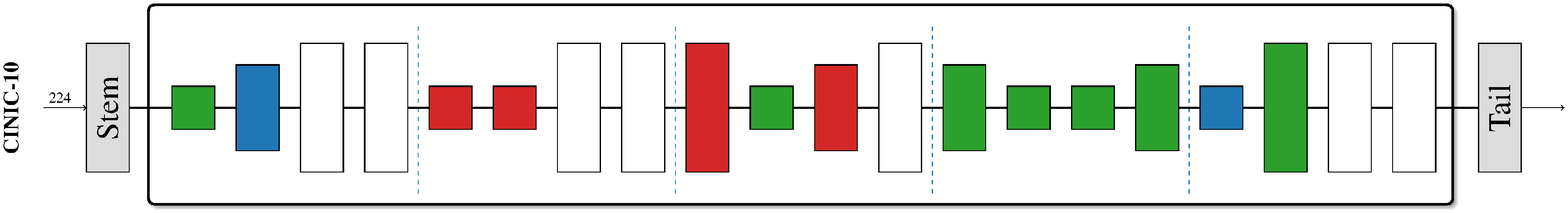}
    \end{subfigure}
\caption{Efficient architectures (350M MAdds) obtained by \ourmethod{} on ten diverse image classification datasets.\label{fig:dataset_archs}}
% \vspace{-0.5cm}
\end{figure}

Fig.~\ref{fig:dataset_anno} shows the accuracy and \#MAdds trade-off for each dataset across a wide range of models, including \ourmodel{}s, existing NAS and hand-designed models. \textcolor{black}{Across all datasets, \ourmodel{}s consistently achieve better accuracy while being an order of magnitude more efficient (\#MAdds) than existing models, suggesting that searching directly on the targeted datasets is a more effective alternative to the conventional transfer learning that fine-tunes weights of architectures learned on standard datasets (i.e. ImageNet and CIFAR-10)}. Under mobile settings ($\leq$ 600M), \ourmodel{}s achieve the state-of-the-art on these datasets, and a new state-of-the-art accuracy on both STL-10 \cite{stl-10} and CINIC-10\footnote{According to \cite{wang2019enaet} for STL-10, and \cite{xnas} for CINIC-10.} \cite{cinic10} datasets. Surprisingly, on small scale datasets e.g. Oxford Flowers102 \cite{flowers102}, Oxford-IIIT Pets \cite{pets}, DTD \cite{dtd} and STL-10 \cite{stl-10}, we observe that \ourmodel{}s are significantly more effective than conventional fine-tuning. Even on fine-grained datasets such as Stanford Cars and FGVC aircraft, where conventional fine-tuning did not improve upon training from scratch, \ourmodel{}s improve accuracy while also being significantly more efficient.

Fig.~\ref{fig:dataset_archs} shows a visualization of architectures with 350M MAdds for each dataset. \textcolor{black}{The lack of similarity in the networks suggest that different datasets require different architectures to be efficient in \emph{accuracy-MAdds}, and \ourmethod{} is able to generate these customized networks for each dataset.} Additional visualization of architectures searched on all datasets is provided in Section~\ref{sec:heatmap}.

\subsection{Scalability to Objectives}
Practical applications of NAS can rarely be considered from the point of view of a single objective, and most often, they must be evaluated from many different, possibly competing, objectives. We demonstrate the scalability of \ourmethod{} to more than two objectives, and evaluate its effectiveness.

We use \ourmethod{} to simultaneously optimize for three objectives---namely, model accuracy on ImageNet, model size (\#params), and model computational efficiency. We consider three different metrics to quantify computational efficiency---\#MAdds, CPU latency, and GPU latency. In total, we run three instances of three-objective search---i.e. maximize accuracy, minimize \#params, and minimize one of \#MAdds, CPU latency or GPU latency. We follow the settings from the ImageNet experiment in Section~\ref{sec:imagenet}, except the fine-tuning step.

\begin{figure*}[t]
    \centering
    \begin{subfigure}{0.9\textwidth}
    \centering
    \includegraphics[width=0.95\textwidth]{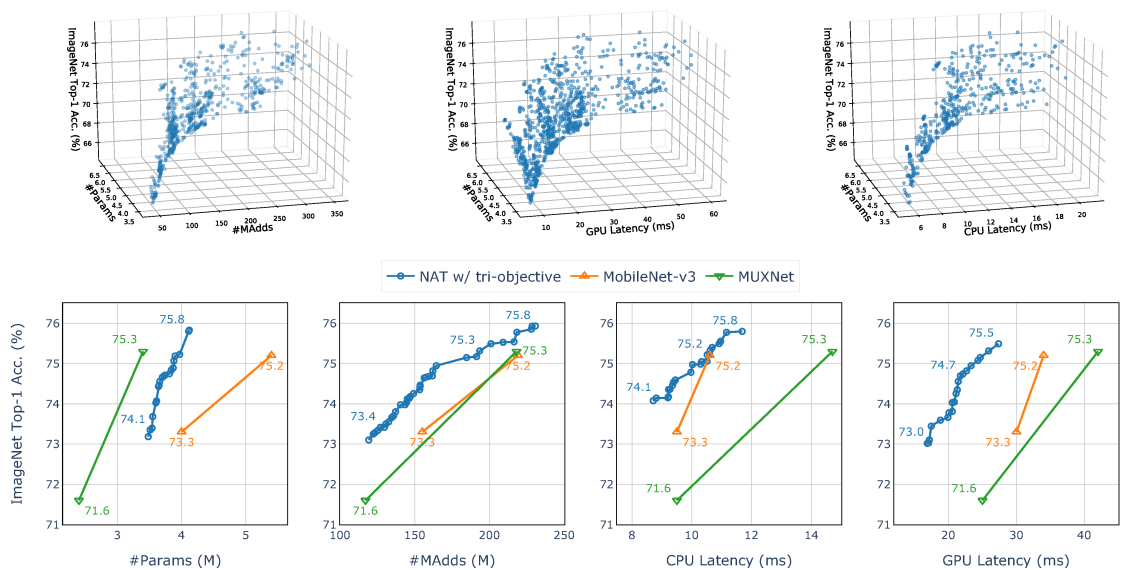}
    \end{subfigure}
\caption{\textbf{Top row:} \ourmodel{}s obtained from tri-objective search to maximize ImageNet top-1 accuracy, minimize model size (\#Params), and minimize \{\#MAdds, CPU latency, GPU latency\} from left to right. {\color{black}Pareto surfaces emerge at higher model complexity regime (i.e. top right corner) suggesting that trade-offs exist between model size (\#params) and model efficiency (\#MAdds and latency)}. \textbf{Bottom row:} 2D projections from above 3D scatter, showing top-1 accuracy vs. each of the four efficiency related measurements. \textcolor{black}{The first two 2D projections are from the first 3D scatter, and the remaining two 2D projections are from the second and third 3D scatters, respectively.} To better visualize (the comparison with MobileNetV3 \cite{mobilenetv3} and MUXNet \cite{muxconv}), partial solutions from the non-dominated frontiers are shown. All top-1 accuracy shown are without fine-tuning.\label{fig:objectives_anno}}
\vspace{-0.3cm}
\end{figure*}

After obtaining the non-dominated (trade-off) solutions, we first visualize the objectives in Fig.~\ref{fig:objectives_anno}. We observe that Pareto surfaces emerge at higher model complexity regime (i.e. high \#params, \#MAdds, etc.), shown in the 3D scatter plot in the top row, suggesting that trade-offs exist between model size (\#params) and model efficiency (\#MAdds and latency). In other words, \#params and \{\#MAdds, CPU, GPU latency\} are not completely correlated---e.g. a model with a fewer \#params is not necessarily more efficient in \#MAdds or latency than another model with more \#params. This is one of the advantages of using a many-objective optimization algorithm compared to optimizing a single scalarized objective (such, as a weighted-sum of objectives \cite{proxylessnas,mnasnet}). 

Fig.~\ref{fig:objectives_anno} visualizes, in 2D, the top-1 accuracy as a trade-off with each one of the four considered efficiency metrics in the bottom row. The 2D projection is obtained by ignoring the third objective. For better visualization we only show the architectures that are close to the performance trade-off of MobilNetV3 \cite{mobilenetv3}. \ourmodel{}s obtained directly from the three-objective search i.e., before any fine-tuning of their weights, consistently outperform MobileNetV3 on ImageNet along all the objectives (top-1 accuracy, \#params, \#MAdds, CPU and GPU latency). Additionally, we compare to MUXNets \cite{muxconv} which are also obtained from a three-objective NAS optimizing \{top-1 accuracy, \#params, and \#MAdds\}. However, MUXNets adopt a search space that is specifically tailored for reducing model size. Therefore, in comparison to MUXNets, we observe that \ourmodel{}s perform favourably on all the remaining three efficiency metrics, except for \#params. Primarily driven by curiosity in terms of pushing the scalability of our approach with respect to number of objectives, we provide an application to 12 objective problem in Section \ref{sec:twelve_obj}.

\subsection{Utility on Dense Image Prediction\label{sec:tasks}}
Dense image prediction is another series of important computer vision tasks, that assigns a label to each pixel in the input image \cite{chen2017deeplab,newell2016stacked}. Success in these tasks relies on both feature extraction via a backbone CNN, e.g. ResNet \cite{resnet}, and feature aggregation, e.g. FPN \cite{lin2017feature}, at multiple scales. In this section, we use \ourmethod{} to design efficient backbone feature extractors for semantic segmentation, to demonstrate its utility beyond image classification. 

Similar to previous studies, we start from the supernet trained on ImageNet (which is created once before all experiments; see Section~\ref{sec:implementation}). We remove the last classification layer and pair it with the BiSeNet segmentation heads \cite{yu2018bisenet}, a lightweight semantic segmentation framework for real-time performance. We modify the searched input resolutions from [192, $\ldots$, 256] to [512, $\ldots$, 1280] and keep other searched options the same as before. \ourmethod{} is applied to minimize \#MAdds and maximize mIoU on validation data (20\% randomly sampled from the training set) for 20 iterations. In each iteration, the supernet is adapted for 2K iterations using SGD with a momentum of 0.9 and weight decay of \num{5e-4}. We use a batch size of eight for each GPU. We use an initial learning rate of 0.01 and follow the ``poly'' learning rate schedule from the original BiSeNet \cite{yu2018bisenet}, in which the initial learning rate is multiplied by $(1 - \frac{{iter}}{{max\_iter}})^{{0.9}}$ in each iteration. All other hyperparameters are set at default values from Table~\ref{tab:hyperparameters}. On the Cityscapes dataset \cite{cityscapes}, the overall \ourmethod{} process takes a day on a server with six Titan RTX GPUs. 

\begin{figure}[!hbt]
    \centering
    \begin{subfigure}{0.45\textwidth}
    \centering
    \includegraphics[width=0.85\textwidth]{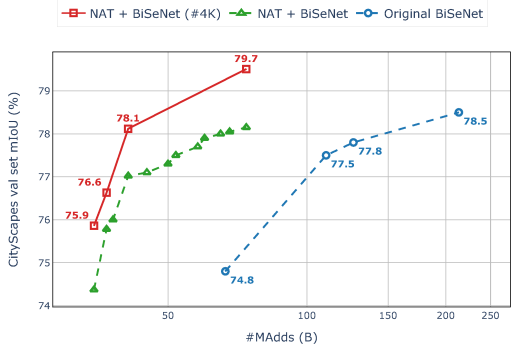}
    \end{subfigure}
\caption{\textbf{MAdds vs. Cityscapes mIoU}. \ourmethod{} obtained backbone feature extractors (\textcolor{green}{green} curve) significantly outperform the original BiSeNet, which are based on ResNets (R18 - R152). With further fine-tuning of 4K iterations, \ourmethod{} achieves the state-of-the-art performance (\textcolor{red}{red} {curve}).\label{fig:cityscapes_results}}
\end{figure}

Fig.~\ref{fig:cityscapes_results} compares the \emph{mIoU-MAdds} trade-off obtained by \ourmethod{} and the original BiSeNet \cite{yu2018bisenet} on the Cityscapes dataset. Empirically, we observe that \ourmethod{} based backbones consistently outperform the original BiSeNets, which are based on ResNets. To realize the full potential of the searched \ourmodel{}s, we further fine-tune the obtained models for 4K iterations. As shown in Table~\ref{tab:cityscapes}, the resulting \ourmethod{} model yields comparable performance against state-of-the-art methods, including PSPNet \cite{pspnet}, DeepLabv3 \cite{deeplabv3+}, Auto-DeepLab-S \cite{autodeeplab}, while being \textbf{4x} - \textbf{28x more efficient} in \#Madds.

\begin{table}[hbt]
    \centering
    \caption{\textbf{Cityscapes Semantic Segmentation \cite{cityscapes}:} All results are based on \emph{single-scale} inputs from validation set.\label{tab:cityscapes}}
    \resizebox{0.4\textwidth}{!}{%
    \begin{tabular}{@{\hspace{2mm}}l|cc|c@{\hspace{2mm}}}
    \toprule
    Method & \#Params & \#Multi-Adds & mIoU (\%) \\ \midrule
    BiSeNet \cite{yu2018bisenet} & 13.4M & 67B & 74.8 \\
    PSPNet \cite{pspnet} & 65.9M & 2,017B & 78.4 \\
    DeepLabv3+ \cite{deeplabv3+} & 43.5M & 1,551B & 79.6 \\
    Auto-DeepLab-S \cite{autodeeplab} & 10.2M & 333B & \textbf{79.7} \\ \midrule
    \ourmethod{} + BiSeNet (ours) & \textbf{8.8M} & \textbf{73B} & \textbf{79.7} \\ \bottomrule
    \end{tabular}%
    }
\end{table}

\section{Ablation Study\label{sec:ablation}}
In this section, we provide additional experiments towards quantifying the impacts of the main components introduced in \ourmethod{} and hyperparameter analysis. 

\subsection{Accuracy Predictor Performance}
\begin{figure*}[t]
    \centering
    \begin{subfigure}{0.95\textwidth}
    \centering
    \includegraphics[width=\textwidth{}]{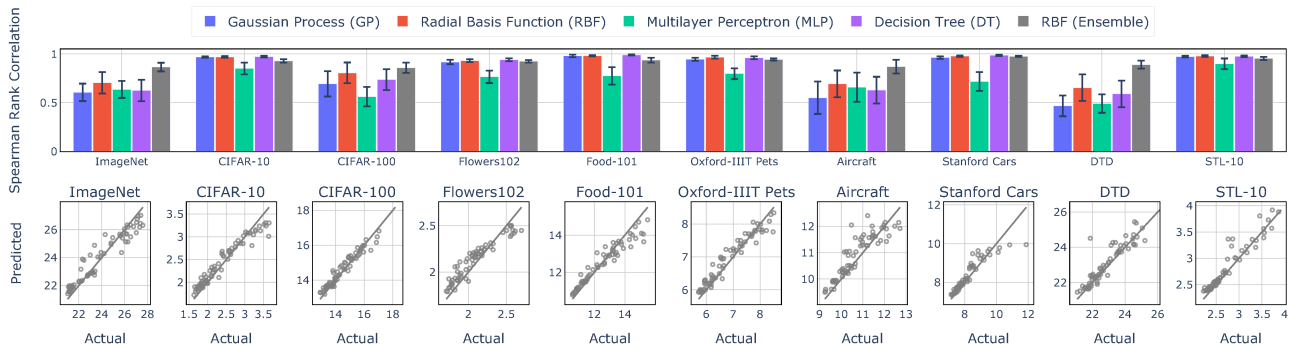}
    \end{subfigure}
\caption{\textbf{Top row:} Spearman rank correlation between predicted accuracy and true accuracy of different surrogate models across many datasets. Each accuracy predictor is constructed from 250 samples (trained architectures). Error bars show mean and standard deviation over ten runs. \textbf{Bottom row:} Goodness of fit visualization of RBF ensemble, the best accuracy predictor.\label{fig:accuracy_predictor2}}
\vspace{-0.3cm}
\end{figure*}

In this subsection, we assess the effectiveness of different accuracy predictor models. We first uniformly sampled 350 architectures from our search space and trained them using SGD for 150 epochs on ImageNet. Each one of them is fine-tuned for 50 epochs on the other ten datasets (Table~\ref{tab:dataset}). From the 350 pairs of architectures and top-1 accuracy computed on each dataset, we reserved a subset (randomly chosen) of 50 pairs for testing, and the remaining 300 pairs are then available for training the predictor models.

Fig.~\ref{fig:accuracy_predictor1} compares the mean (over 11 datasets) Spearman rank correlation between the predicted and the true accuracy for each accuracy predictor as the training sample size is varied from 50 to 300. Empirically, we observe that radial basis function (RBF) has higher Spearman rank correlation compared to the other three models. The proposed RBF ensemble model further improves performance over the standalone RBF model across all training sample size regimes. Fig.~\ref{fig:accuracy_predictor2} shows a visualization of the comparative performance of predictor models on different datasets. From the trade-off perspective of minimizing number of training examples (which reduces the overall computational cost) and maximizing Spearman rank correlation in prediction (which improves the accuracy in ranking architectures during search), we chose the RBF ensemble as our accuracy predictor model and a training size of 100 for all our experiments.

\subsection{Search Efficiency}
\begin{figure}[t]
    \centering
    \begin{subfigure}{0.48\textwidth}
    \centering
    \includegraphics[width=\textwidth{}]{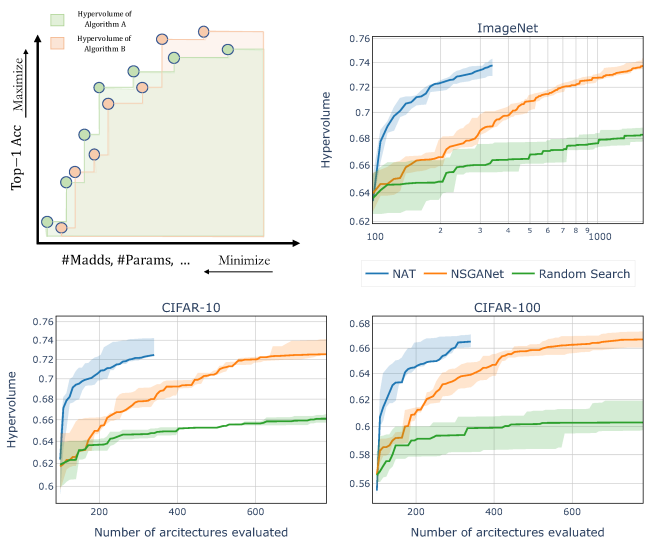}
    \end{subfigure}
\caption{\textbf{Top left:} A sketch visualizing the hypervolume metric \cite{hypervolume}. In case of bi-objective, it measures the dominated area achieved by a multi-objective algorithm. A larger hypervolume indicates a better Pareto front achieved. \textbf{Rest:} Search efficiency comparison between \ourmethod{}, NSGANet \cite{NSGANet}, and random search under a bi-objective setup. Mean hypervolume over five runs are plotted with shaded region showing the standard deviation.\label{fig:search_efficiency}}
\vspace{-0.3cm}
\end{figure}
\begin{table*}[t]
\centering
\caption{Comparing the relative search efficiency of \ourmethod{} to other methods. ``--'' denotes for not applicable, ``WS'' stands for weight sharing and ``SMBO'' stands for sequential model-based optimization \cite{hutter2011sequential}. $^{\dagger}$ is taken from \cite{fbnetv2}, $^{\ddagger}$ estimate based on the \# of models evaluated during search (20K in \cite{nasnet}, 1.2K in \cite{PNAS}, 27K in \cite{amoebanet}). $^{*}$ denotes re-ranking stage where top 100-250 models undergo extended training and evaluation for 300 epochs before selecting the final model.\label{tab:search_cost}}
\resizebox{0.8\textwidth}{!}{%
\begin{tabular}{@{\hspace{2mm}}l|lc|cc|ccc|c@{\hspace{2mm}}}
\toprule
\multirow{2}{*}{} & \multirow{2}{*}{Method} & \multirow{2}{*}{Type} & \multirow{2}{*}{\begin{tabular}[c]{@{}c@{}}Top-1 \\      Acc. (\%)\end{tabular}} & \multirow{2}{*}{\begin{tabular}[c]{@{}c@{}}\#MAdds \\      (M)\end{tabular}} & \multicolumn{4}{c}{Estimated Search Cost   (GPU hours)} \\ \cmidrule(l){6-9} 
 &  &  &  &  & Prior-search & During-search & Post-search & Total \\ \midrule
\multirow{3}{*}{ImageNet} & MnasNet \cite{mnasnet} & gradient & 75.2 & 312 & - & - & - & 91k$^{\dagger}$ \\
 & OnceForAll \cite{onceforall} & WS+EA & 76.9 & 230 & 1,200 & 40 & 75 & 1.3k \\
 & \textbf{\ourmethod{} (ours)} & WS+EA & \textbf{77.5} & \textbf{225} & 1,200 & 150 & 75 & 1.4k \\ \midrule
\multirow{5}{*}{CIFAR-10} & NASNet \cite{nasnet} & RL & 97.4 & 569 & - & 10,000$^{\ddagger}$ & 6000$^{*}$ & 16k \\
 & PNASNet \cite{PNAS} & SMBO & 96.6 & 588 & - & 600$^{\ddagger}$ & 36 & 0.6k \\
 & DARTS \cite{darts} & WS+gradient & 97.3 & 595 & - & 96 & 36 & 0.1k \\
 & AmoebaNet \cite{amoebanet} & EA & 97.5 & 555 & - & 13,500$^{\ddagger}$ & 2400$^{*}$ & 16k \\
 & \textbf{\ourmethod{} (ours)} & transfer+EA & \textbf{98.4} & \textbf{468} & - & 150 & - & 0.1k \\ \bottomrule
\end{tabular}
\vspace{-0.2cm}
}
\end{table*}
The overall computation cost consumed by a NAS algorithm can be factored into three phases: (1) \emph{Prior-search}: Cost incurred prior to architecture search, e.g. training supernet in case of one-shot approaches \cite{one-shot,onceforall} or constructing accuracy predictor \cite{chamnet}, etc; (2) \emph{During-search}: Cost associated with measuring the performance of candidate architectures sampled during search through inference. It also includes the cost of training the supernet in case it is coupled with the search, like in \cite{darts} and \ourmethod{}; (3) \emph{Post-search}: Cost associated with choosing a final architecture, and/or fine-tuning/re-training the final architectures from scratch. For comparison, we select representative NAS algorithms, including those based on reinforcement learning (RL), gradients, evolutionary algorithm (EA), and weight sharing (WS). Table~\ref{tab:search_cost} shows results for ImageNet and CIFAR-10. The former is the dataset on which the supernet is trained and the latter is a proxy for transfer learning to a non-standard dataset. \ourmethod{} consistently achieves better performance, both in terms of top-1 accuracy and model efficiency (e.g. \#MAdds), compared to the baselines while computational cost is similar or lower. The primary computational cost of \ourmethod{} is the \emph{prior-search} training of supernet for 1200 hours. We emphasize, again, that it is a one-time cost that is amortized across all subsequent deployment scenarios (e.g. 10 additional datasets in Section~\ref{sec:dataset}).

Comparing the search phase contribution to the success of different NAS algorithms is challenging and ambiguous due to substantial disparities in search spaces and training procedures. So, we conduct the following controlled experiment where we replace only the evolutionary search component in the \ourmethod{} pipeline with (1) a \emph{random search} that uniformly samples (with possible repetition) from the search space, and (2) \emph{NSGANet} \cite{NSGANet}, another multi-objective EA-based NAS algorithm. This experiment is under a bi-objective setup: maximize top-1 accuracy and minimize \#MAdds. We run each method five times on three datasets to capture the variance in performance due to inherent stochasticity in the optimization initialization. We use hypervolume \cite{hypervolume}, a widely-used metric for comparing algorithms under multiple objectives, as the evaluation metric. Fig.~\ref{fig:search_efficiency} shows the mean and the standard deviation of the hypervolume achieved by each method. The evolutionary search component in \ourmethod{} is $3\times$ - $5\times$ more sample efficient than the baselines for the same hypervolume.

\subsection{Analysis of Crossover}
\begin{figure*}[t]
    \centering
    \begin{subfigure}{0.9\textwidth}
    \centering
    \includegraphics[width=0.95\textwidth{}]{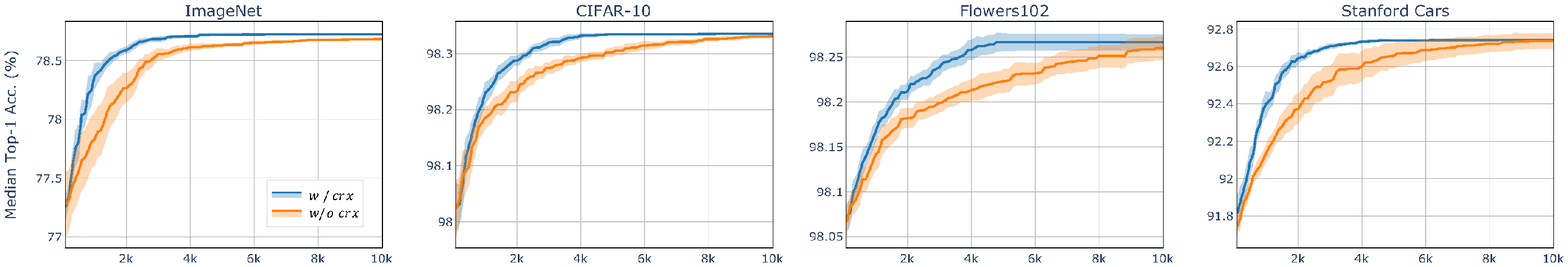}
    \caption{Effect of Crossover\label{fig:crx_types}}
    \end{subfigure}\\
    \begin{subfigure}{0.9\textwidth}
    \centering
    \includegraphics[width=0.95\textwidth{}]{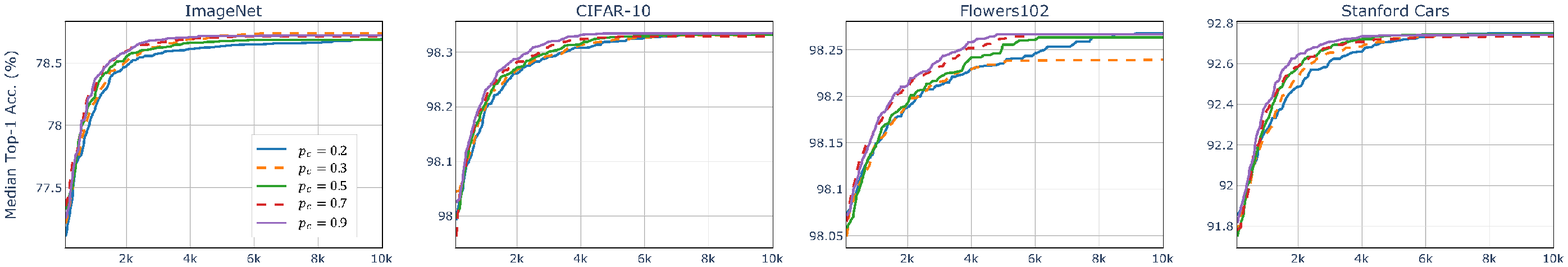}
    \caption{Effect of Crossover Probability\label{fig:crx_prob}}
    \end{subfigure}
\caption{Ablation study on the crossover operator: (a) the median performance from eleven runs of our evolutionary algorithm with and without the crossover operator. (b) the median performance deteriorates as the crossover probability reduces from 0.9 to 0.2.\label{fig:crossover_ablation}}
\vspace{-0.3cm}
\end{figure*}
Crossover is a standard operator in evolutionary algorithms, but has largely been avoided by existing EA-based NAS methods \cite{real2017large,liu2018hierarchical,amoebanet}. But as we demonstrate here, a carefully designed crossover operation can significantly improve search efficiency. We run the evolutionary search of \ourmethod{} with and without the crossover operator on four datasets; ImageNet \cite{imagenet}, CIFAR-10 \cite{cifar}, Oxford Flowers102 \cite{flowers102}, and Stanford Cars \cite{stanford_cars}. The hyperparameters that we compare are:
\begin{enumerate}
    \item \emph{w/ crx}: crossover probability of 0.9; mutation probability of 0.1; mutation index $\eta_m$ of 3. 
    \item \emph{w/o crx}: crossover probability of 0.0; mutation probability of 0.2; mutation index $\eta_m$ of 3.
\end{enumerate}
We double the mutation probability when crossover is not used to compensate for the reduced exploration ability of the search. On each dataset, we run each setting to maximize the top-1 accuracy 11 times and report the median performance as a function of the number of architecture sampled in Fig~\ref{fig:crx_types}. On all four datasets, the crossover operator significantly improves the efficiency of the evolutionary search algorithm. To further validate, we sweep over the probability of crossover while maintaining the rest of the settings. The median performance (over 11 runs) deteriorates as the crossover probability is reduced from 0.9 to 0.2 (see Fig.~\ref{fig:crx_prob}).

\subsection{Analysis of Mutation Hyperparameters}
\begin{figure*}[t]
    \centering
    \begin{subfigure}{0.9\textwidth}
    \centering
    \includegraphics[width=0.95\textwidth{}]{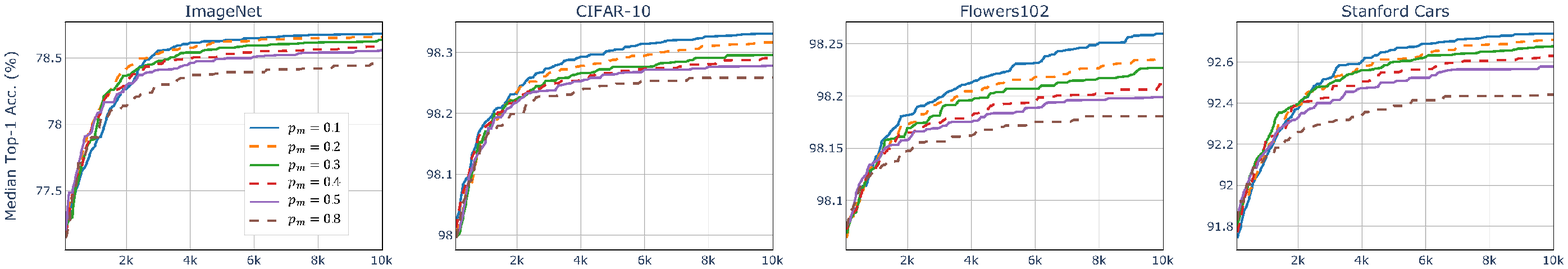}
    \caption{Effect of Mutation Probability\label{fig:mut_prob}}
    \end{subfigure}\\
    \begin{subfigure}{0.9\textwidth}
    \centering
    \includegraphics[width=0.95\textwidth{}]{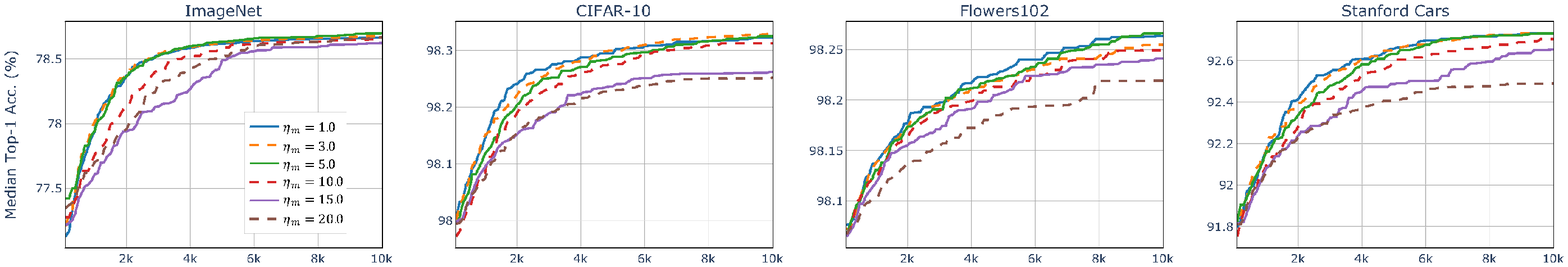}
    \caption{Effect of Mutation Hyperparameter $\eta_m$\label{fig:mut_eta}}
    \end{subfigure}
\caption{Hyperparameter study on (a) mutation probability $p_m$ and (b) mutation index parameter $\eta_m$. For each study, we run \ourmethod{} eleven times on four datasets to maximize top-1 accuracy and report the median performance.\label{fig:hyperparameters}}
\vspace{-0.3cm}
\end{figure*}

The mutation operator used in \ourmethod{} is controlled via two hyperparameters---namely, the mutation probability $p_m$ and mutation index $\eta_m$. To identify the optimal hyperparameter values, we conduct the following parameter sweep experiments. Setting the rest of the hyperparameters to their default values (see Table~\ref{tab:hyperparameters}), we sweep the value of $p_m$ from 0.1 to 0.8, and $\eta_m$ from 1.0 to 20. And for each setting, we run \ourmethod{} eleven times on four datasets (same as the crossover experiment) to maximize the top-1 accuracy. Figs.~\ref{fig:mut_prob} and \ref{fig:mut_eta} show the effect of mutation probability $p_m$ and mutation index $\eta_m$, respectively. We observe that increasing the mutation probability has an adverse effect on performance. Similarly, low values of $\eta_m$, which encourages the mutated offspring to be further away from parent architectures, improves the performance. Based on these observations, we set the mutation probability $p_m$ and mutation index $\eta_m$ parameters to 0.1 and 1.0, respectively, for all our experiments in Section~\ref{sec:experiments}.

\subsection{Effectiveness of Supernet Adaptation\label{sec:abl_supernet}}
\begin{figure*}[t]
    \centering
    \begin{subfigure}{0.9\textwidth}
    \centering
    \includegraphics[width=0.95\textwidth{}]{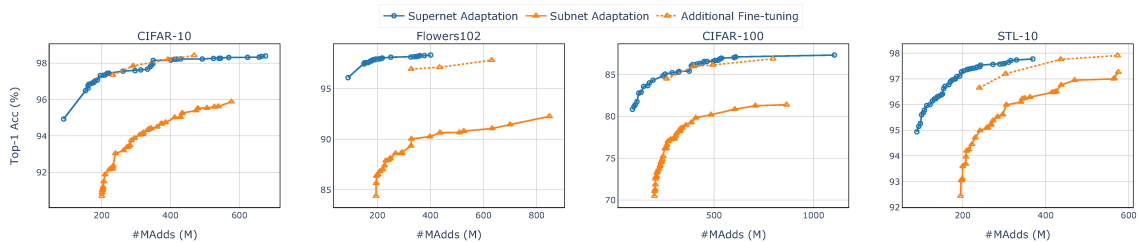}
    \end{subfigure}
\caption{Comparing the performance of \emph{adapting supernet}, \emph{adapting subnet} and \emph{additional fine-tuning} under a bi-objective search setup on four datasets. Details are provided in Section~\ref{sec:abl_supernet}.\label{fig:abl_supernet}}
\vspace{-0.3cm}
\end{figure*}

Recall that \ourmethod{} adopts any supernet trained on a large-scale dataset, e.g. ImageNet, and seeks to efficiently transfer to a task-specific supernet on a given dataset. Here, we compare this procedure to a more conventional approach of adapting every subnet (candidate architectures in search) directly. Specifically, we consider the following,

\begin{enumerate}
    \item \emph{Supernet Adaptation}: fine-tune supernet for 5 epochs in each iteration and use accuracy from inherited weights (without further training) to select architectures during search (adopted in \ourmethod{}).
    \item \emph{Subnet Adaptation}: fine-tune each subnet for 5 epochs from the inherited weights, then measure the accuracy.
    \begin{comment}
    \item \emph{Subnet Training}: train the subnets for 155 epochs from the inherited weights, then measure the accuracy.
    \end{comment}
\end{enumerate}

We apply these two approaches to a bi-objective search of maximizing top-1 accuracy and minimizing \#MAdds on four datasets, including CIFAR-10, CIFAR-100, Oxford Flowers102, and STL-10. Figure~\ref{fig:abl_supernet} compares the final Pareto fronts. Adapting the supernet yields significantly better performance than adapting individual subnets. Furthermore, we select a subset of searched subnets after \emph{subnet adaptation} and fine-tune their weights for an additional 150 epochs. We refer to this as \emph{additional fine-tuning} in Fig.~\ref{fig:abl_supernet}. Empirically, we observe that further fine-tuning can match the performance of \emph{supernet adaptation} on datasets with larger training samples per class (e.g. 4,000 in CIFAR-10). On datasets with fewer samples per class (e.g. 20 in Flowers 102), there is still a large performance gap between \emph{supernet adaptation} and \emph{additional fine-tuning}. Overall the results suggest that \emph{supernet adaptation} is more effective on tasks with limited training samples.

\subsection{Towards Quantifying Architectural Advancement\label{sec:abl_architecture}}
Comparing the architectural contribution to the success of different NAS algorithms can be difficult and ambiguous due to substantial differences in training procedures, e.g. data augmentation, training hyperparameters, etc. Therefore, to quantify the architectural advancement made by \ourmethod{} alone, we train \ourmethod{}-M1 from randomly initialized weights (instead of inheriting them from the supernet) with standard training hyperparameters (see Table~\ref{tab:trn_hyper_settings}). We then compare the outcome to two other recently proposed efficient models, MobileNetV3 \cite{mobilenetv3} and FBNetV2 \cite{fbnetv2}. The results are summarized in Table~\ref{tab:abl_architectural_contribution}, where we observe that the NAT searched model, NAT-M1, is \textbf{0.5} - \textbf{1.0\% more accurate} on ImageNet than compared models using similar or less \#MAdds. 

\begin{table}[!hbt]
\centering
\caption{Details of training hyperparameter settings. Advance settings are in addition to standard settings.\label{tab:trn_hyper_settings}}
\resizebox{0.48\textwidth}{!}{%
\begin{tabular}{@{\hspace{2mm}}l|c|c|c|c@{\hspace{2mm}}}
\toprule
Setting & Data Augmentation & Regularization & Optimizer & LR Schedule \\ \midrule
Standard & Horizontal Flop + Crop & Drop out & \multirow{3}{*}{\begin{tabular}[c]{@{}c@{}} \\ RMSProp + Exponential   \\      Moving Averaging\end{tabular}} & \multirow{3}{*}{\begin{tabular}[c]{@{}c@{}} \\ Step LR w/ Decay\\ + Linear Warm-up \cite{lr-warm-up}\end{tabular}} \\\cmidrule(r){1-3}
Advance & \begin{tabular}[c]{@{}c@{}}+ Random Augmentation \cite{Cubuk_2020_CVPR_Workshops}\\ + Random Erase Pixel \cite{zhong2020random}\end{tabular} & + Drop path \cite{drop_path} &  &  \\ \bottomrule
\end{tabular}%
}
\end{table}

To further quantify the architectural advancement made by NAT, we use NAT-M1 as a drop-in replacement of the backbone feature extractor for three dense image prediction tasks, including object detection, semantic segmentation, and instance segmentation. More specifically, we replace the EfficientNet-B0 \cite{efficientnet} in EfficientDet-D0 \cite{tan2020efficientdet} for object detection; the ResNet-18 \cite{resnet} in BiSeNet \cite{yu2018bisenet} for semantic segmentation; and the ResNet-50 \cite{resnet} in YOLACT \cite{bolya2019yolact} for instance segmentation. For comparison, we apply the same procedure to both MobileNetV3 and FBNetV2 as well. The results are reported in Table~\ref{tab:abl_architectural_contribution}. In general, our NAT searched model, NAT-M1, is consistently better than peer competitors across all tasks and datasets using similar or less \#MAdds. Specifically, NAT-M1 is better than the compared models on all three datasets for semantic segmentation, achieving \textbf{1.0} - \textbf{2.3 higher mIoU}. 

\begin{table}[!hbt]
\centering
\caption{Comparison between \ourmethod{} searched model and representative models on ImageNet classification under standard training setup, and as feature extractors on MS COCO \cite{mscoco} object detection task, PASCAL VOC \cite{pascal-voc} instance segmentation task and semantic segmentation tasks.\label{tab:abl_architectural_contribution}}
\resizebox{0.49\textwidth}{!}{%
\begin{tabular}{@{}ll|ccc@{}}
\toprule
\multicolumn{2}{l|}{Backbone} & MobileNetV3 \cite{mobilenetv3} & FBNetV2 \cite{fbnetv2} & NAT-M1 (ours) \\ \midrule
\multicolumn{2}{l|}{\#MAdds} & 219M & 238M & 225M \\ \midrule
\multicolumn{2}{l|}{ImageNet   Top-1 Acc.} & 74.7 & 75.2 & \textbf{75.7} \\ \midrule
\multicolumn{1}{l|}{\multirow{2}{*}{\begin{tabular}[c]{@{}l@{}}Object \\ Detection\end{tabular}}} & AP & 31.8 & 31.1 & \textbf{32.2} \\
\multicolumn{1}{l|}{} & AP s/m/l & 10.4 / 37.3 / \textbf{50.1} & 10.9 / 36.6 / 48.4 & \textbf{11.5} / \textbf{37.9} / 49.7 \\ \midrule
\multicolumn{1}{l|}{\multirow{2}{*}{\begin{tabular}[c]{@{}l@{}}Instance \\ Segmentation\end{tabular}}} & AP bbox & 44.0 & 44.8 & \textbf{45.2} \\
\multicolumn{1}{l|}{} & AP mask & 43.6 & 43.9 & \textbf{44.3} \\ \midrule
\multicolumn{1}{l|}{\multirow{3}{*}{\begin{tabular}[c]{@{}l@{}}Semantic\\ Segmentation\end{tabular}}} & Cityscapes \cite{cityscapes} & 73.0 & 72.6 & \textbf{74.0} \\
\multicolumn{1}{l|}{} & PASCAL VOC \cite{pascal-voc} & 73.8 & 73.6 & \textbf{75.9} \\
\multicolumn{1}{l|}{} & COCO-Stuff \cite{coco-stuff} & 28.5 & 28.5 & \textbf{29.5} \\ \bottomrule
\end{tabular}%
}
\end{table}

Finally, we break down the effect of different training settings and additional fine-tuning for the Top-1 accuracy of the searched models in Table~\ref{tab:abl_train_setting}. The advance setting in Table~\ref{tab:trn_hyper_settings} also uses knowledge distillation \cite{onceforall,yu2020bignas}. 

\begin{table}[!hbt]
\centering
\caption{Effect of different training setups. Details of the standard and advanced settings under \emph{Random Initialization} are provided in Table~\ref{tab:trn_hyper_settings}.\label{tab:abl_train_setting}}
\resizebox{0.4\textwidth}{!}{%
\begin{tabular}{@{\hspace{2mm}}l|cc|cc@{\hspace{2mm}}}
\toprule
\multirow{2}{*}{\begin{tabular}[c]{@{}l@{}}Training\\ Settings\end{tabular}} & \multicolumn{2}{c|}{Random Initialization} & \multicolumn{2}{c}{Inherited from Supernet} \\ \cmidrule(l){2-5} 
 & standard & advanced & w/o fine-tune & w/ fine-tune \\ \midrule
\ourmethod{}-M1 & 75.7 & 77.1 & 75.9 & 77.5 \\
\ourmethod{}-M2 & 76.9 & 78.0 & 77.4 & 78.6 \\
\ourmethod{}-M3 & 78.2 & 79.1 & 78.9 & 79.9 \\
\ourmethod{}-M4 & 78.8 & 79.5 & 79.4 & 80.5 \\ \bottomrule
\end{tabular}%
}
\end{table}

\section{Conclusion}
This paper considered the problem of designing custom neural network architectures that trade-off multiple objectives for a given image classification task. We introduced \emph{Neural Architecture Transfer} (\ourmethod{}), a practical and effective approach for this purpose. We described our efforts to harness the concept of a supernet and an evolutionary search algorithm for designing task-specific neural networks trading-off accuracy and computational complexity. We also showed how to use an online regressor, as a surrogate model to predict the accuracy of subnets in the supernet. Experimental evaluation on eleven benchmark image classification datasets, ranging from large-scale multi-class to small-scale fine-grained tasks, showed that networks obtained by \ourmethod{} outperform conventional fine-tuning based transfer learning, while being orders of magnitude more efficient under mobile settings ($\leq$ 600M Multiply-Adds). \ourmethod{} was especially effective for small-scale fine-grained tasks where fine-tuning pre-trained ImageNet models is ineffective. Finally, we also demonstrated the utility of \ourmethod{} in optimizing up to twelve objectives with a subsequent trade-off analysis procedure for identifying a single preferred solution. Overall, \ourmethod{} is the first large scale demonstration of many-objective neural architecture search for designing custom task-specific models on diverse image classification datasets.

\bibliographystyle{IEEEtran}
\bibliography{egbib}

\begin{appendices}
    \section{Relation to Existing One-Shot NAS\label{sec:one-shot}}
\begin{figure*}[!hbt]
    \centering
    \begin{subfigure}{0.98\textwidth}
    \centering
    \includegraphics[width=0.98\textwidth{}]{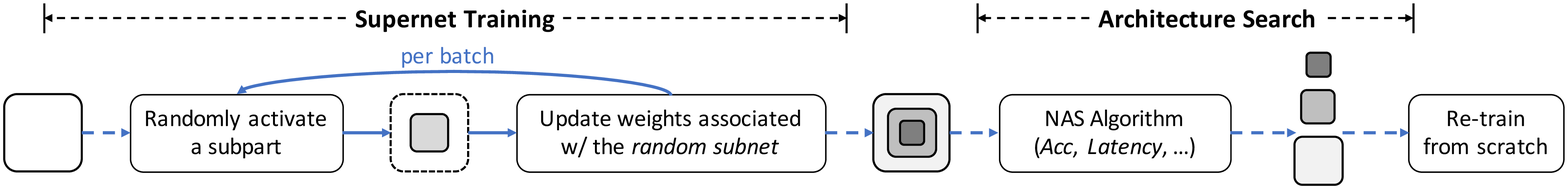}
    \caption{A typical process that most early One-Shot NAS approaches follow \cite{one-shot,guo2019single}.\label{fig:typical_oneshotnas}}
    \end{subfigure}
    \centering
    \begin{subfigure}{0.98\textwidth}
    \centering
    \vspace{3mm}
    \includegraphics[width=0.98\textwidth{}]{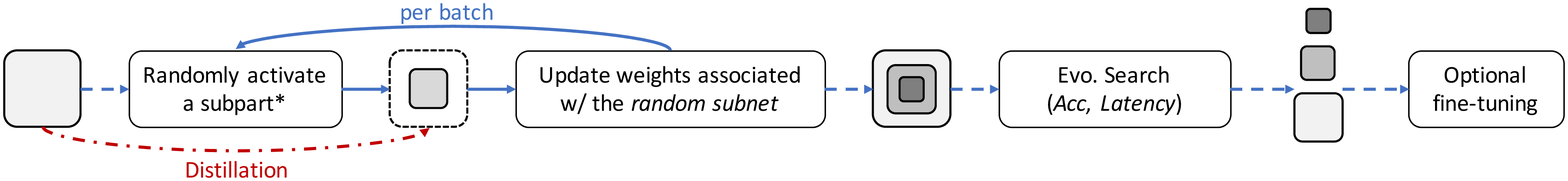}
    \caption{The Progressive Shrinking algorithm proposed in OnceForAll \cite{onceforall}. It pre-trains the supernet at full scale before subnet sampling and use the supernet at full scale to supervise the training of subnets. *And the searched dimensions are gradually added to the search space, i.e. kernel size --$>$ kernel size + depth --$>$ kernel size + depth + width.\label{fig:progressive_shrinking}}
    \end{subfigure}
\caption{Overview of existing one-shot NAS approaches, which decouples the supernet training and architecture search to two sequential steps.}
\end{figure*}

Most existing one-shot NAS approaches follow a two-step process, where the \emph{supernet training} and the \emph{architecture search} are disentangled into two sequential stages. This process starts with training a supernet (in which searchable architectures become \emph{subnets}) offline as a one-time process prior to the search. Then the performance of the subnets, evaluated with the inherited weights, is used to guide the selection of architectures during search. Early one-shot approaches \cite{one-shot,guo2019single,chu2019fairnas} follow a conventional (rather na\"{i}ve) way to train the supernet, i.e. train a randomly chosen sub-part (subnet) of the supernet directly from randomly initialized weights for each mini-batch (see Fig.~\ref{fig:typical_oneshotnas}). Consequently, the searched subnets need to be re-trained thoroughly from scratch as the performance evaluated with inherited weights are far below the true performance and can only be used as a proxy indicator to compare the relative difference between subnets. 

The \emph{progressive shrinking} algorithm proposed in OnceForAll \cite{onceforall} also trains the supernet in an offline fashion, but differs in three aspects---(i) it pre-trains the supernet at full scale before sampling subnets; (ii) it gradually adds the searched dimensions (kernel size, depth, width) into the search space; and (iii) it uses the full-scale supernet to supervise the training of subnets. However, the supernet weights update is still based on randomly sampled subnets. See Fig.~\ref{fig:progressive_shrinking} for a visualization. Empirically, OnceForAll shows that the supernet trained with progressive shrinking enables subnets with inherited weights to be directly deployed without re-training.

Despite the success shown in OnceForAll, we argue that such an offline training process of supernet is fundamentally limited by the fact that it requires \emph{all} subnets to be learned \emph{simultaneously}. To elaborate, without prior knowledge on the distribution of the optimal subnets for the tasks at hand, the supernet training has to cover the search space of subnets \emph{globally} as the training is performed prior to the search as a one-time process. However, training the supernet weights in such a way that all subnets are optimized simultaneously is practically infeasible. For instance, progressive shrinking \cite{onceforall} sampled roughly 634K\footnote{Estimated based on the batch size of 2,048 and the training epochs of 1,000 provided by the \href{https://drive.google.com/file/d/1IK9N9rkBfmuFiRG4N8ml9Ssa-poC1MZg/view}{OnceForAll paper} \cite{onceforall}} subnets during supernet training, which is less than \num{e-12}\% of the its total subnet volume. Any additional options added to the search space (one more kernel size and expand ratio choice) will require 100x more training epochs (100K vs 1K) to cover the same volume of subnets, which is obviously not scalable. Moreover, we argue that simultaneously training all subnets is also unnecessary as not all subnets are equally important for the tasks at hand. Specifically, existing NAS works have shown that different hardware requires different architectures to be efficient, e.g. CPU favors deeper networks with fewer channels in each layer, while GPU favors shallower networks with more channels in each layer, from the latency perspective \cite{proxylessnas,fbnet}. 

\begin{figure*}[!hbt]
    \centering
    \begin{subfigure}{0.98\textwidth}
    \centering
    \includegraphics[width=0.98\textwidth{}]{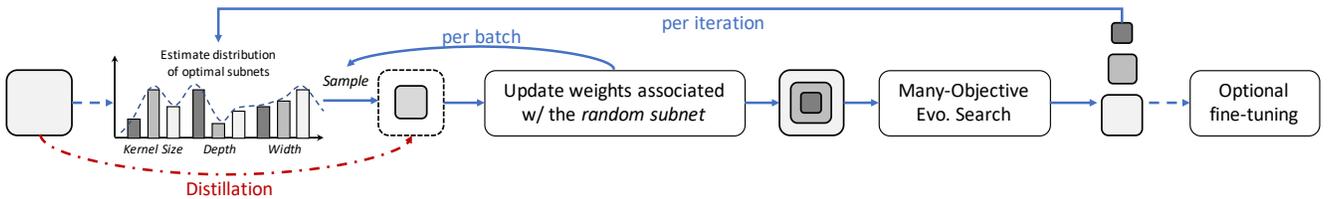}
    \end{subfigure}
\caption{Overview of our proposed \ourmethod{}. The distribution of optimal subnets is estimated from the promising architectures returned by architecture search. Then it is used to guide the training of the supernet. The ``per iteration'' refers to the iteration in Algorithm 1 in the main paper.\label{fig:rebuttal_nat}}
\end{figure*}

To overcome the aforementioned limitations of existing one-shot approaches, we propose \ourmethod{}. The key difference is that \ourmethod{} trains the supernet online. Instead of randomly sampling subnets to train the supernet all at once, \ourmethod{} estimates the distribution (in the variable space) of the optimal subnets from the subnets returned by a many-objective search algorithm, and trains the supernet in correspondence to the estimated distribution. \ourmethod{} does so in a progressive manner, where the estimated distribution and supernet training are gradually refined through iterations (see Fig.~\ref{fig:rebuttal_nat}). We argue that our approach is conceptually more scalable and efficient than existing one-shot approaches since the supernet training now can focus on the promising task-specific subnets recommended by the search algorithm, instead of on all subnets globally.  

To visualize the difference between the existing approach of disentangling supernet training from architecture search, and our approach that use architecture search to guide the supernet training, let us consider the following problem of minimizing a two-variable Rosenbrock function \cite{rosenbrock}:

\begin{equation}
\begin{aligned}
\Minimize & \hspace{3mm} f(x_1, x_2) = (1 - x_1)^2 + 100(x_2 - x_1^2)^2, \\
     & \hspace{3mm} x_1, x_2 \in [-2.048, 2.048].
\end{aligned}
\label{def:rosenbrock}
\end{equation}

\begin{figure*}[!hbt]
    \centering
    \begin{subfigure}{0.8\textwidth}
    \centering
    \includegraphics[width=0.8\textwidth{}]{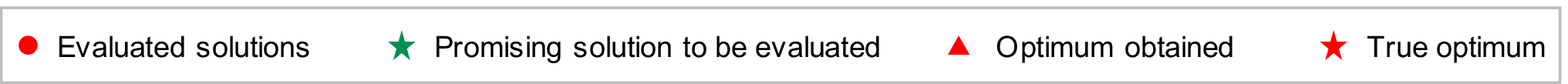}
    \end{subfigure} \\
    \begin{subfigure}[b]{0.24\textwidth}
    \centering
    \includegraphics[width=0.9\textwidth{}]{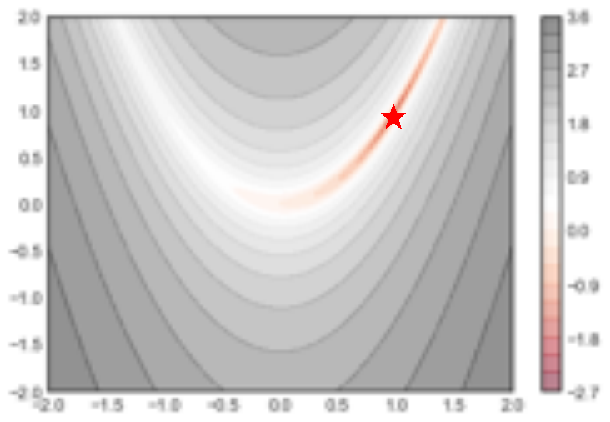}
    \caption{\label{fig:rosenbrock_target}}
    \end{subfigure} \hfill
    \begin{subfigure}[b]{0.74\textwidth}
    \centering
    \includegraphics[width=0.95\textwidth{}]{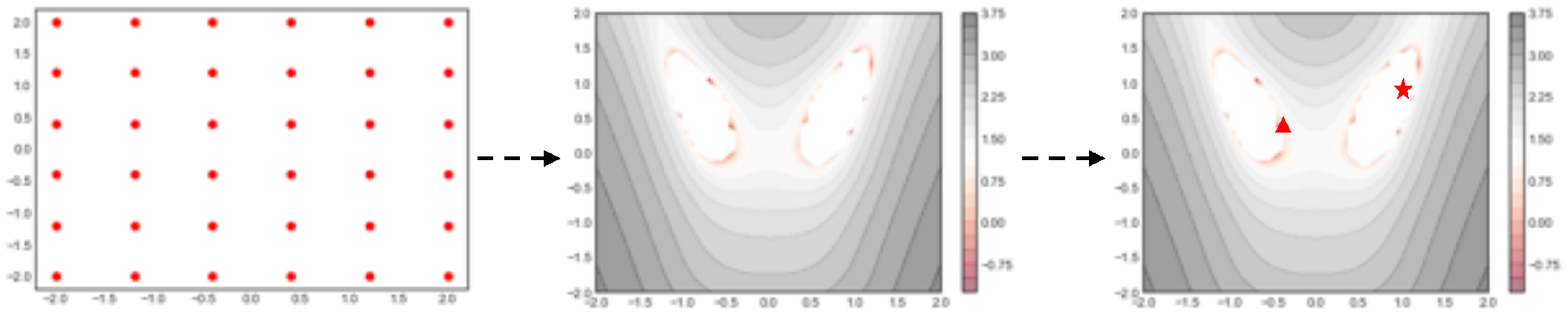}
    \caption{\label{fig:rosenbrock_existing}}
    \end{subfigure} \\
    \centering
    \begin{subfigure}{0.98\textwidth}
    \centering
    \vspace{2mm}
    \includegraphics[width=0.98\textwidth{}]{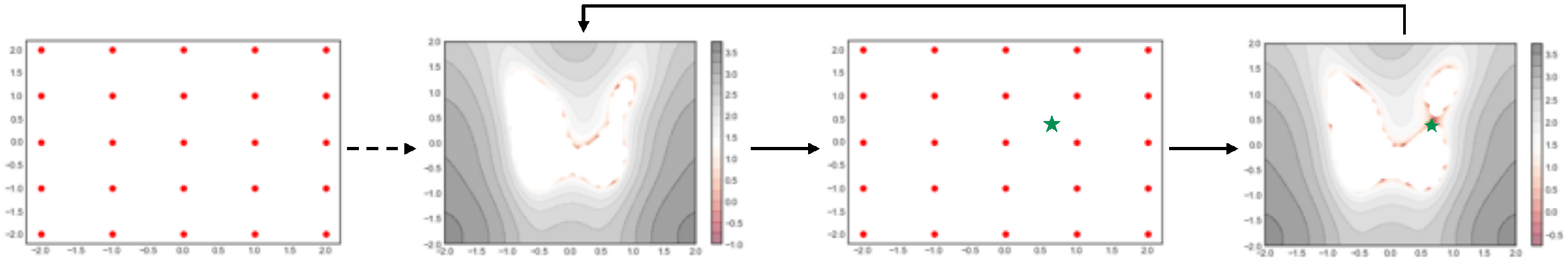}
    \caption{\label{fig:rosenbrock_ours}}
    \end{subfigure}
\caption{(a) True objective landscape (contour) of a two-variable Rosenbrock function. (b) Offline surrogate modelling approach (adopted by existing one-shot NAS methods \cite{guo2019single,chu2019fairnas,onceforall}): the objective landscape is interpolated through uniformly sampled solutions, then the optimization is carried out on the interpolated landscape. (c) Online surrogate modelling approach (ours): a coarse interpolation of the objective landscape is firstly learned using partial budget, then the landscape is gradually refined by adding the optimization outcome on the current landscape to the interpolation. See Fig.~\ref{fig:rosenbrock_results} for comparison on the obtained results.}
\end{figure*}

\begin{figure*}[!hbt]
    \begin{subfigure}[b]{0.74\textwidth}
    \centering
    \includegraphics[width=0.96\textwidth{}]{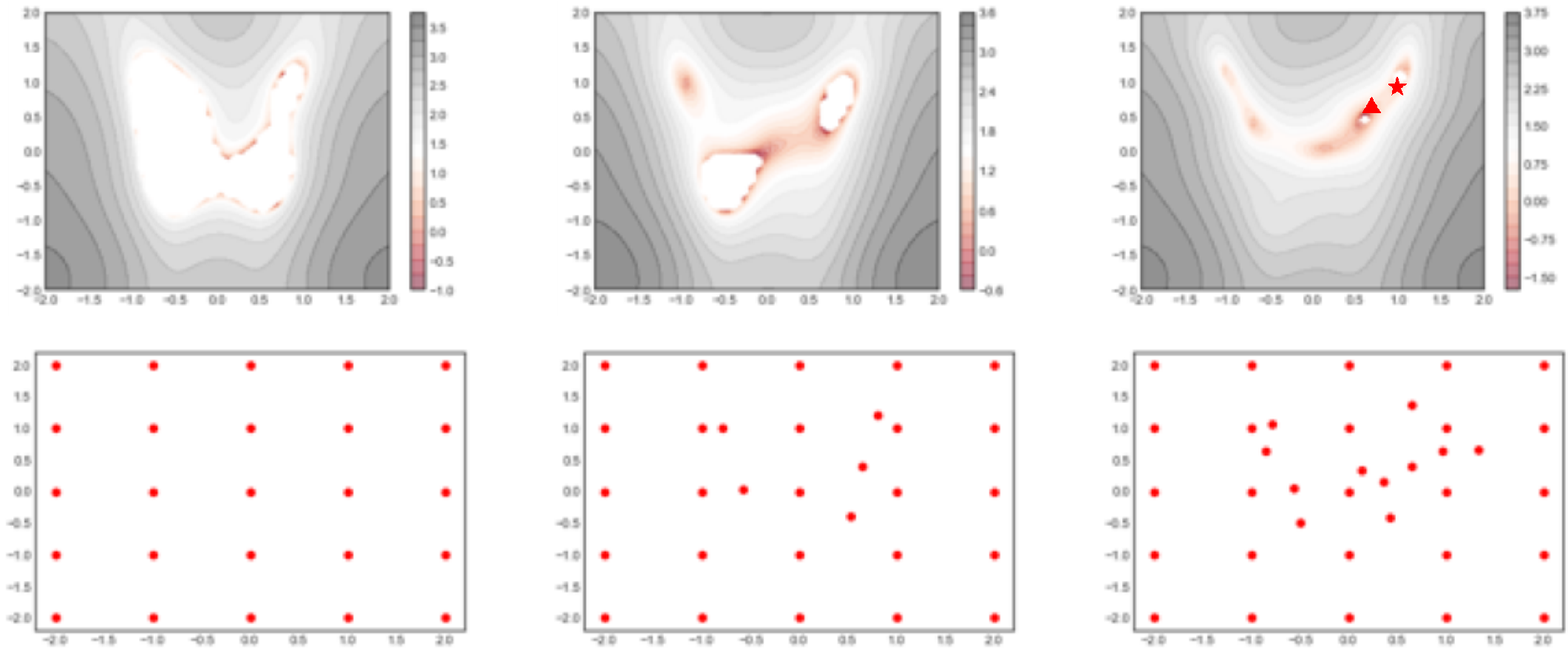}
    \caption{}
    \end{subfigure} \hfill
    \begin{subfigure}[b]{0.24\textwidth}
    \centering
    \includegraphics[width=0.9\textwidth{}]{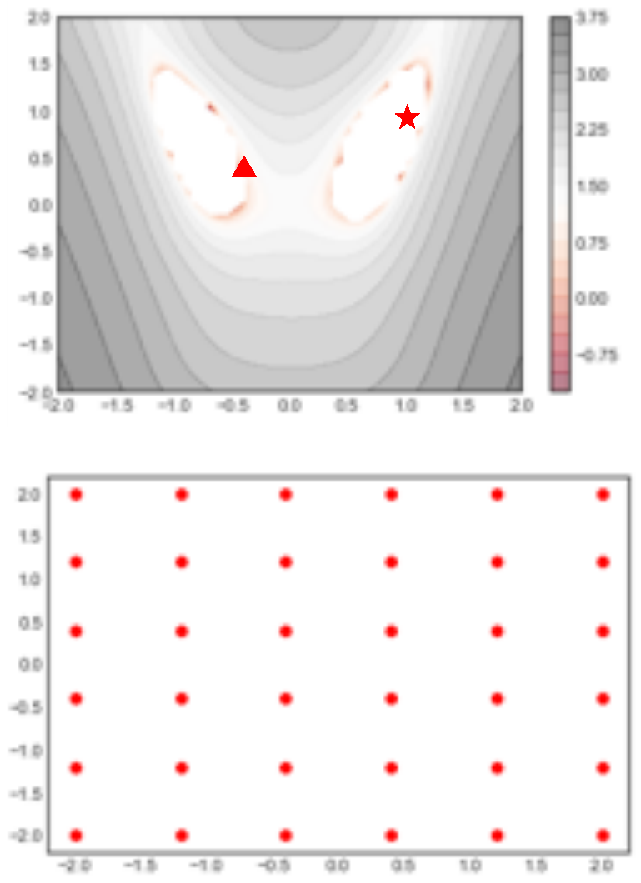}
    \caption{}
    \end{subfigure}
\caption{\textbf{Top row} compares the interpolated landscapes and the obtained optimum by (a) our online surrogate modeling (Fig.~\ref{fig:rosenbrock_ours}) with initial, 3/4, and full budget from \emph{Left} to \emph{Right}, and (b) offline surrogate modeling (existing one-shot NAS approaches; Fig.~\ref{fig:rosenbrock_existing}). \textbf{Bottom row} visualizes the evaluated solutions by the two approaches. Even though the offline approach of uniformly sampling provides a better \emph{global} interpolation of the landscape (i.e. sub-figure (b)), the online approach achieves a better \emph{local} interpolation around the optimum (i.e. sub-figure (a) \emph{Right}). The true landscape is shown in Fig.~\ref{fig:rosenbrock_target}.\label{fig:rosenbrock_results}}
\end{figure*}

\noindent The objective landscape (contour) of the above two-variable Rosenbrock function is shown in Fig.~\ref{fig:rosenbrock_target}. Let's also assume that each function evaluation of $f(x_1, x_2)$ in Eq~(\ref{def:rosenbrock}) is expensive and hence extensively probing the true value is prohibitive (as in the case of NAS). To efficiently optimize this problem, we may learn a meta-model, $\tilde{f}(x_1, x_2)$, to interpolate the landscape (from limited true evaluations). The meta-model should be quick to compute, and hence can be called extensively by an optimization algorithm (as in the case of one-shot NAS). One way is to spend all the true evaluation budget on randomly sampled (from a uniform distribution) solutions at the beginning to learn a meta-model; then the optimization is carried out on the meta-model (as in the case of existing one-shot NAS approaches \cite{guo2019single,chu2019fairnas,onceforall}). See Fig.~\ref{fig:rosenbrock_existing} for a visualization. Another way is to adaptively learn a meta-model in an online fashion. Instead of uniformly exhausting all the true evaluation budget at the beginning, the online approach (as in the case of \ourmethod{}) constructs an initial coarse meta-model from uniformly sampled solutions using partial budget, then a gradual refinement is applied using the solutions optimized based on the current meta-model. See Fig.~\ref{fig:rosenbrock_ours} for a visualization. As shown in Fig.~\ref{fig:rosenbrock_results}, the online approach allows the meta-model to focus on local regions where potential optimal solutions are more likely to reside, eventually leading to a better solution. 

\section{Many-Objective Selection Continued\label{sec:many-continued}}
\begin{figure}[!hbt]
    \centering
    \begin{subfigure}{0.48\textwidth}
    \centering
    \includegraphics[width=0.95\textwidth{}]{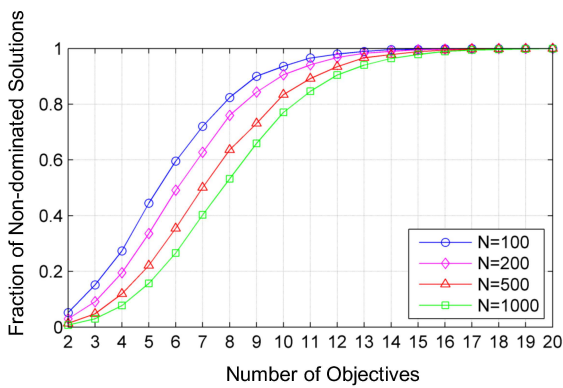}
    \end{subfigure}
\caption{Mean ratio of non-dominated solutions from a set of randomly generated solutions. N is the sample size of the randomly generated solutions.\label{fig:non_dominated}}
\end{figure}

Recall from Section~\ref{sec:selection} in the main paper that \emph{domination} is a widely-adopted partial ordering concept to compare solutions with two or more objectives. It is used to sort solutions into different ranks of importance, where solutions in lower rank are lexicographically better than solutions in higher rank; and solutions in the same rank are non-dominated, i.e. \emph{equally good}. However, as well recognized by the evolutionary many-objective optimization community \cite{nsga3,moeadd}, an increasing larger fraction of randomly generated solutions becomes non-dominated as the number of objectives increases (see Fig.~\ref{fig:non_dominated} for a visualization). As a result, the selection pressure provided from domination diminishes quickly as the number of objectives increases, leading to a slow convergence towards the Pareto front.

\begin{figure}[!hbt]
    \centering
    \begin{subfigure}{0.48\textwidth}
    \centering
    \includegraphics[width=0.95\textwidth{}]{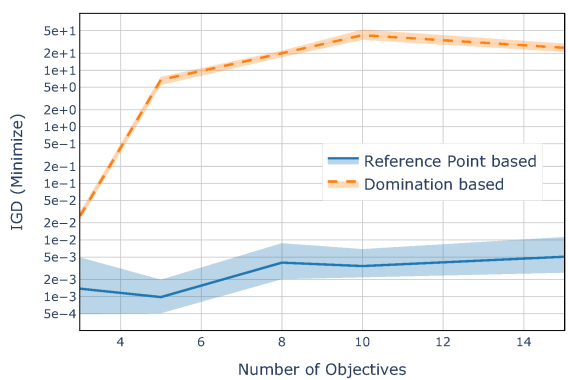}
    \end{subfigure}
\caption{Performance comparison of reference point based (Algorithm 4 in the main paper) and domination based selections \cite{nsga2} on DTLZ1 problem \cite{dtlz}.\label{fig:ref_point_dtlz1}}
\end{figure}

To compensate for the degradation in selection pressure from domination alone, many recently proposed many-objective optimization algorithms \cite{moead,nsga3,moeadd,rvea} opt for the route of reference point based selection, including this work. The reference points serve as a set of pre-defined targets to aid the selection whenever domination concept finds two solutions indistinguishable, i.e. non-dominated. To demonstrate the effectiveness of the reference point based selection, we select the DTLZ1 problem \cite{dtlz}, a benchmark problem that is scalable in number of objectives, and compare the IGD metric\footnote{Note that Hypervolume, another multi-objective performance metric that is used in the main paper, is computationally infeasible to calculate under large numbers of objectives \cite{hv_cal}.} \cite{igd}, a widely-used performance assessment indicator for comparing many-objective optimization algorithms. We vary the number of objectives in DTLZ1 from 3 to 15 and perform 31 independent runs for each selection method. The mean IGD values along with the standard deviations are plotted in Fig.~\ref{fig:ref_point_dtlz1}. The consistently lower IGD values across different numbers of objectives confirm the effectiveness of the reference point based selection method.

\color{black}

\section{Choosing Best Trade-off Solution}\label{sec:decision}
The proposed many-objective EA is expected to produce $N$ (population size) solutions trading-off all $m$ objectives. These solutions are guaranteed to have one property: a gain in one objective between $i$-th and $j$-th solutions comes only from a loss in at least one other objective between them. We calculate the trade-off of $i$-th solution as the average loss per unit average gain among $m$ nearest neighbors ($B(\bm{i})$) based on normalized Euclidean distance are used here), as follows \cite{deb-book-01}:
% \footnotesize
\begin{equation}
\label{def:trade-off}
    \mbox{Trade-off}(\bm{i}) = \max_{j=1}^{|B(\bm{i})|}
 \frac{{\rm Avg. Loss}(\bm{i},\bm{j})}{{\rm Avg. Gain}(\bm{i},\bm{j})}
\end{equation}
where 
\begin{eqnarray*}
{\rm Avg. Loss}(\bm{i},\bm{j}) &=& \frac{\sum_{k=1}^m \max\left(0, f_k(\bm{j}) -
  f_k(\bm{i})\right)}{\sum_{k=1}^m \left\{1| f_k(\bm{j}) >
  f_k(\bm{j})\right\}}\\
{\rm Avg. Gain}(\bm{i},\bm{j}) &=& \frac{\sum_{k=1}^M \max\left(0, f_k(\bm{i}) -
  f_k(\bm{j})\right)}{\sum_{k=1}^m \left\{1| f_k(\bm{i}) >
  f_k(\bm{j})\right\}}
\end{eqnarray*}
% \normalsize
Thereafter, the solutions having the highest trade-off value indicates that it causes the largest average loss in some objectives to make a unit average gain in other objectives to choose any of its neighbors. If this highest trade-off value is much larger statistically than other solutions, then the highest trade-off solution is the preferred choice, in case of no preferences provided from users.

\section{Comparison to Existing ConvNets\label{sec:additional_comparison}}
\begin{figure}[!hbt]
    \centering
    \begin{subfigure}{0.48\textwidth}
    \centering
    \includegraphics[width=0.95\textwidth{}]{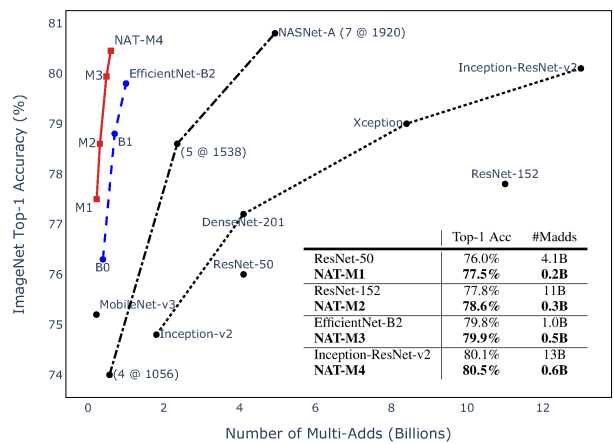}
    \end{subfigure}
\caption{\textbf{MAdds vs. ImageNet Accuracy}. Our \ourmodel{}s significantly outperform other models from NAS algorithms and human experts. In particular, \ourmethod{}-M4 achieves new state-of-the-art 80.5\% top-1 accuracy under mobile setting (600M MAdds).\label{fig:preview_table}}
\end{figure}

Figure~\ref{fig:preview_table} visualizes the \#MAdds-accuracy trade-off curve, where our \ourmodel{}s achieve better top-1 accuracy with much fewer \#MAdds than other CNN models. Notably, \ourmethod{}-M1 is more accurate, and \textbf{20x more efficient} in \#MAdds than ResNet-50 \cite{resnet}; \ourmethod{}-M4 is more accurate, and \textbf{21x more efficient} in \#MAdds than Inception-ResNet-v2 \cite{inceptionv4}. 

\begin{figure*}[!hbt]
    \centering
    \begin{subfigure}{0.33\textwidth}
    \centering
    \includegraphics[width=0.98\textwidth{}]{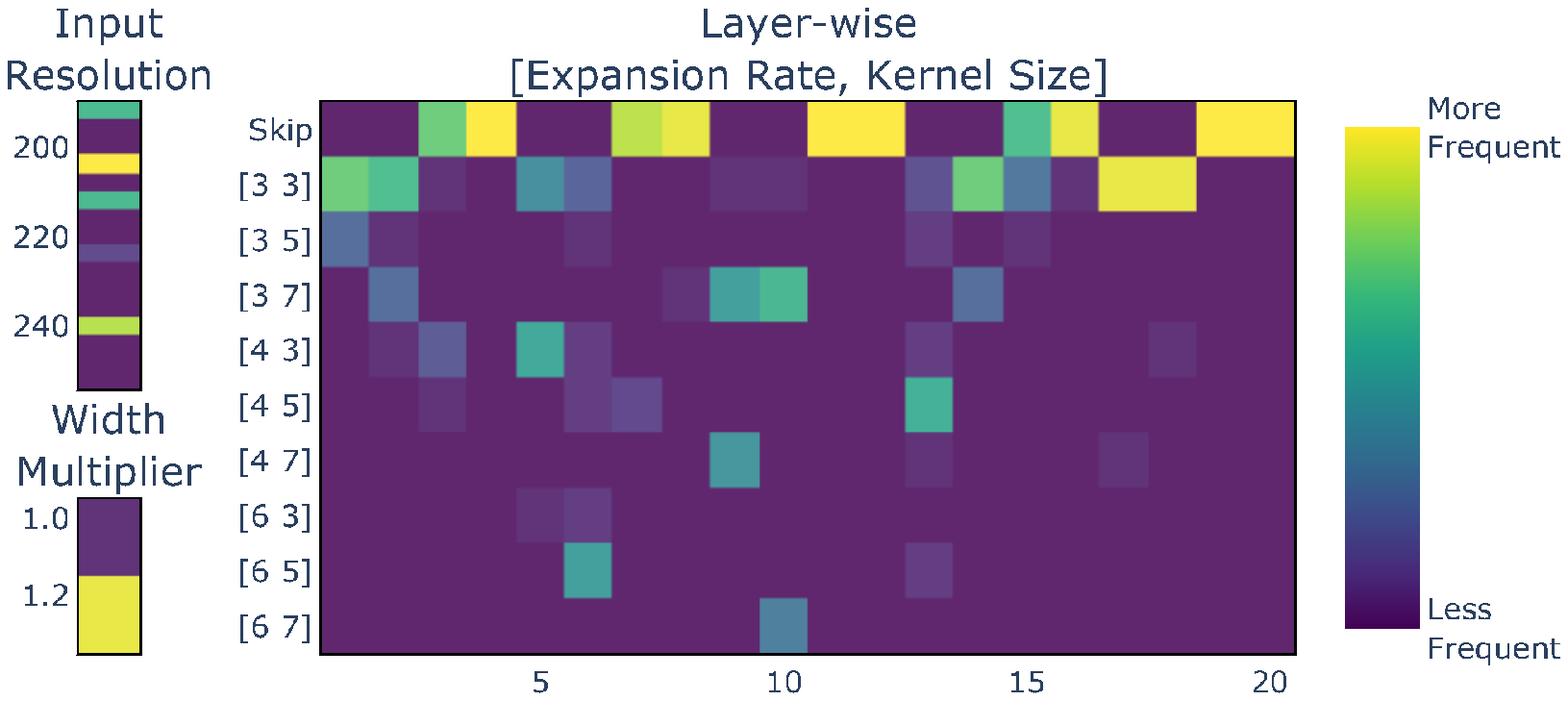}
    \caption{Flowers102\label{fig:flowers102-heatmap}}
    \end{subfigure}
    \begin{subfigure}{0.33\textwidth}
    \centering
    \includegraphics[width=0.98\textwidth{}]{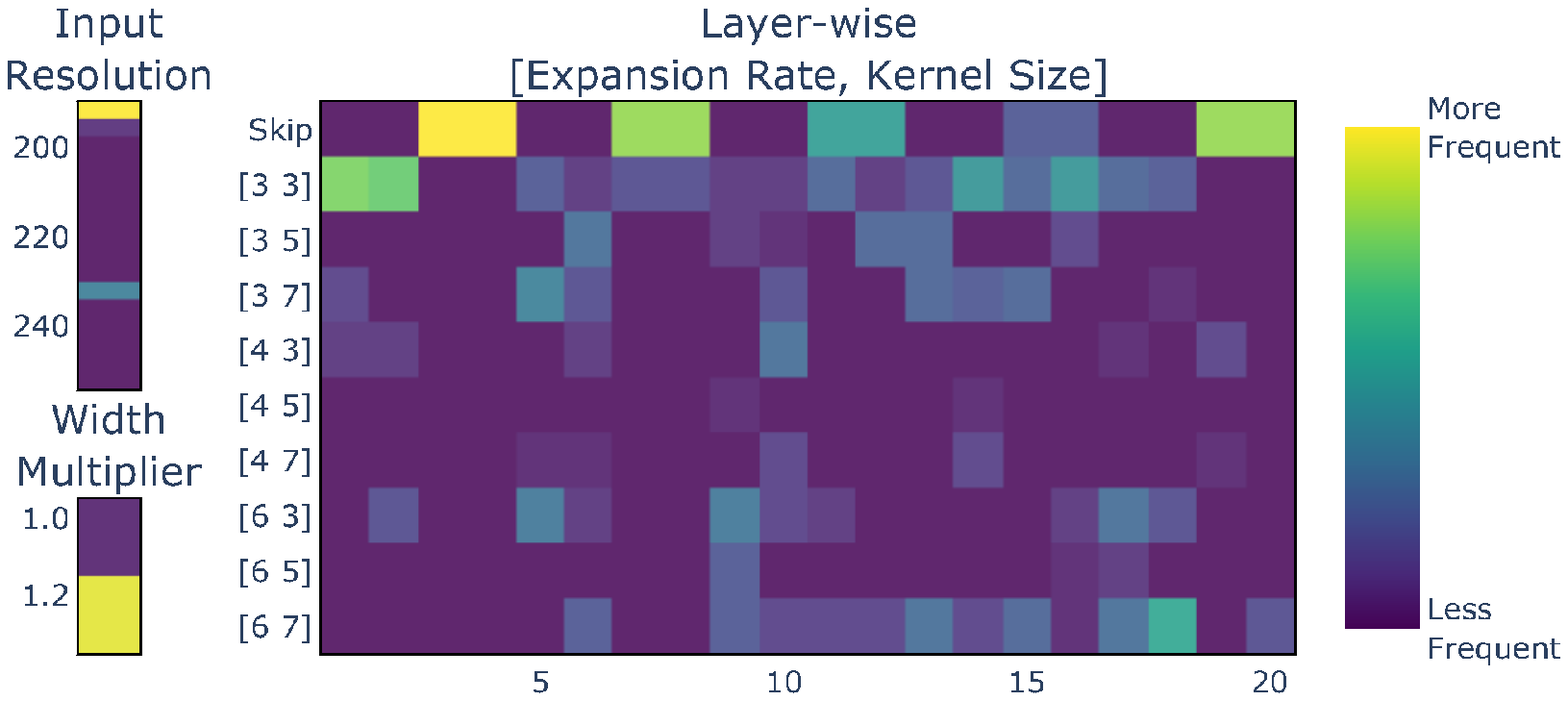}
    \caption{Pets\label{fig:pets-heatmap}}
    \end{subfigure}
    \begin{subfigure}{0.33\textwidth}
    \centering
    \includegraphics[width=0.98\textwidth{}]{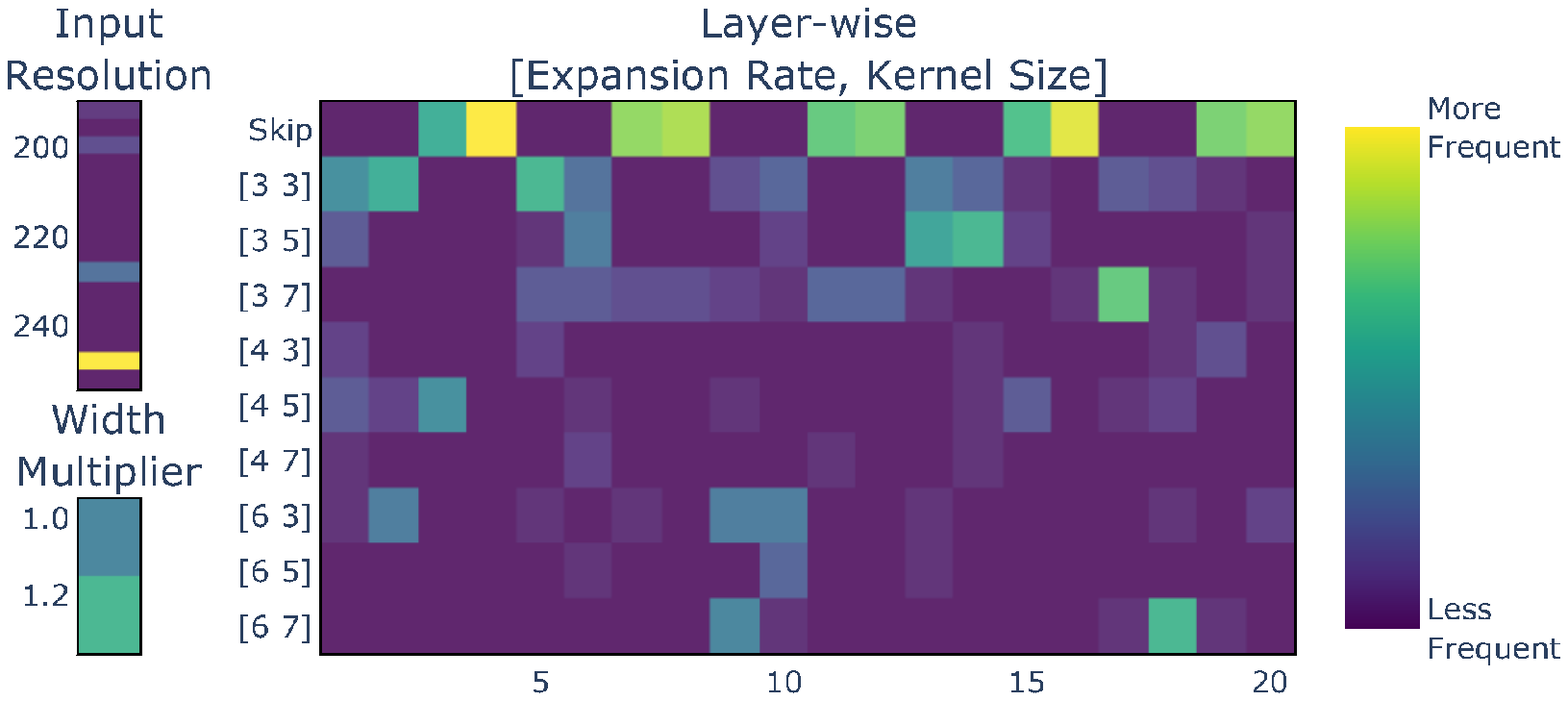}
    \caption{DTD\label{fig:dtd-heatmap}}
    \end{subfigure}\\
    \begin{subfigure}{0.33\textwidth}
    \centering
    \includegraphics[width=0.98\textwidth{}]{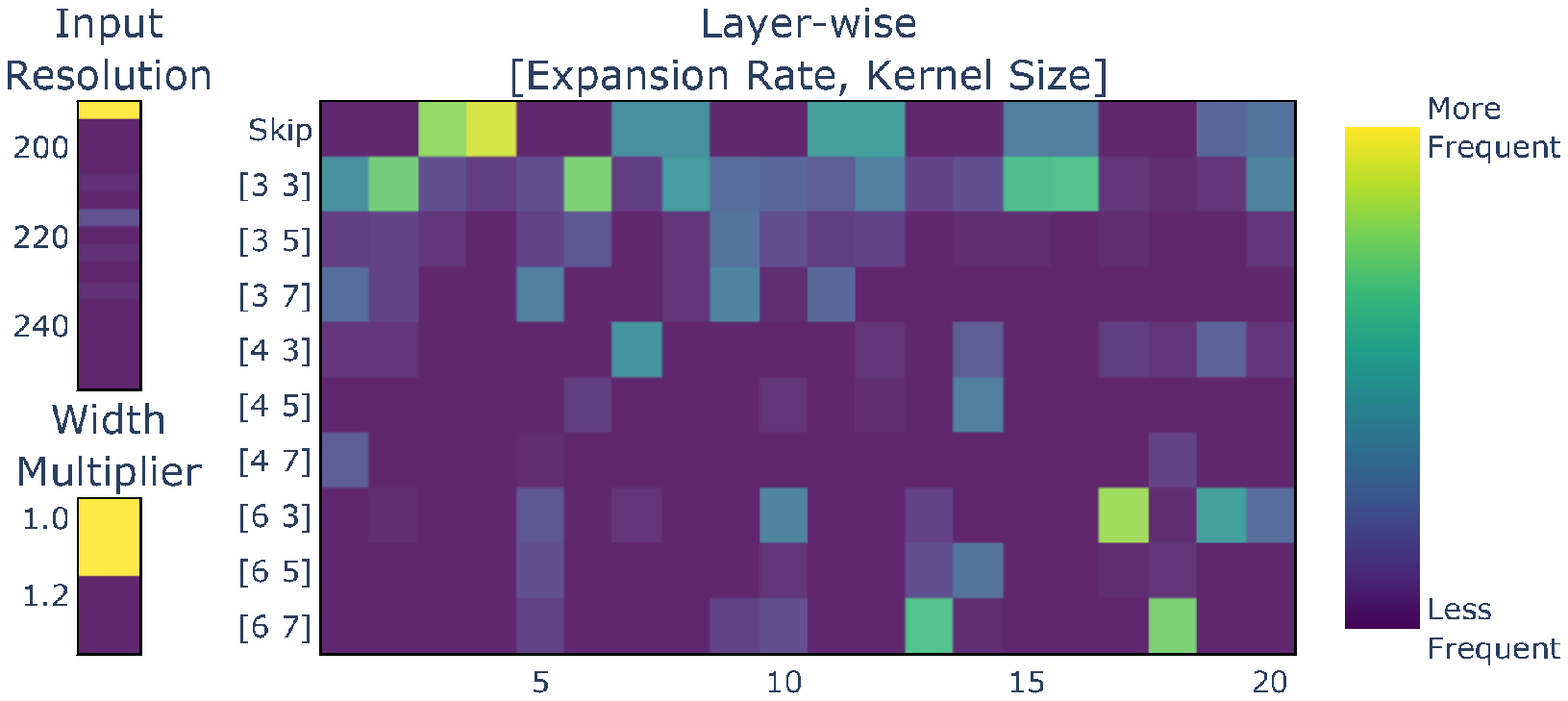}
    \caption{STL-10\label{fig:stl10-heatmap}}
    \end{subfigure}
    \begin{subfigure}{0.33\textwidth}
    \centering
    \includegraphics[width=0.98\textwidth{}]{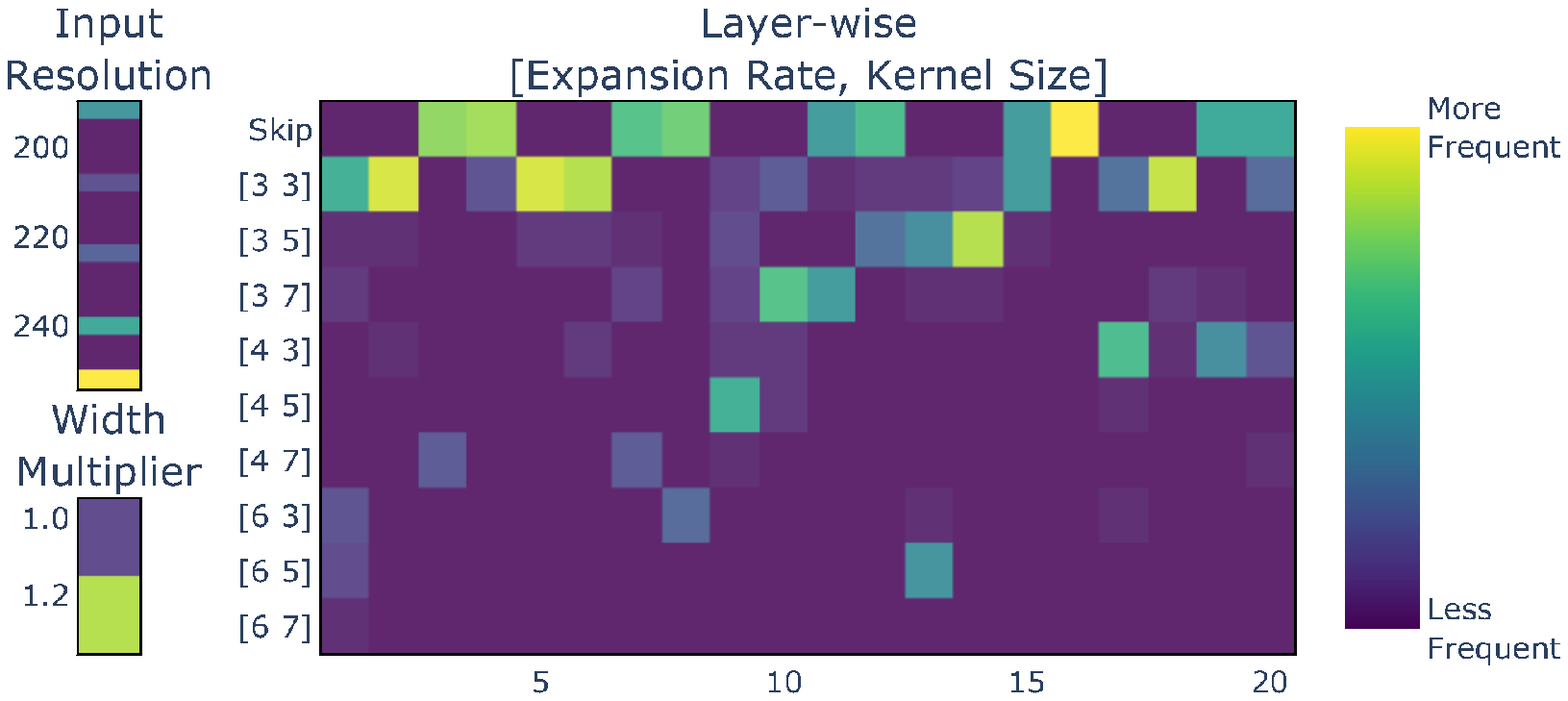}
    \caption{Aircraft\label{fig:aircraft-heatmap}}
    \end{subfigure}
    \begin{subfigure}{0.33\textwidth}
    \centering
    \includegraphics[width=0.98\textwidth{}]{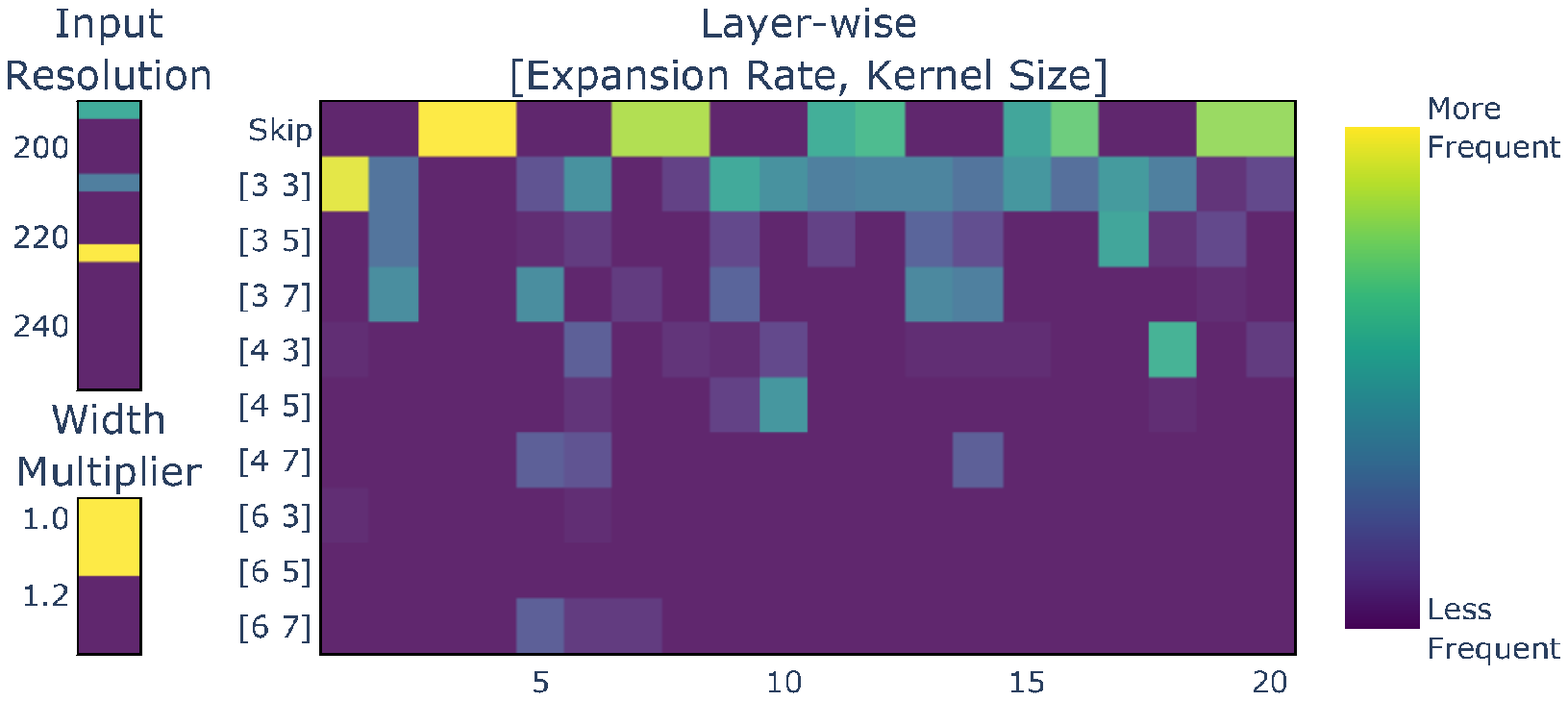}
    \caption{Stanford Cars\label{fig:cars-heatmap}}
    \end{subfigure}\\
    \begin{subfigure}{0.33\textwidth}
    \centering
    \includegraphics[width=0.98\textwidth{}]{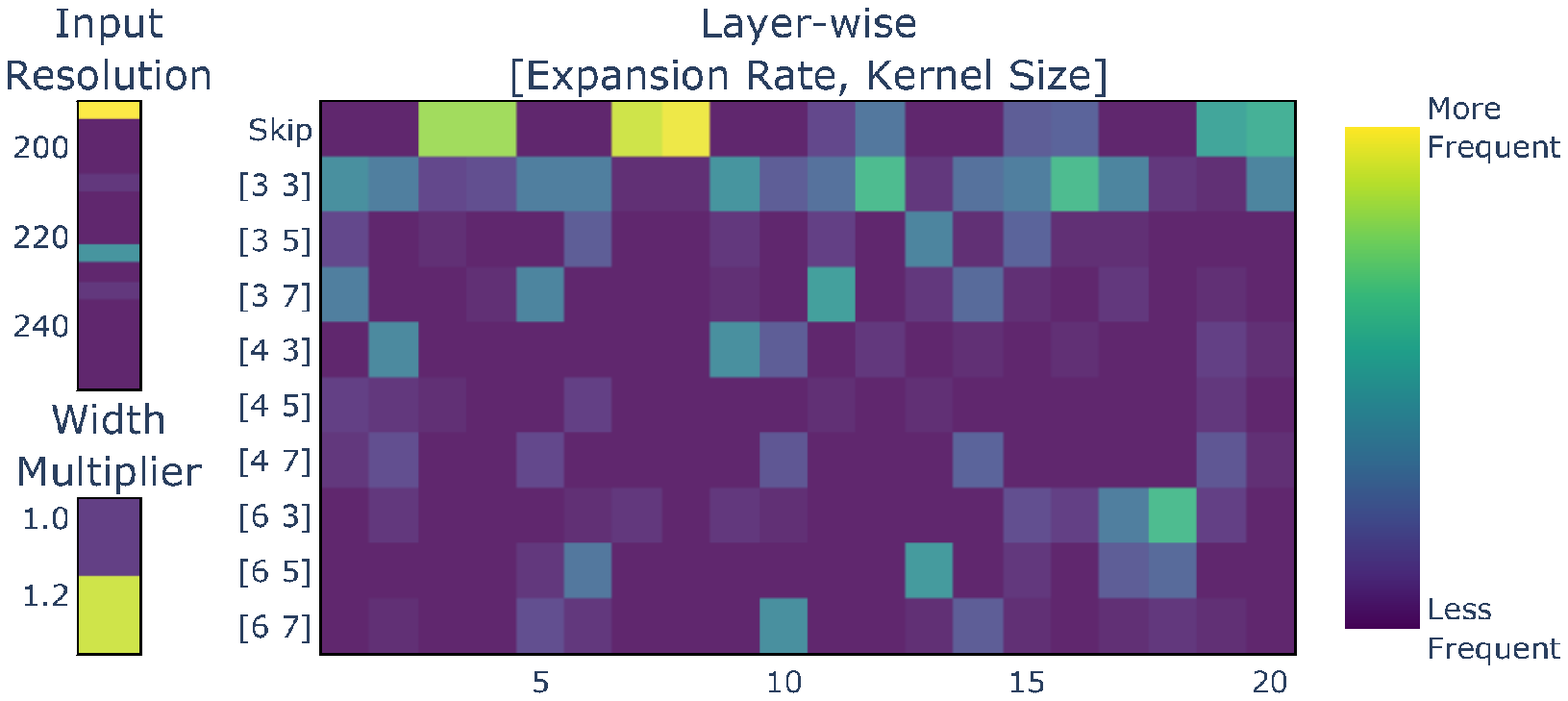}
    \caption{CIFAR-10\label{fig:cifar10-heatmap}}
    \end{subfigure}
    \begin{subfigure}{0.33\textwidth}
    \centering
    \includegraphics[width=0.98\textwidth{}]{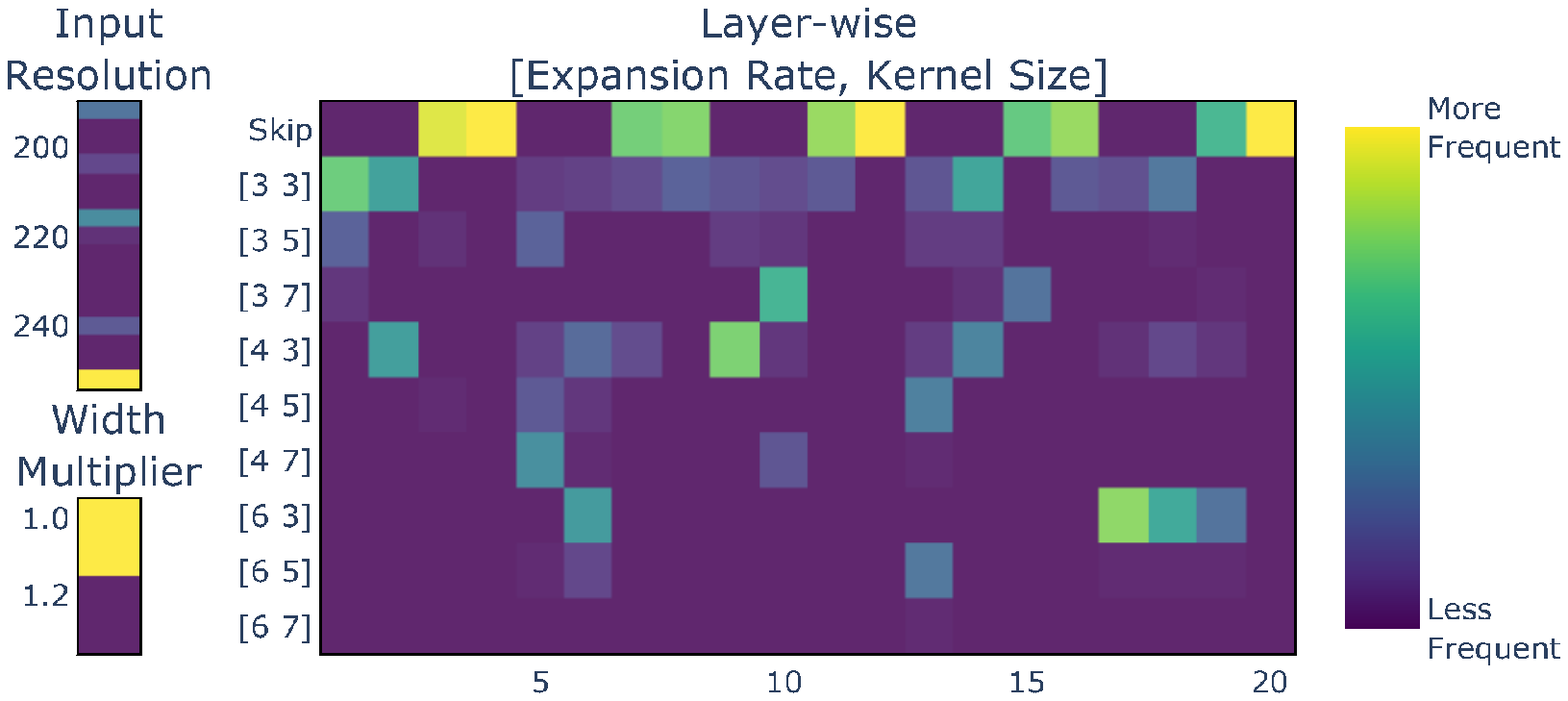}
    \caption{Food-101\label{fig:food101-heatmap}}
    \end{subfigure}
     \begin{subfigure}{0.33\textwidth}
    \centering
    \includegraphics[width=0.98\textwidth{}]{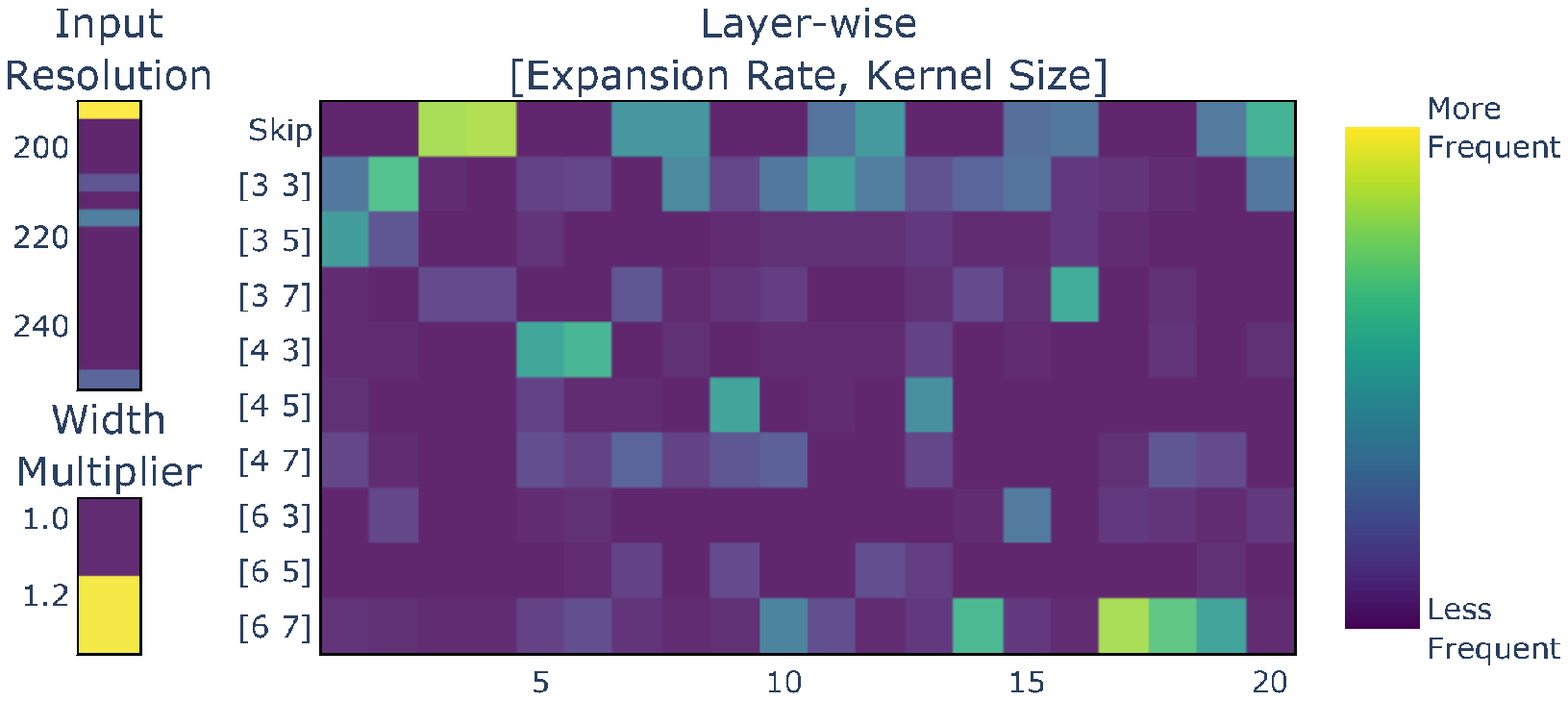}
    \caption{ImageNet\label{fig:imagenet-heatmap}}
    \end{subfigure}
\caption{Non-dominated architectures to \{top-1 accuracy, \#MAdds\} obtained by \ourmethod{} on different datasets.\label{fig:arch_heatmap}}
\end{figure*}

\begin{table*}[!hbt]
\centering
\caption{NAT model performance corresponding to Fig.~9 in main paper.\label{tab:dataset_anno}}
\resizebox{\textwidth}{!}{%
\begin{tabular}{@{\hspace{2mm}}ccc||ccc||ccc||ccc||ccc@{\hspace{2mm}}}
\toprule
\multicolumn{3}{c||}{Flowers102 \cite{flowers102}} & \multicolumn{3}{c||}{Oxford-IIIT Pets \cite{pets}} & \multicolumn{3}{c||}{DTD \cite{dtd}} & \multicolumn{3}{c||}{STL10 \cite{stl-10}} & \multicolumn{3}{c}{FGVC Aircraft \cite{aircraft}} \\\midrule
\#Params & \#MAdds & Top-1 Acc (\%) & \#Params & \#MAdds & Top-1 Acc (\%) & \#Params & \#MAdds & Top-1 Acc (\%) & \#Params & \#MAdds & Top-1 Acc (\%) & \#Params & \#MAdds & Top-1 Acc (\%) \\
3.3M & 152M & 97.5 & 4.0M & 160M & 91.8 & 2.2M & 136M & 76.1 & 4.4M & 240M & 96.7 & 3.2M & 175M & 87.0 \\
3.4M & 195M & 97.9 & 5.5M & 306M & 93.5 & 4.0M & 297M & 77.6 & 5.1M & 303M & 97.2 & 3.4M & 235M & 89.0 \\
3.7M & 250M & 98.1 & 5.7M & 471M & 94.1 & 4.1M & 347M & 78.4 & 7.5M & 436M & 97.8 & 5.1M & 388M & 90.1 \\
4.2M & 400M & 98.3 & 8.5M & 744M & 94.3 & 6.3M & 560M & 79.1 & 7.5M & 573M & 97.9 & 5.3M & 581M & 90.8 \\ \midrule
\multicolumn{3}{c||}{Stanford Cars \cite{stanford_cars}} & \multicolumn{3}{c||}{CIFAR-100 \cite{cifar}} & \multicolumn{3}{c||}{CIFAR-10 \cite{cifar}} & \multicolumn{3}{c||}{Food-101 \cite{food-101}} & \multicolumn{3}{c}{CINIC-10 \cite{cinic10}} \\\midrule
\#Params & \#MAdds & Top-1 Acc (\%) & \#Params & \#MAdds & Top-1 Acc (\%) & \#Params & \#MAdds & Top-1 Acc (\%) & \#Params & \#MAdds & Top-1 Acc (\%) & \#Params & \#MAdds & Top-1 Acc (\%) \\
2.4M & 165M & 90.9 & 3.8M & 261M & 86.0 & 4.3M & 232M & 97.4 & 3.1M & 198M & 87.4 & 4.6M & 317M & 93.4 \\
2.7M & 222M & 92.2 & 6.4M & 398M & 87.5 & 4.6M & 291M & 97.9 & 4.1M & 266M & 88.5 & 6.2M & 411M & 94.1 \\
3.5M & 289M & 92.6 & 7.8M & 492M & 87.7 & 6.2M & 392M & 98.2 & 3.9M & 299M & 89.0 & 8.1M & 501M & 94.3 \\
3.7M & 369M & 92.9 & 9.0M & 796M & 88.3 & 6.9M & 468M & 98.4 & 4.5M & 361M & 89.4 & 9.1M & 710M & 94.8 \\ \bottomrule
\end{tabular}%
}
\end{table*}

\begin{table*}[!hbt]
\centering
\caption{Accuracy predictor model mean (standard deviation) performance corresponding to Fig.~13 in main paper.\label{tab:accuracy_predictor}}
\resizebox{\textwidth}{!}{%
\begin{tabular}{@{\hspace{2mm}}l|c|c|c|c|c|c|c|c|c|c@{\hspace{2mm}}}
\toprule
Method & ImageNet \cite{imagenet} & CIFAR-10 \cite{cifar} & CIFAR-100 \cite{cifar} & Flowers102 \cite{flowers102} & Food-101 \cite{food-101} & Oxford-IIIT Pets \cite{pets} & Aircraft \cite{aircraft} & Stanford Cars \cite{stanford_cars} & DTD \cite{dtd} & STL-10 \cite{stl-10}\\ \midrule
GP & 0.606 (0.09) & 0.969 (0.01) & 0.693 (0.13) & 0.918 (0.02) & 0.980 (0.01) & 0.945 (0.02) & 0.551 (0.17) & 0.964 (0.01) & 0.467 (0.11) & 0.973 (0.11)\\
RBF & 0.705 (0.11) & 0.969 (0.01) & 0.806 (0.08) & 0.932 (0.03) & 0.981 (0.01) & 0.967 (0.01) & 0.693 (0.08) & 0.977 (0.01) & 0.653 (0.06) & 0.979 (0.01) \\
MLP & 0.635 (0.09) & 0.851 (0.06) & 0.562 (0.10) & 0.766 (0.06) & 0.775 (0.09) & 0.798 (0.05) & 0.658 (0.15) & 0.717 (0.10) & 0.490 (0.09) & 0.899 (0.06)\\
DT & 0.625 (0.11) & 0.974 (0.01) & 0.736 (0.11) & 0.940 (0.02) & 0.990 (0.01) & 0.961 (0.01) & 0.629 (0.14) & 0.986 (0.01) & 0.590 (0.14) & 0.976 (0.01) \\
RBF Ensemble & 0.866 (0.04) & 0.959 (0.02) & 0.858 (0.05) & 0.931 (0.01) & 0.967 (0.03) & 0.943 (0.01) & 0.870 (0.07) & 0.975 (0.01) & 0.890 (0.04) & 0.964 (0.02) \\ \bottomrule
\end{tabular}%
}
\end{table*}

\section{Architecture Visualization\label{sec:heatmap}}
One of the main advantages of multi-objective optimization is that it generates a set of non-dominated solutions in a single run. These non-dominated solutions are special in the sense that one has to sacrifice on one objective to gain on another. Thereby, ``mining'' on these non-dominated solutions oftentimes yields important design principles for the task at hand, in this case, to efficiently construct an architecture specific to the objectives and dataset. To demonstrate this concept, we visualize the non-dominated architectures (to maximize top-1 accuracy and minimize \#MAdds) resulting from \ourmethod{} on a diverse set of datasets in Fig.~\ref{fig:arch_heatmap}. Each sub-figure is a heat map showing the distribution of the searched, input image resolutions, width multipliers, and layer settings.

It is clear from Fig.~\ref{fig:arch_heatmap} that even under the same objectives, the optimal architectures for different datasets are different. For example, the most frequent input image resolution is 192 (the lowest value in our searched options) for Oxford-IIIT Pets \cite{pets} and STL-10 \cite{stl-10}. While on FGVC Aircraft \cite{aircraft} and Food-101 \cite{food-101}, the most frequent choice of resolution is 256, which is the highest value in our searched option. Similar observations can be made in case of width multiplier and layer settings. This example provides empirical evidence necessary for finding dataset-specific optimal architectures, as opposed to conventional transfer learning. And as demonstrated in the main paper, our proposed \ourmethod{} presents an efficient and effective way to achieve this goal. 

\begin{figure*}[t]
    \centering
    \begin{subfigure}{0.62\textwidth}
    \centering
    \includegraphics[width=\textwidth{}]{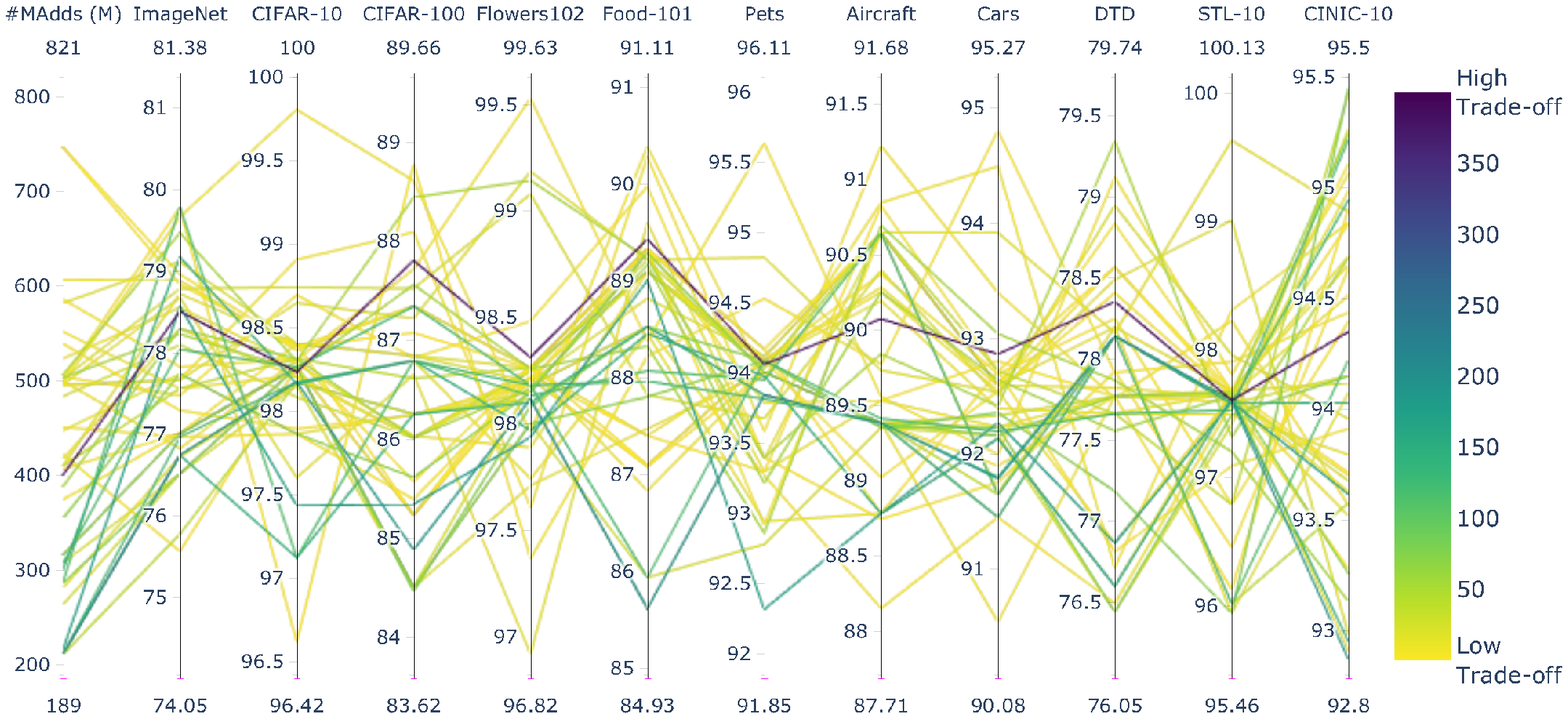}
    \end{subfigure}
    \centering
    \begin{subfigure}{0.33\textwidth}
    \centering
    \includegraphics[width=\textwidth{}]{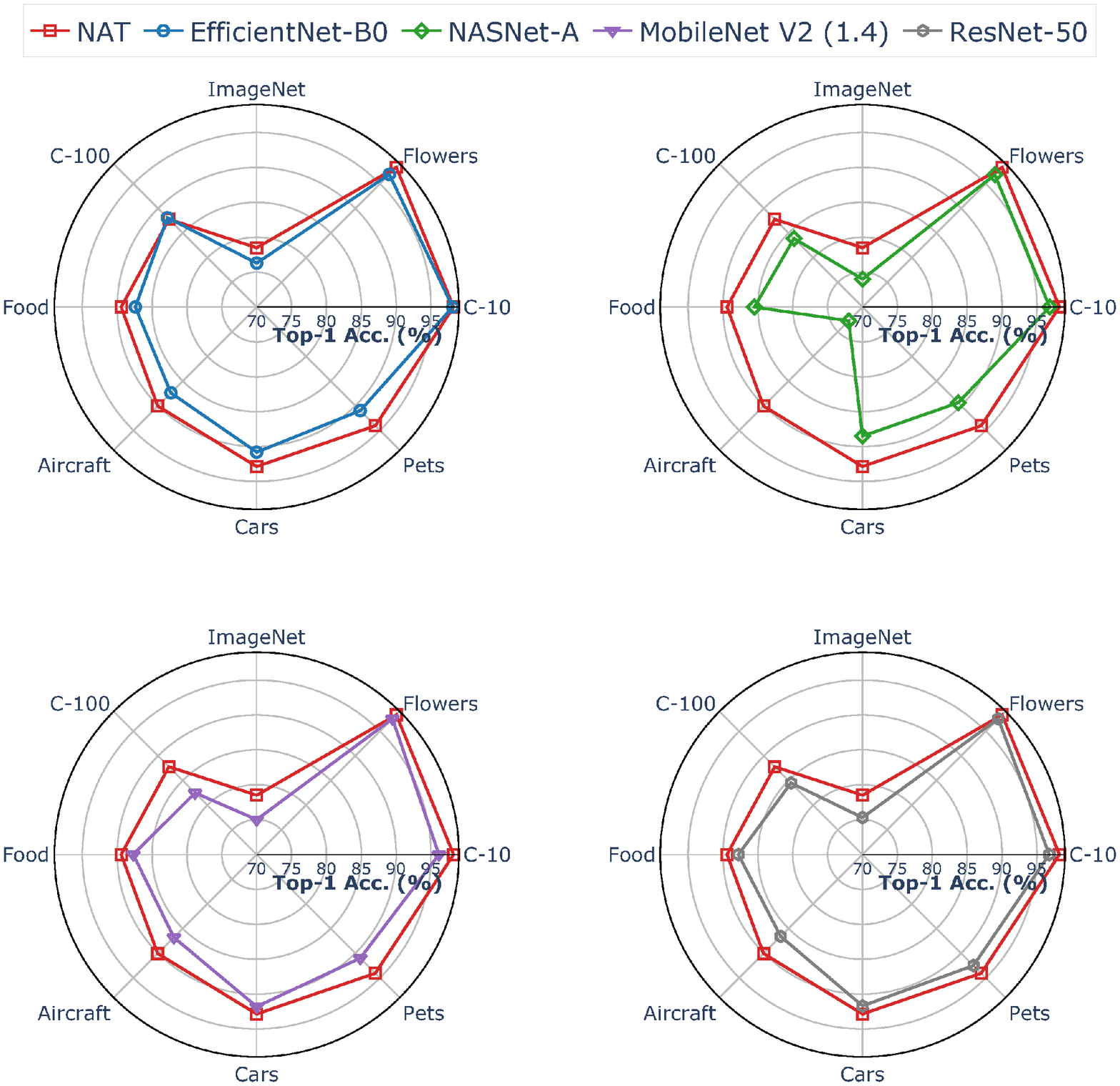}
    \end{subfigure}
\caption{\textbf{Left:} Parallel Coordinate Plot (PCP) where each vertical bar is an objective and each line is a non-dominated architectures achieved by \ourmethod{} from a 12-obj optimization of minimizing \#MAdds and maximizing accuracy on the 11 datasets. The model with the best trade-off (see Section~\ref{sec:one-shot} for details) is highlighted in dark blue. \textbf{Right:} 1-on-1 comparison between the selected \ourmodel{} (top-ranked in trade-off) and representative peer models on top-1 accuracy on various datasets. Method with larger area is better.\label{fig:many_obj}}
\vspace{-0.3cm}
\end{figure*}

\section{Scalability to Objectives Continued\label{sec:twelve_obj}}
To further validate the scalability of \ourmethod{} to a large number of objectives, we consider the top-1 accuracy on each of the 11 datasets shown in Table~3 (main paper) along with \#MAdds, as separate objectives, resulting in a 12-objective optimization problem. Not only is such a large-scale optimization plausible with \ourmethod{}, it also reveals important information, which a low-dimensional optimization may not. During search, the accuracy on each dataset is computed by inheriting weights from the dataset-specific supernets generated from previous experiments (Section~\ref{sec:dataset} in the main paper). Since the supernets are already adapted to each dataset, we exclude the supernet adaptation step in \ourmethod{} for this experiment.

Fig.~\ref{fig:many_obj} (Left) shows the 12 objective values for all 45 non-dominated architectures obtained by \ourmethod{} in a parallel coordinate plot (PCP), where each vertical bar is an objective and each line connecting all 12 vertical bars is an architecture. We now apply the trade-off decision analysis presented in Section~\ref{sec:one-shot} and observe that the highest trade-off solution is more than $(\mu+3\sigma)$ trade-off away from the rest of 44 solutions. This solution is highlighted in dark blue in Fig.~\ref{fig:many_obj} (Left). Its intermediate performance in all objectives indicate that this best trade-off solution makes a good compromise on all 12 objectives among all 45 obtained solutions. In Fig.~\ref{fig:many_obj} (Right), we compare this solution with different baseline models that are fine-tuned to each dataset separately. Notably, our \ourmodel{} achieves better accuracy on all datasets with similar or less \#MAdds than EfficientNet-B0 \cite{efficientnet}, MobileNetV2 \cite{mobilenetv2}, NASNet-A \cite{nasnet}, and ResNet-50 \cite{resnet}, making our highest trade-off solution a preferred one.

The above analysis alludes to a computational mechanism for choosing a single preferred trade-off solution from the Pareto solutions obtained by a many-objective optimization algorithm. If such an overwhelmingly high trade-off solution exists in the Pareto front, it becomes one of the best choices and can outperform solutions found by a single-objective optimization algorithm. Without resorting to a many-objective optimization to find multiple trade-off solutions, identification of such a high trade-off solution is very challenging.

\end{appendices}

\end{document}